\documentclass{article}
\usepackage{thestyle,times}

\iclrpreprintcopy

\usepackage{amsmath,amsfonts,bm}

\def\eqref#1{equation~\ref{#1}}

\def\1{\bm{1}}

\DeclareMathAlphabet{\mathsfit}{\encodingdefault}{\sfdefault}{m}{sl}
\SetMathAlphabet{\mathsfit}{bold}{\encodingdefault}{\sfdefault}{bx}{n}

\usepackage{hyperref}
\usepackage{url}
\usepackage{graphicx}
\usepackage{float}
\usepackage{amsmath,amssymb,amsthm,mathtools}
\usepackage{wrapfig}
\usepackage{booktabs}
\usepackage{glossaries}     

\usepackage{subcaption}
\usepackage{tabularx}
\usepackage{multirow}
\usepackage{tcolorbox}
\usepackage{listings}

\usepackage{xcolor}

\usepackage{xcolor}
\usepackage{listings}

\definecolor{vsbg}{HTML}{F5F5F5}
\definecolor{vstext}{HTML}{000000}
\definecolor{vsvariable}{HTML}{001080}
\definecolor{vscontrol}{HTML}{AF00DB}
\definecolor{vskeyword}{HTML}{0000FF}
\definecolor{vsfunction}{HTML}{795E26}
\definecolor{vsnamespace}{HTML}{267F99}
\definecolor{vsstring}{HTML}{A31515}
\definecolor{vsnumber}{HTML}{098658}
\definecolor{vscomment}{HTML}{008000}
\definecolor{vslinenum}{HTML}{237893}

\lstdefinestyle{pythoncode}{
  language=Python,
  basicstyle=\ttfamily\small\color{vstext},
  identifierstyle=\color{vsvariable},
  keywords=[1]{import,from,as,for,in,if,elif,else,while,return,break,
               continue,pass,try,except,finally,raise,with,yield,await,async},
  keywordstyle=[1]\color{vscontrol},
  keywords=[2]{def,class,lambda,True,False,None,and,or,not,is},
  keywordstyle=[2]\color{vskeyword},
  emph=[1]{make_vec,reset,step,propose_embodiment,policy,range},
  emphstyle=[1]\color{vsfunction},
  emph=[2]{codesign_gym,cgym},
  emphstyle=[2]\color{vsnamespace},
  commentstyle=\color{vscomment},
  stringstyle=\color{vsstring},
  literate={0}{{\color{vsnumber}0}}1 {1}{{\color{vsnumber}1}}1
           {2}{{\color{vsnumber}2}}1 {3}{{\color{vsnumber}3}}1
           {4}{{\color{vsnumber}4}}1 {5}{{\color{vsnumber}5}}1
           {6}{{\color{vsnumber}6}}1 {7}{{\color{vsnumber}7}}1
           {8}{{\color{vsnumber}8}}1 {9}{{\color{vsnumber}9}}1,
  backgroundcolor=\color{vsbg},
  frame=single,
  rulecolor=\color{vsbg},
  framesep=6pt,
  xleftmargin=2.2em,
  framexleftmargin=1.8em,
  numbers=left,
  numberstyle=\tiny\color{vslinenum},
  numbersep=8pt,
  showstringspaces=false,
  columns=fullflexible,
  keepspaces=true,
  breaklines=true,
  tabsize=4,
}

\lstdefinestyle{jsonstyle}{
    basicstyle=\ttfamily\small,
    breaklines=true,
    frame=none,
    showstringspaces=false,
    keywordstyle=\color{blue},
    stringstyle=\color{red},
    commentstyle=\color{green!50!black},
}

\newtcolorbox{promptbox}{
    colback=gray!10,     
    colframe=gray!50,    
    boxrule=0.5pt,       
    arc=3pt,             
    left=6pt, right=6pt, top=6pt, bottom=6pt,
}

\theoremstyle{definition}

\title{Co-Design Gym: A Unified Benchmark for Embodiment-Policy Co-Optimization}

\author{
\vspace{4px}
Aviraj Newatia$^{*,1}$ \And
Yordan Tsvetkov$^{*,1}$ \And
Leonard Pleiss$^{2}$ \And
Andrew Spielberg$^{3,4}$ \And
Rika Antonova$^{1}$\\[0.5em]
\vspace{4px}
$^{*}$Equal contribution\hspace{1em}
$^{1}$University of Cambridge \\
\vspace{4px}
$^{2}$Technical University of Munich\hspace{1em}
$^{3}$Carnegie Mellon University \hspace{1em}
$^{4}$MorphoAI \hspace{1em}
}

\newcommand{\hlfigdir}{expr_figures}

\newcommand{\embody}{\mathbf{e}}
\newcommand{\embodyset}{\mathbf{E}}

\newacronym{acr:rl}{RL}{Reinforcement Learning}

\begin{document}

\maketitle
\vspace{-1.5em}

\begin{abstract}
Finding an optimal behaviour policy within a given environment is a widely studied problem in domains as diverse as games, robotics, energy infrastructure, communication networks, and multi-agent systems. Numerous benchmarks have been developed to support such research, but the vast majority assume that  the agent's embodiment (design) is fixed, focusing instead on policy learning alone.
\mbox{Lifting} this assumption gives rise to a broader class of problems in which optimizing embodiment and policy separately is highly suboptimal. An agent's embodiment strongly shapes which control policies can be discovered, while the optimal embodiment is in turn defined by the policies it admits. To help the research community study this class of problems explicitly and systematically, we introduce \textsc{Co-Design Gym} -- a suite of benchmark environments for jointly optimizing embodiment and policy. Our environments span domains such as robotic manipulation and locomotion, multi-robot cooperation, deformable and soft dynamics, video games, electricity grids, wireless networks, F1 racing, multi-agent warehouses, and optimal control, offering 20 environment families (domains), with over 85 distinct co-design presets in total. We further contribute a systematic evaluation of representative co-design algorithms, characterizing the current state of the art. Together, these contributions lay the groundwork for cumulative, comparable progress in co-design. 

\end{abstract}

\vspace{-1em}

\begin{figure}[H]
    \centering
    \includegraphics[width=0.8\linewidth]{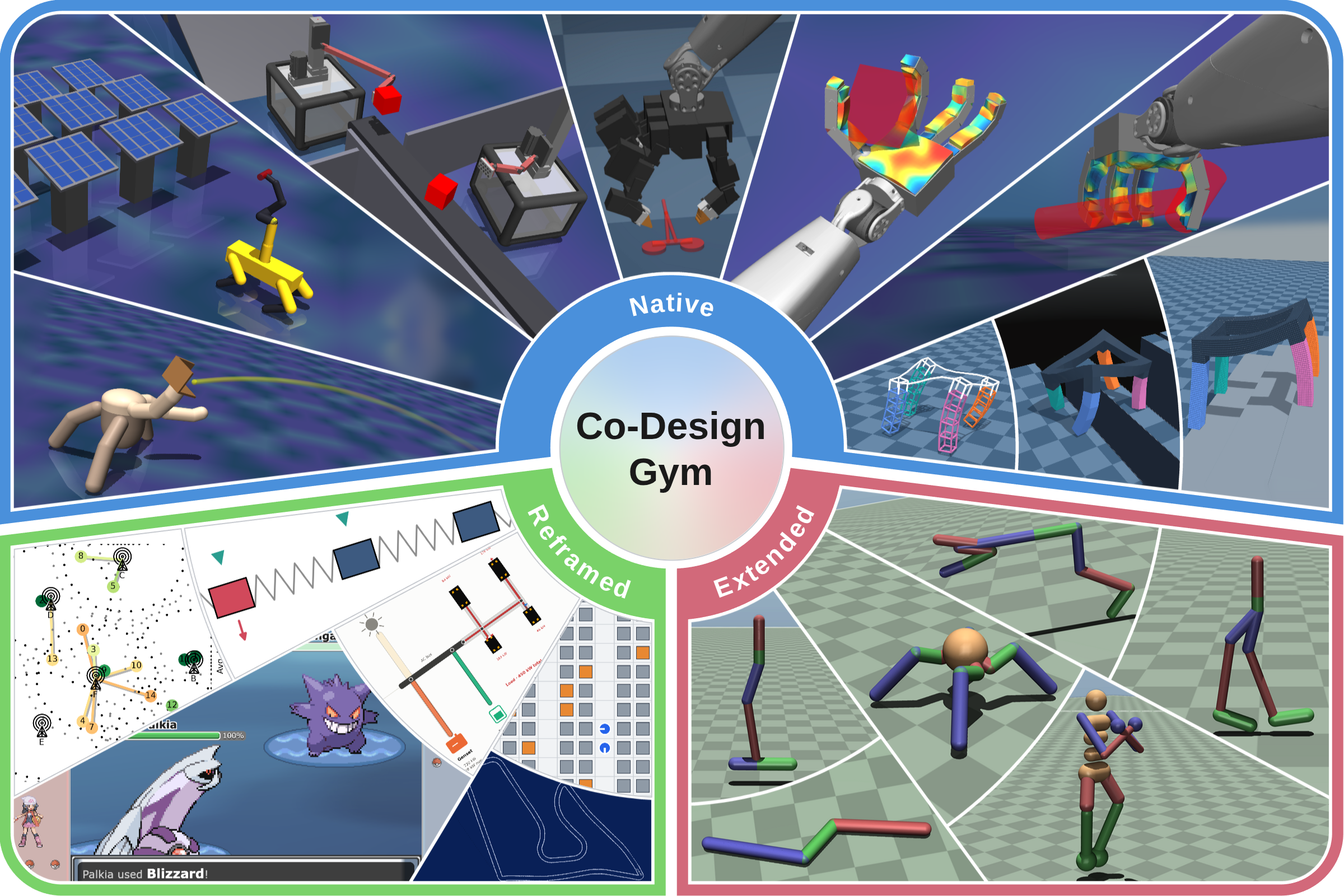}
    \captionsetup{skip=5pt}
    \caption{Co-Design Gym overview with three environment categories. 
    \textbf{Native:} new domains we developed to study the dynamic interplay between embodiment and policy, covering manipulation, locomotion and deformable body dynamics.
    \textbf{Extended:} classic MuJoCo control tasks from \mbox{OpenAI} Gym extended to support morphology variations.
    \textbf{Reframed:} existing problems from games, \mbox{energy}, networking, racing, logistics and control, recast as co-design tasks.
    \textit{Co-Design Gym is available at} 
    \textit{\url{https://camral.github.io/codesigngym/}}}
    \label{fig:main}
    \vspace{-1.0em}
\end{figure}

\section{Introduction}

An agent's ability is a function of both its \textit{embodiment} and its behavioral \textit{policy}.  An embodiment determines what an agent is \textit{able to do}: it is an interface for transforming perception into action upon an external environment.  
The policy determines what the agent  \textit{actually does}, using a given  physical interface. The process of jointly improving the embodiment and the policy toward performance goals is referred to as \textit{co-design}.

Embodiment and policy are coupled: the quality of an embodiment is the performance of the best policy it can achieve for downstream tasks. Hence, fully assessing each candidate embodiment requires finding an optimal policy for it on a variety of tasks. However, running expensive policy optimization to completion before knowing the potential of an embodiment is wasteful. Tighter co-optimization is essential, especially when the embodiment and policy subspace structures yield challenging high-dimensional optimization landscapes.

The above calls for extending optimization paradigms such as Reinforcement Learning (RL), usually framed as singly non-stationary optimization. Recognizing co-design as  a \textit{doubly \mbox{non-stationary}} problem is more suitable, since both the policy and the embodiment can change rapidly. 
Finding global or even high-quality local optima for this extended problem is expensive, and gauging real progress is nontrivial. To date, no unified benchmark tracks the performance of co-design methods across a variety of domains, or disentangles progress in embodiment search and policy optimization from progress specific to co-design.

Co-design today stands in a similar place to where RL stood before the introduction of unified benchmarks, when agents still struggled to solve basic tasks~\citep{Mnih2013PlayingAW}. Benchmarks and unifying frameworks, \textit{e.g.} OpenAI Gym and Gymnasium~\citep{brockman2016openai, Towers2024GymnasiumAS}, enabled compound progress that accelerated the field. RL agents have since reached superhuman performance in a wide variety of areas, such as strategic game playing in Go~\citep{Silver2016MasteringTG} \& Dota~2~\citep{Berner2019Dota2W}, drone racing \citep{geles2026superhumansafeagileracing}, control of tokamak plasmas for nuclear fusion \citep{Degrave2022MagneticCO}, and robot locomotion~\citep{peng2017deeploco, peng2018deepmimic, margolis2021learning}, to name a few.  A unified benchmark for co-design should spark comparable progress.

In pursuit of that, we introduce \textsc{Co-Design Gym} -- a unified benchmark for co-design algorithms. Our contribution is threefold. 
First, we extend the canonical Gymnasium API with an explicit design space, so that any policy-learning task can be lifted into a co-design task. 
Second, we leverage this generalized API to convert prominent existing policy optimization environments into co-design tasks. 
Finally, we present novel co-design environments in which embodiment optimization is \mbox{decisive} for what the policy can achieve: from reshaping fingertips for precise manipulation of small objects \& reshaping a hand for fast in-hand reorientation, to dynamic ball catching, multi-robot truck unloading, and evolving soft robots for locomotion.
Overall, we offer 20 environment families and over 85 presets, with accelerated GPU-based resets that make it practical to evaluate thousands of embodiments in parallel. Benchmarking representative evolutionary, reinforcement-learning, hybrid and transformer-based co-design methods under matched compute budgets, we find that no single method dominates, and that many tasks remain far from solved. 
This leaves substantial headroom for new methods, and we call upon the community to take advantage of this opportunity.

\section{Background and Related Work}
\label{sec:background}

Co-design problems appear naturally across disciplines in design tasks that feature both a static embodiment and a time-varying policy component, and in which choices made in each domain can affect choices in the other.  Over the last several decades, methods have been developed for the co-design of computer systems hardware and software \citep{de1997hardware, milder2012computer}, optics and their processing algorithms \citep{dowski1995extended}, aeronautical materials and stabilizers \citep{salama1989recent}, and chemical plants and processes \citep{luyben1994analyzing}, just to name a few.  More recently, co-design of the embodiments and policies of virtual and physical robots has seen growing interest. This problem is exceptionally relevant now that investment in robotic hardware is at all-time high: advanced hardware prototyping is finally affordable, thus co-design solutions don't have to be confined to simulation. Robots typically must solve tasks that involve a complex combination of dynamical moves, medium-to-long-term-planning, a mix of discrete and continuous optimization in both morphological structure and actions all at once, while still enabling fast (real-time) operation of the whole system. A further challenge is to transfer the success from simulation to reality. All these aspects make co-optimization of robot systems particularly challenging, yielding an ideal ground for testing novel co-design methods.

Early research built upon the seminal works of \cite{ventrella1994explorations} and \cite{sims1994evolving} and employed evolutionary algorithms \citep{hiller2011automatic, cheney2014unshackling, cheney2014evolved, corucci2016evolving}, a trend that still sees application in modern learning-based approaches \citep{wang2019neural, yuan2021transform2act} and are enjoying a renaissance in combination with off-the-shelf large models \citep{chen2026natural, gao2026vlmgineer}.  While these approaches produce robust specimens \citep{bongard2011morphological, hejna2021task} their sampling-based search can be inefficient when simulation time becomes the computational bottleneck.  In such cases, complementary approaches that use gradient-based search tend to more efficient \citep{matthews2023efficient}.  Given continuous and smoothly differentiable parameterizations of morphology and controllers, these approaches employ unconstrained and constrained optimizers to search over optimal parameterizations of geometry \citep{wampler2009optimal, coros2013computational, spielberg2017functional, ha2017joint, ma2021diffaqua}, materials, \citep{spielberg2023advanced}, and controllers and observers \citep{spielberg2019learning} for solving specified tasks.  As the field has begun to mature in the last decade, holistic frameworks have emerged that have driven toward a unifying theory of co-design \citep{censi2015mathematical, zardini2023co}. Sampling- and gradient-based approaches have also begun to merge under the umbrella of generative systems \citep{wang2023diffusebot, zhao2020robogrammar, xu2021multi, dai2026stackelbergppo, ha2026transformer}.  

Though much of the above work focuses on the algorithmic nature of co-design algorithms, co-designed creatures are not confined to the virtual world. Successful physical demonstrations of co-designed robots reach back as far as \citet{park1994concurrent} and have been applied to form factors such as quadrupeds \citep{wang2026efficient}, grippers \citep{xu2024dynamics, Fay2025HouseOD}, full humanoids \citep{wang2026bridge}, and more at large \citep{yu2026agile} and small scales \citep{kriegman2020scalable, balciunaite2026co}.

Despite algorithmic progress, evaluation progress has lagged.
Two precedents have made some progress in this regard.  SoftZoo \citep{wang2023softzoo} provides benchmarking for soft robot co-design algorithms across terrains, materiality, and morphological representations, but lacks physical realism and is limited to 3D elastic dynamics.  EvoGym \citep{Bhatia2022EvolutionGA} provides an extensive array of tasks and environments and emphasizes complex neural control tasks, but it is also limited to soft robots, and in this case, 2D, low-fidelity pixel representations. Further evaluation setups relevant to co-design exist, but either focus on a narrow set of tasks, or lack realistic environment modeling.
Our work is most inspired by the OpenAI Gym \citep{brockman2016openai} and its successor, Gymnasium \citep{Towers2024GymnasiumAS}, which have been seminal for RL research under a shared framework, but lack the extensibility needed to accommodate search over morphologies and other embodiment aspects.

\section{An Optimization Perspective on Co-Design}
\label{sec:problem-statement}

Co-design treats embodiment and policy as coupled choices and a single problem. An embodiment determines aspects of what an agent can do and how it interacts with the environment; a policy determines how those capabilities are used. 
The usefulness of an embodiment therefore depends on the policies available to it, while the usefulness of a policy depends on the embodiment through which it acts. 
For example, changing a robot's leg geometry can change which actuations are effective, so evaluating a new geometry using only a controller developed for the previous one may give a misleading picture of its potential.
Hence, these two problems are not independent.

The term \textit{embodiment} is not limited to a physical body. The capabilities of a virtual character or the interfaces available to a software agent can play a similar role.
In each case, design choices are subject to overarching rules and constraints. Environment structure, tasks, and objectives can also become design variables, but here we focus on embodiment and policy under a shared objective.  It is often useful to select an embodiment before interaction (the beginning of the episode) and hold it fixed during execution when the policy acts.
In such cases, we can  say that the problem of choosing an embodiment (design) and choosing a policy (actions) possess a temporal hierarchy, where an embodiment is chosen once, while actions are chosen multiple times within a single episode.

In order to describe a co-design problem $\mathcal I$, let $\embodyset$ be a nonempty set of admissible embodiments. Each $\embody\in \embodyset$ gives rise to a control problem
$M_{\embody}$ and a nonempty set $\Pi_{\embody}$ of policies available to that embodiment. 
This set incorporates both what the embodiment can execute and any restrictions imposed on the controller. 
We impose no particular form on the policies in $\Pi_{\embody}$, beyond the feasibility and controller restrictions specified by the problem. 
Their action choices may be probabilistic, history-dependent, or time-dependent.
Different embodiments may induce different dynamics, available actions, initial conditions, or information available
to the controller. These details are specified by $M_{\embody}$ and $\Pi_{\embody}$, allowing different embodiments to use different representations.

Let us use $J(\embody,\pi)$ to denote the expected performance of policy $\pi$ with embodiment $\embody$. 
For concreteness, in a discrete-time discounted setting this is
\[
J(\embody,\pi)
=
\mathbb{E}_{M_{\embody},\pi}
\left[
    \sum_{t=0}^{\infty}\gamma^t R_{\embody}(S_t,A_t)
\right],
\qquad 0\leq\gamma<1,
\]
where $R_{\embody}$ is the reward associated with the specified task. 
We assume that the underlying models and policies define this expectation and that it is finite for every pair considered.
Embodiment-dependent rewards can include design costs or other effects as long as their values remain comparable under the same objective.
The argument below also applies to other well-defined scalar performance criteria.

The feasible choices are the embodiment-policy pairs
\[
\mathcal F = \{(\embody,\pi):\embody\in \embodyset,\ \pi\in\Pi_{\embody}\}.
\]
This set is not necessarily a Cartesian product: a policy executable with one embodiment may be unavailable with another. 
It is nevertheless useful to picture performance over these pairs as a joint optimization landscape.
This picture is schematic; neither the embodiment choices nor the policy classes need to form a continuous or finite-dimensional space and often they may not.

\begin{wrapfigure}{r}{0.3\textwidth}
    \centering
    \includegraphics[width=\linewidth]{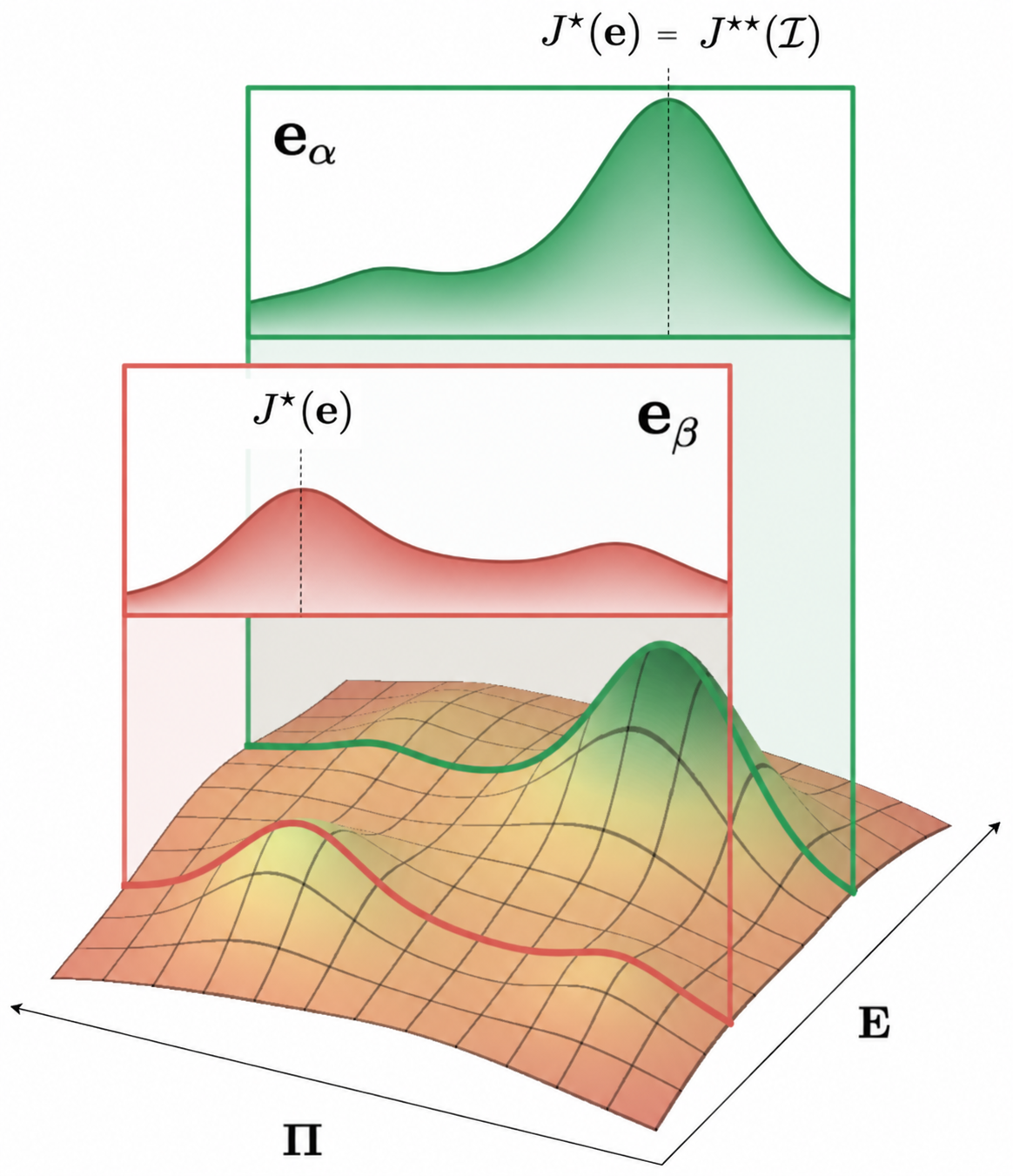}
    \caption{
        An illustrative embodiment-policy optimization landscape.
        The axes $\mathbf{\Pi}$ and $\embodyset$ represent policy and embodiment spaces; surface height represents expected return. 
        The panels $\embody_\alpha$ and $\embody_\beta$ label two fixed-embodiment slices.
        Each $J^\star(\embody)$ refers to optimization over only its panel's embodiments.
        In this example, the green slice contains a joint optimum, giving $J^\star(\embody)=J^{\star\star}(\mathcal I)$.
    }
    \label{fig:problem-formulation}
\end{wrapfigure}

Figure~\ref{fig:problem-formulation} illustrates the interaction between embodiment and policy. 
Each colored slice holds an embodiment fixed while allowing the policy to vary. 
Changing the embodiment changes the slice, potentially changing both the best available return and the policy that achieves it. 
An embodiment's potential therefore depends on the policies it enables.

For a fixed embodiment $\bar{\embody}$, policy optimization searches the corresponding slice. 
Its best achievable value, understood as a supremum, is
\[
J^\star(\bar{\embody})
=
\sup_{\pi\in\Pi_{\bar{\embody}}}
J(\bar{\embody},\pi).
\]
Co-design also searches across embodiments, giving
\[
J^{\star\star}(\mathcal I)
=
\sup_{(\embody,\pi)\in\mathcal F}J(\embody,\pi)
=
\sup_{\embody\in\embodyset}J^\star(\embody).
\]
The nested expression assesses each embodiment through the best performance available with its policies. 
It motivates approaches that optimize policies within a search over embodiments, although co-design algorithms are not restricted to this organization. 
We use suprema because a best policy or embodiment-policy pair need not be attained.

In the illustrated example, the highest point on the red slice lies below the highest point on the green slice, thus only the latter slice contains a joint optimum. 
Policy optimization within the red slice cannot close this difference: doing so requires another embodiment together with a suitable policy. 
The \emph{design gap} is then the difference between a fixed-embodiment performance ceiling and a global optimum. 
When both supremal values are finite, we have
\[
\mathrm{gap}(\bar{\embody};\mathcal I)
=
J^{\star\star}(\mathcal I)-J^\star(\bar{\embody})
\geq 0.
\]
The gap is nonnegative because every pair available with $\bar{\embody}$ remains available in the joint problem. 
A positive gap means that fixing this embodiment excludes some pairs (with alternative embodiments) whose returns exceed its performance ceiling. 
A zero gap means that fixing the given embodiment still allows to attain a global optimum. 
Thus, in the figure, the red embodiment has a positive design gap and the green embodiment has zero gap.

This distinction separates two reasons why a learned policy may fall short of the joint performance ceiling. 
It may not yet use its embodiment effectively, leaving room for improvement within its slice, or the slice itself may also have a lower ceiling than the joint problem.
Improved policy optimization can address the first shortfall, but cannot remove the second while the embodiment, available policy class, and objective remain fixed. 
The design gap captures this remaining opportunity after accounting for the best policies available to the fixed embodiment.

The same reasoning applies to other restrictions of the joint problem.
Restricting embodiments, policies, or both gives a subproblem with a smaller feasible set. 
For any nonempty
$\mathcal F'\subseteq\mathcal F$,
\[
\sup_{(\embody,\pi)\in\mathcal F'}J(\embody,\pi)
\leq
\sup_{(\embody,\pi)\in\mathcal F}J(\embody,\pi),
\]
provided each retained pair is evaluated in exactly the same way.
Every choice in the restricted problem remains available in the larger one. 
If rewards, initial conditions, or other evaluation conditions change for retained pairs, feasible-set inclusion alone no longer establishes this comparison.

These comparisons describe available performance, rather than what a particular algorithm will find. 
A positive design gap motivates searching over embodiments, but the additional search may require more data or computation. 
Under a limited budget, optimizing a policy for a fixed embodiment can still produce a better result than searching jointly over embodiments and policies.

Determining whether a design gap exists also requires looking beyond a single learning curve. 
A plateau at a fixed embodiment establishes neither that its policy has reached the slice's ceiling nor that other embodiments offer no improvement. Assessing the gap requires comparisons across embodiments or additional knowledge connecting their behavior, such as a model of how design affects the dynamics.
Benchmarks with explicit design choices allow these comparisons to be studied. 
Their measured gains, however, reflect both the embodiments considered and how effectively their policies were optimized; they need not equal the theoretic design gap, which compares supremal returns.

There is also an important distinction between this explicit separation of design and control and the generality of MDPs. 
An MDP can itself represent embodiment choices through actions and state variables.
Our formulation exposes those choices as a separate optimization variable so that their interaction with policy can be discussed directly.
In this sense, \textit{policy optimization with an externally fixed embodiment is a restricted case of co-design}, recovered by taking $E=\{\bar e\}$, and can be considered a subproblem in our discussion, as it admits a contained feasible set.
The perspective helps identify which design choices a problem leaves open, which it fixes in advance, and how those decisions shape the policies and performance available to the agent.

While fundamentally a part of a joint problem, embodiment and policy can be viewed as residing in distinct subspaces. This is useful for creating optimizers with different properties for these subspaces, incorporating structural priors, or enabling hierarchy. For instance, sampling from the policy subspace (e.g. realtime action) may be a lot faster than from the embodiment subspace (e.g. making prototypes of novel hardware). The former can benefit from a scalable optimizer that performs best when ample data is available, while the latter requires data-efficient methods. Hierarchical considerations can be accommodated as well. As discussed above, temporal hierarchy can be respected by selecting embodiment first in each optimization trial, then rolling out policy actions. Hence, the above co-design formulation treats optimizing embodiment and policy as a joint problem, while still providing enough structure to accommodate their differences.

Overall, \textsc{Co-Design Gym} provides a unified interface to connect policy and design optimization, offering the broader research community a view on co-design as a general doubly non-stationary optimization problem. 

\newpage

\section{Co-Design Gym}
\label{sec:gym}

\begin{figure}[ht]
    \centering
    \includegraphics[width=1\linewidth]{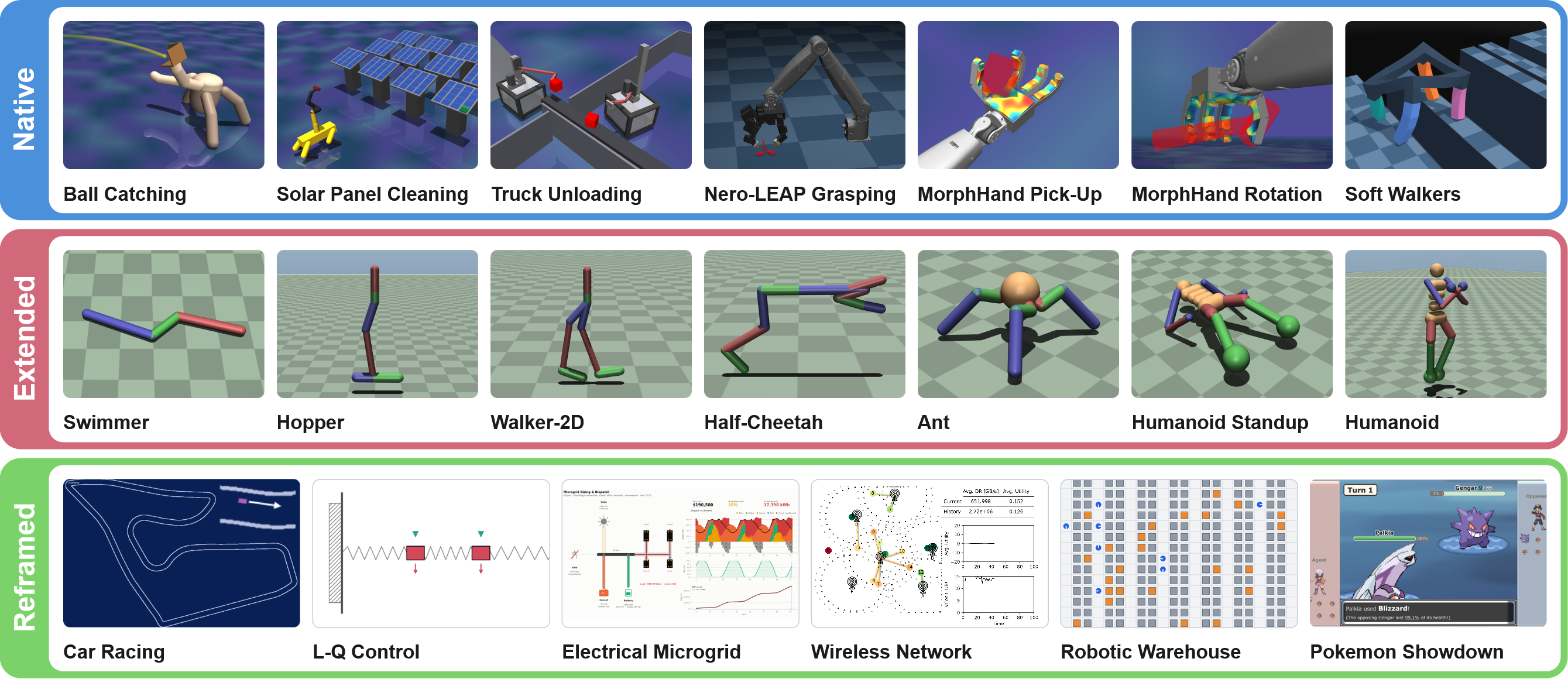}
    \caption{Overview of the available environments. For additional illustrations of the environment presets, refer to \nameref{sec:envs}.}
    \label{fig:env_overview}
\end{figure}

Co-design gym combines environments that span a wide range of domains, including but not limited to robotics, video games, and grid management. Our environments include `from-scratch' setups that natively support efficient and flexible embodiment changes, co-design extensions of the popular and OpenAI MuJoCo Control tasks ported to MuJoCo Warp, and lifts of existing RL environments from various fields as scalable co-design problems. 

Each environment pairs an inner policy optimization loop with an outer embodiment problem, and ships with several presets of varying complexity. All rigid-body robotics tasks are built in MuJoCo~\citep{todorov_mujoco_2012} and MuJoCo Warp~\citep{google-deepmind_mujoco_warp_2025, macklin_warp_2022}. Most of our environments are natively GPU-accelerated. Below is a brief summary of the main environment families. For full descriptions, reward definitions, and additional presets, please refer to Appendix~\ref{sec:envs}.

\textbf{Ball Catching.} A robot must catch a thrown ball in a glove while remaining upright. The policy controls the robot, and the embodiment varies its limbs and torso. Three presets modify the throw: fast and flat, high arc, and varied 3D fielding: testing co-design under dynamic conditions inspired by \citet{Morgan_2026}.

\begin{wrapfigure}{r}{0.73\textwidth}
 \begin{center}
 \vspace{-1.75em}
     \includegraphics[width=0.73\textwidth]{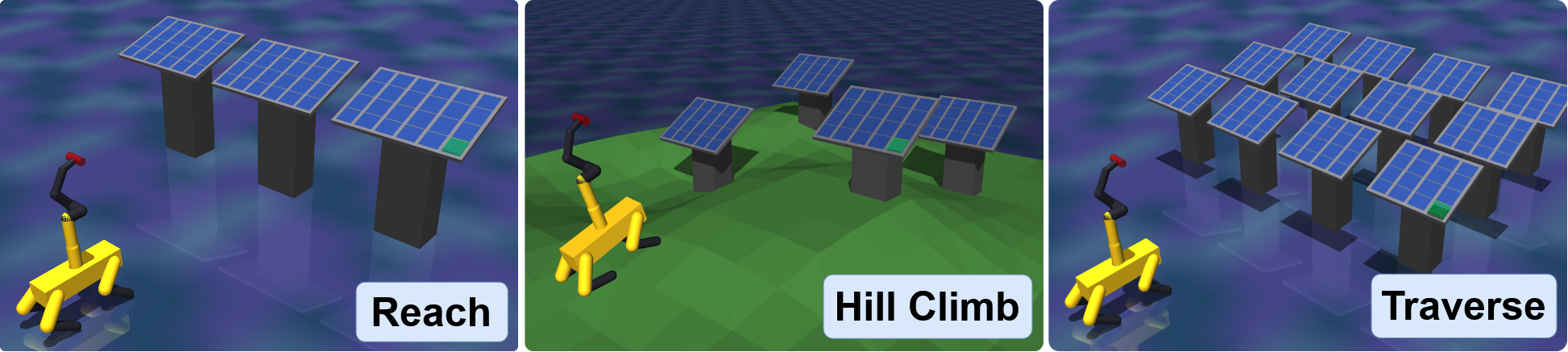}
     \label{fig:envs:solarcleaner}
     \vspace{-1.5em}
     \caption{Solar Panel Cleaning environment.}
     \vspace{-1.5em}
 \end{center}
\end{wrapfigure}
\textbf{Solar Panel Cleaning.} A quadruped robot with a roller-tipped arm must clean solar panels without any other body part touching them. The policy controls body and arm, and the embodiment modifies all limbs. Presets cover a variety of solar panel layouts and terrains: targeting practical deployments of co-designed locomoting robots for intricate manipulation scenarios.

\textbf{Truck Unloading.} A mobile manipulator unloads boxes from a truck onto a conveyor, where another sorts and shelves or stacks them by colour. A single policy controls all robots, and the embodiment modifies the arms of each robot. Presets vary the number of unloaders and different box colours: presenting a coordination challenge at the embodiment and control levels.

\textbf{Nero-LEAP Grasping.} A LEAP hand on a Nero arm must stably lift small tools and hardware from a table. The policy controls the hand, and the embodiment replaces the thumb and index fingertip meshes. Presets cover different groups of objects: testing high-dimensional mesh-based shape design coupled with fine, contact-rich interactions.

\newpage

\begin{wrapfigure}{r}{0.65\textwidth}
  \begin{center}
      \includegraphics[width=0.65\textwidth]{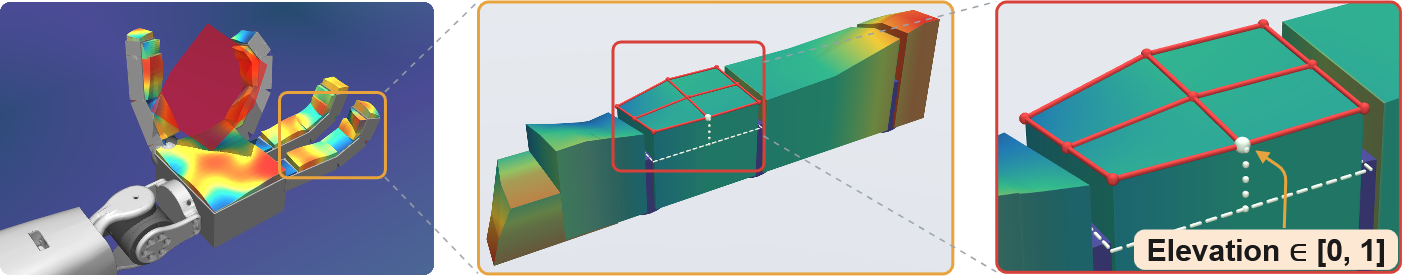}
      \label{fig:envs:rotate_in_hand}
      \vspace{-1.5em}
      \caption{The MorphHand. Colors show elevation at each vertex.}
      \vspace{-1.5em}
  \end{center}
  \end{wrapfigure}
\textbf{MorphHand Pick-Up \& Rotation.} A four-fingered hand must either reorient an object in-hand or pick up and pose a cube, ball, or hammer. Supporting transition-level embodiment changes, these environments enable the use of a new class of design optimization algorithms for  complex single-hand manipulation.

\begin{wrapfigure}{r}{0.38\textwidth}
\vspace{-1.5em}
\begin{center}
    \includegraphics[width=0.38\textwidth]{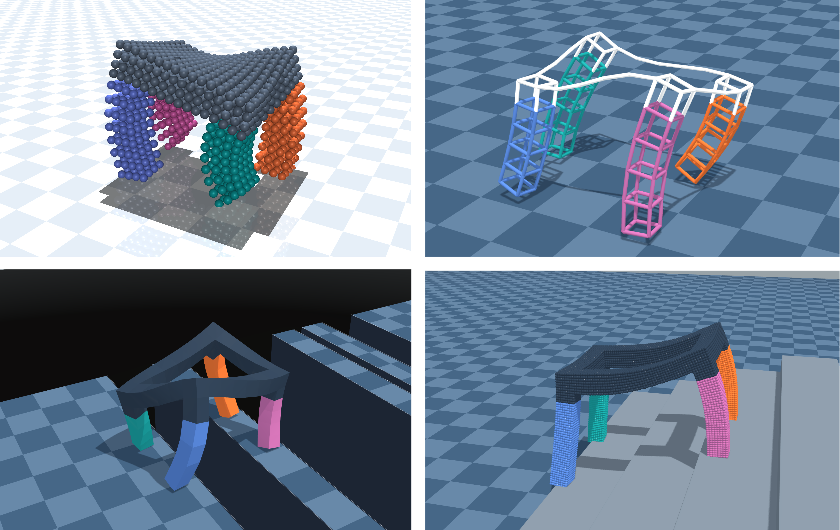}
    \label{fig:envs:softrobot}
    \vspace{-1.5em}
    \caption{Soft Walkers environment.}
    \vspace{-1.5em}
\end{center}
\end{wrapfigure}
\textbf{Soft Walkers.} Following \citet{Bhatia2022EvolutionGA} and \citet{cheney2014unshackling}, a soft robot must traverse flat, stepped, or gapped terrain. Each terrain is paired with five 2D/3D spring, MPM, beam, and FEM solvers (15 presets). Co-designing soft robots presents a unique challenge due to their deformable bodies and the large and varied design space spanning shape, mass, material, and actuator properties.

\textbf{Car Racing.} A scaled F1Tenth car~\citep{okelly2020f1tenth} must minimise lap time on the Levine or Spielberg track. The policy drives the car. The embodiment sets the vehicle setup, such as tyre grip, cornering stiffness, and drive power, and each parameter trades off against the others. 

\textbf{Linear-Quadratic Control.} A mass-spring-damper chain must be regulated against disturbances. The embodiment includes the masses, stiffnesses, damping, and actuator and sensor placement. Three presets cover continuous design, combinatorial placement under noise, and budget-constrained design with saturated inputs, so the environment serves as a lightweight sanity check for co-design algorithms.

\textbf{Electrical Microgrid.} A microgrid must meet the randomised demand of one or more buildings over 24--72 hours, either grid-connected or off-grid~\citep{Henri2020pymgridAO}. The policy dispatches energy, and the embodiment sizes solar, battery, and diesel capacity. Design costs are charged in the reward, making this a budgeted problem. A variety of presets test the idiosyncratic problem of grid control with careful resource allocation.

\textbf{Wireless Network.} Base stations must give strong, consistent coverage to moving mobile users~\citep{schneider2022mobileenv}. The policy assigns users to stations, and the embodiment includes the number, placement, and transmit power of the stations. Presets range from slow urban pedestrians to sparse, high-speed motorway traffic: combining the challenges of control in networking with geographic layouts and resource allocation.

\textbf{Robotic Warehouse.} A robot fleet fetches requested shelves and delivers them to goals, with the aim of maximising throughput~\citep{bonnet2024jumanji, semitable_multi_robot_warehouse_rware_2020}. The policy controls the fleet, and the embodiment lays out shelves on a fixed floor, trading route length against congestion. Presets scale from 2 to 8 robots: representing the challenges of finding optimal layout patterns for fleets of uniform agents.

\textbf{Pokemon Showdown.} The agent must win single battles, or a gauntlet of five League opponents, in Pokemon Showdown~\citep{smogon_pokemon_showdown_2011, hsahovic_poke_env_2019}. The policy chooses moves, and the embodiment is the six-Pokemon team. Opponents range from random players to the League teams of all nine game generations, with or without healing between battles. With a combinatorially explosive discrete design space, and intricate environment dynamics, this tests sparse design space navigation in addition to complex environment rule-learning.

\subsection{API}

To facilitate ease of use we keep API changes minimal relative to the widely used Gymnasium \citep{Towers2024GymnasiumAS} API. That way,  existing control implementations remain usable and co-design methods differ only where necessary.

Any Gymnasium environment becomes a co-design environment by declaring an
\texttt{embodiment\_space}, a map from
$\mathbf{e} \in \mathbf{E}$ to a realised body, and by forwarding the new argument through \texttt{reset}; the dynamics, reward, and termination logic are untouched.
The original environment is recovered exactly as the base embodiment
$\mathbf{e}_0$, so the conversion adds designs without altering the original task.

The \texttt{embodiment\_space} is a Gymnasium \texttt{Space} representing
$\mathbf{E}$ (see Section~\ref{sec:problem-statement}), alongside the usual \texttt{observation\_space} and \texttt{action\_space}. It supports
\texttt{sample()} and \texttt{contains()} with the standard semantics, so a design space can be
enumerated, sampled from, or checked for membership by the same machinery used for actions.

Each environment declares a \emph{base embodiment} $\mathbf{e}_0 \in \mathbf{E}$: the conventionally chosen body, with which it is realised until some other design is requested. Passing \texttt{None} retains the embodiment currently in force, which for an environment never given an explicit one is $\mathbf{e}_0$ throughout; passing $\mathbf{e} \in \mathbf{E}$ requests that the agent be realised as $\mathbf{e}$. An environment called with \texttt{embodiment=None} everywhere is therefore a standard Gymnasium environment with embodiment $\mathbf{e}_0$. 

\begin{wrapfigure}{r}{0.6\textwidth}
\vspace{-1.75em}
\includegraphics[width=0.6\textwidth]{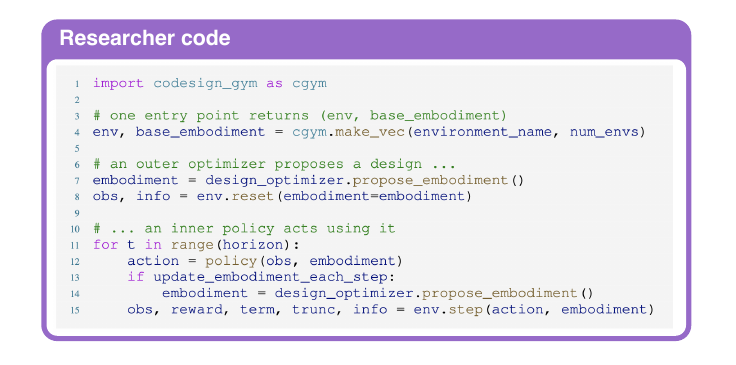}
\caption{Minimal usage of \texttt{codesign\_gym}. A single entry point,
\texttt{make\_vec}, returns a natively parallelized environment together with its base
embodiment supporting GPU-acceleration. An outer design optimizer proposes an embodiment that is passed to
\texttt{reset}, and an inner policy then acts against it. The embodiment can
optionally be re-proposed and passed to \texttt{step} at every timestep.}
\label{fig:codesign-gym-api}
\end{wrapfigure}
\paragraph{Within-task modifications}
Our API admits the embodiment argument in \texttt{step} as well as \texttt{reset}. This choice allows the embodiment to change during
an episode. This is not physically meaningful for many tasks---bodies do not deform on command mid-trajectory---but it is potentially useful during training, where it permits a single rollout to experiment with multiple designs. Because within-episode modification is physically infeasible for many tasks, results obtained with it are not directly interpretable as task performance. A method that exploits it during evaluation would report returns no physical system could achieve. Each environment therefore carries an \texttt{is\_eval\_env} flag, fixed at construction, \texttt{env = CoDesignEnv(..., is\_eval\_env=True)}. If true, changes that alter the agent's embodiment take effect only when supplied to
\texttt{reset}.

\section{Experiments with Representative Optimization Methods}
\label{sec:expr}

To demonstrate our environments, we benchmark four co-design algorithms on them. Existing research in co-design has not had a unified interface/setup and therefore makes state-of-the-art algorithms difficult to compare. Our selection of baselines represents different frontiers of co-design algorithms from evolutionary approaches and reinforcement learning (RL) to transformer-based search and hybrid approaches.

For all baselines, every 5\% of training the current best embodiment and controller are evaluated in parallel environments on held-out seeds.

\textbf{CMA-ES.} CMA-ES searches over controller and embodiment parameters jointly as a single vector. The controller parameters are the weights of a neural-network policy.

\textbf{FastTD3~\citep{seo_fasttd3_2025}.} We add the embodiment parameters to the action space of FastTD3, an off-policy RL algorithm for continuous control. The embodiment actions are applied at the first step of each rollout and the control actions at every later step. Unused action components are zeroed in the replay buffer, so the policy learns only from actions that were actually taken.

\textbf{PPO+NGOpt~\citep{Schulman2017ProximalPO, nevergrad}} PPO trains the controller. After each PPO update, NGOpt proposes one embodiment per training environment and scores it by its mean return in the rollouts where it was used.

\textbf{LOKI.} We generalise \citet{jeon2026convergent} to arbitrary embodiment parametrisations by replacing its robot-centric tokenization with a naive one. We pre-train its Transformer VAE on 500k randomly sampled valid embodiments. We then run the original online loop with PPO-trained Transformer controllers, using 6 clusters instead of 40 for comparability with the other baselines. To test this reduction we run a 40-cluster ablation on LOKI's best- and worst-scoring presets and show that neither result changes materially (Appendix~\ref{sec:loki40}). At evaluation, the controller of the best cluster is paired with that cluster's top embodiments.

\textbf{LA-MCTS~\citep{wang2020lamcts} and SPSA~\citep{spall1992spsa}.} To broaden the range of optimizer families, we additionally run two black-box optimizers over the same joint vector of embodiment parameters and controller weights as CMA-ES: LA-MCTS, a Monte Carlo tree search that learns to partition the search space, and SPSA, a stochastic finite-difference gradient method. Together, the six baselines span evolutionary, tree-search, stochastic-gradient, reinforcement-learning, alternating, and generative approaches. We run LA-MCTS and SPSA on 20 presets and report them in Appendix~\ref{sec:extra-baselines}.

For a detailed description of our baselines, we refer to Appendix~\ref{sec:baselines}.

We evaluate each of these methods on our co-design gym tasks, reporting performance across evaluation environments, embodiment rejection rates, and training behaviour. Each baseline and environment are trained under matched hardware conditions with 4-hour wall-clock time budgets other than LOKI. For comparability, we perform online training for LOKI for 3-hours, as an average of 1-hour was consumed in data generation, VAE pre-training, and cluster generation. Results are reported across five seeds on the 35 presets for which all four baselines were run.

\textbf{Normalized score.} As raw returns differ by orders of magnitude across presets, we follow \citet{Mnih2013PlayingAW} and report $100 \times (R - R_{\text{rand}}) / (R_{\text{ref}} - R_{\text{rand}})$, where $R$ is the final evaluation return, $R_{\text{rand}}$ that of a uniform-random policy on the base embodiment, and $R_{\text{ref}}$, in the absence of a human reference, the theoretical maximum return or, where it is unbounded, a physics-based cap (Appendix~\ref{sec:normalization}). Raw returns, anchors, environment interactions and model sizes are in Appendix~\ref{sec:detailed-results}.

\begin{table}[t]
\centering
\footnotesize
\setlength{\tabcolsep}{1.5pt}
\begin{tabular}[t]{@{}lrrrr@{}}
\toprule
Preset & \shortstack[r]{CMA-\\ES} & \shortstack[r]{Fast\\TD3} & \shortstack[r]{PPO+\\NGOpt} & LOKI \\
\midrule
\multicolumn{5}{@{}l}{\textit{Native}} \\
BallCatcher-Pitch & 2.1 & \textbf{33.6} & 3.6 & $-$0.2 \\
BallCatcher-Arc & 2.4 & \textbf{41.0} & 15.2 & 0.1 \\
BallCatcher-Fielding & 0.5 & \textbf{30.9} & 3.8 & $-$0.2 \\
NeroGraspAll & \textbf{9.8} & $-$1081.9 & 0.3 & 0.0 \\
HandPickUpCube & \textbf{1.7} & 1.2 & 0.8 & 0.6 \\
HandPickUpHammer & \textbf{27.5} & 12.1 & 9.3 & 4.9 \\
RotateInHandCube & 1.2 & \textbf{1.6} & 1.6 & 0.4 \\
SoftWalkerBeam3D & 4.4 & 4.4 & \textbf{4.5} & 2.5 \\
SolarCleanerReach & 0.3 & 0.1 & \textbf{1.9} & 0.1 \\
SolarCleanerTraverse & \textbf{0.2} & 0.0 & 0.0 & 0.1 \\
TruckUnloadSingle & 8.0 & \textbf{22.3} & 7.5 & 4.7 \\
\midrule
\multicolumn{5}{@{}l}{\textit{Extended}} \\
HalfCheetah & \textbf{36.1} & 15.6 & 22.0 & 16.0 \\
Hopper & 7.9 & 5.6 & 9.9 & \textbf{10.0} \\
Walker2d & \textbf{16.0} & 14.4 & 15.2 & 6.7 \\
Swimmer & \textbf{11.9} & 2.4 & 2.0 & 1.5 \\
Ant & 18.0 & 17.1 & \textbf{22.7} & 0.4 \\
Humanoid & 2.9 & \textbf{16.4} & 2.6 & 1.2 \\
HumanoidStandup & \textbf{31.4} & 6.5 & 15.3 & 7.5 \\
\bottomrule
\end{tabular}
\hfill
\begin{tabular}[t]{@{}lrrrr@{}}
\toprule
Preset & \shortstack[r]{CMA-\\ES} & \shortstack[r]{Fast\\TD3} & \shortstack[r]{PPO+\\NGOpt} & LOKI \\
\midrule
\multicolumn{5}{@{}l}{\textit{Reframed}} \\
\multicolumn{5}{@{}l}{Racing} \\
\hspace{0.8em}Spielberg & 1.5 & \textbf{21.9} & 18.2 & 10.7 \\
\hspace{0.8em}Levine & 7.3 & \textbf{74.6} & 60.9 & 25.3 \\
\multicolumn{5}{@{}l}{Warehouse} \\
\hspace{0.8em}Small & $-$0.0 & $-$0.0 & \textbf{1.3} & 0.1 \\
\hspace{0.8em}Medium & 0.0 & 0.0 & \textbf{0.0} & 0.0 \\
\hspace{0.8em}Congested & $-$0.0 & $-$5.0 & \textbf{0.0} & $-$0.0 \\
\multicolumn{5}{@{}l}{Network} \\
\hspace{0.8em}Urban & $-$7.6 & $-$15.0 & $-$4.3 & \textbf{$-$1.3} \\
\hspace{0.8em}Suburban & $-$20.7 & $-$100.1 & \textbf{$-$11.5} & $-$35.8 \\
\hspace{0.8em}Motorway & 28.2 & $-$117.2 & \textbf{49.4} & 36.5 \\
\multicolumn{5}{@{}l}{Microgrid} \\
\hspace{0.8em}Random & 93.9 & \textbf{96.3} & 62.1 & 52.1 \\
\hspace{0.8em}Forecast & 89.5 & \textbf{96.6} & 57.5 & 51.5 \\
\hspace{0.8em}Campus & 92.1 & \textbf{96.0} & 55.5 & 51.9 \\
\hspace{0.8em}OffGridSingle & 76.0 & \textbf{79.8} & $-$18.1 & 71.1 \\
\hspace{0.8em}OffGridCampus & \textbf{85.4} & 68.9 & 0.5 & 71.0 \\
\multicolumn{5}{@{}l}{Pokenv} \\
\hspace{0.8em}Random & 26.0 & 38.8 & 47.0 & \textbf{49.8} \\
\hspace{0.8em}MaxPower & 30.2 & 24.9 & 33.2 & \textbf{41.3} \\
\hspace{0.8em}Heuristic & 5.9 & 18.3 & \textbf{29.2} & 18.3 \\
LQStructure (LQR) & 82.3 & \textbf{91.0} & 86.9 & 81.8 \\
\midrule
Best on & 9 & 13 & 9 & 4 \\
IQM & 10.5 & 13.5 & 9.1 & 7.0 \\
\bottomrule
\end{tabular}

\caption{Main scoreboard with normalized evaluation scores (0 = uniform-random policy, 100 = theoretical maximum return or physics-based cap), mean over 5 seeds. Bold indicates best mean. 
Each baseline was given a 4-hour wall-clock budget (LOKI: 3 hours online after pretaining 1 hour).
Reporting runs for 35 presets: 4 baselines, 4 hours and 5 seeds each, for a total of 2,800 hours.
35/85 presets were selected as follows: one preset from each environment family to ensure variety; beyond that: all BallCatcher presets (to test locomotion and manipulation, with sparser rewards than locomotion-only OpenAI Gym extensions), additional Solar Cleaner presets (to test loco-manipulation with a robot based on the Boston Dynamics Spot); 
all MorphHand presets (to test rapid per-step shape updates); all presets for reframed families (except LQR being the simplest was limited to  1 preset, and we also skipped pokemon presets with expert-curated strategies, since those go much beyond baselines).
IQM: interquartile mean of the normalized score over all runs. FastTD3 on NeroGraspAll: every proposed design failed structural validation and received the rejection reward ($-250$). Raw returns are in Table~\ref{table:full-results-raw}.}
\label{table:baselines}
\vspace{-10px}
\end{table}

\begin{figure}[h!]
  \centering
  \includegraphics[width=\linewidth]{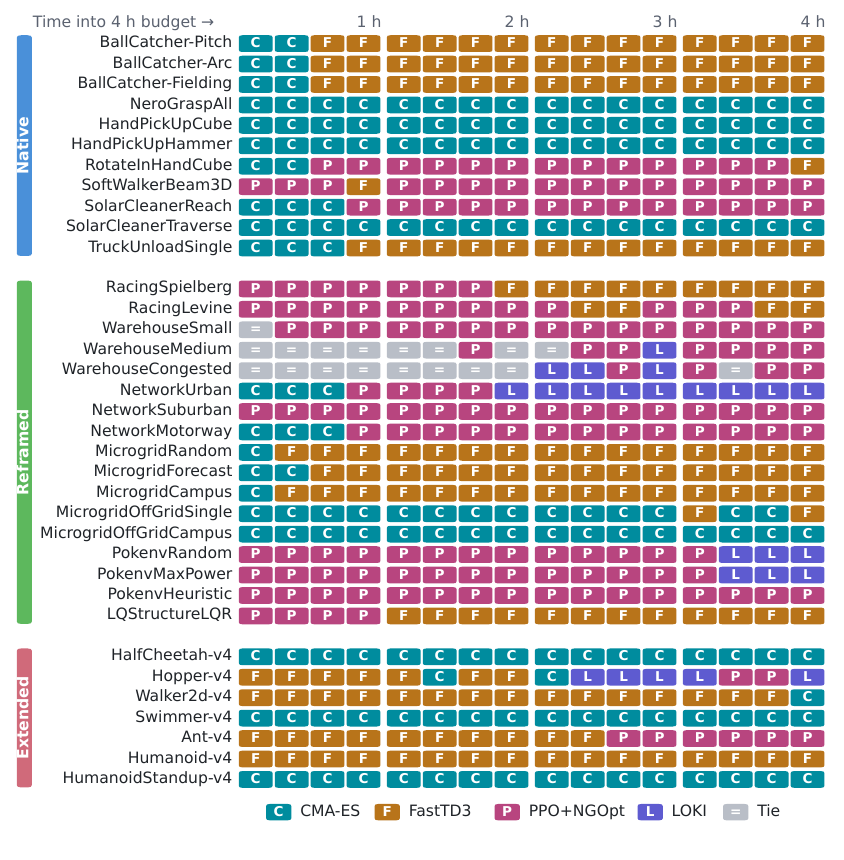}
  \caption{\textbf{Leading baseline over training} on the 35 presets run with all four baselines. Each cell names the method with the
  highest mean eval return (5 seeds) at that point, using the latest evaluation so far; LOKI's evaluations start after its 1-hour preparation.
  CMA-ES leads at the first checkpoint on 20 presets but at the end on only 9; FastTD3 ends ahead on 13, PPO+NGOpt on 9 and LOKI on 4.
  The leader changes at least once on 24 of 35 presets, and no method leads on every domain group, so the ranking of co-design baselines
  depends on both the task and the budget.}
  \label{fig:leader-loki}
\end{figure}

\textbf{The benefit of co-optimization.} Two controls under the same budget isolate policy-only optimization (PPO on the fixed base embodiment) and design-only optimization (NGOpt under a frozen random policy). Design search alone never matches the best co-design baseline, yet fixed-design PPO outperforms every co-design baseline on 9 of 22 presets (Appendix~\ref{sec:controls}).

\textbf{Exploratory Study: LLM Co-Design.} Large language models (LLMs) have shown strong capabilities across a wide range of domains~\citep{Minaee2024LargeLM}, which raises the question of whether they can act as co-design optimizers. We examine this in an exploratory study in which an LLM (Gemma 4 31B) proposes robot embodiments in the outer loop while PPO learns the controller in the inner loop. Because this setup mirrors PPO+NGOpt with the LLM in place of NGOpt, it serves as an ablation of the outer-loop design optimizer. Interestingly, PPO+LLM stays near zero return throughout training, even though its prompt provides additional task-relevant information to the model. We refer to Appendix~\ref{appx:llm-ppo} for a detailed description.

\section{Discussion and Conclusion}
\label{sec:discussion}

\textbf{No single dominating co-design method.} Across the 35 presets on which all four baselines were run, no method leads: FastTD3 is best on 13 presets, CMA-ES and PPO+NGOpt on 9 each and LOKI on 4 (Table~\ref{table:baselines}), and the emerged ranking depends on both the task and the budget (Figure~\ref{fig:leader-loki}). The final winners also cluster by domain. CMA-ES performs best on most of the extended locomotion and grasping tasks, FastTD3 on the ball-catching, racing and most microgrid presets, PPO+NGOpt on the warehouse and most wireless presets, and LOKI on most Pok\'emon presets. Across the variety of domains represented in our benchmark, none of the evaluated co-design methods can claim general capability. This is made visible by the presentation of our unified benchmark; and fragmented evaluation settings of prior work would not reveal this.

\textbf{Substantial headroom remains.} Many presets are far from solved on which no co-design algorithm approached the return ceiling: the best method's normalized score is below 10\% on 11 of 35 presets, and its median across presets is 22.7\%. For instance, on BallCatcher-Pitch the best method completes only 16.6\% of catches. Additionally, all methods score near zero on more challenging benchmarks such as SolarCleaner, the hand-manipulation cube presets, WarehouseMedium and WarehouseCongested. The benchmark leaves ample room for methods that explore the co-optimization of embodiment and policy more efficiently and thoroughly than current approaches across continuous and discrete tasks. We hope that this encourages exciting new development of co-design algorithms to push the frontier on many fun, practical, and challenging domains.

\textbf{Limitations and future work.} For many proposed environments, the design space does not encompass all conceivable degrees of freedom; further extensions are possible, leaving room for richer parametrizations. Moreover, due to the breadth of the proposed benchmark and the generality of our underlying simulators, our baselines cover 35 of the over 85 available presets. Furthermore, the fixed wall-clock budget in our experiments favors faster implementations, so part of the performance gap between baselines may reflect engineering rather than algorithmic differences. Our findings characterize the evaluated methods rather than all co-design techniques, and the native environments abstract actuator internals and materials, leaving hardware transfer to future work. Finally, achieving high throughput with embodiment changes in parallelized, GPU-accelerated physics simulation represents a substantial amount of work in structuring the compilation, caching, and workload scheduling. Incorporating more complex phenomena, such as fluids, wind, or smoke, may require integrating new simulators, with comparably large engineering effort.

\textbf{Conclusion.} We presented \textsc{Co-Design Gym}, a unified benchmark for the joint optimization of embodiment and policy. 
In our work we formalized co-design problems and solutions focusing on pairs of design spaces and policy classes, and situated standard RL as a special case of co-design with a single admissible design. We realized this formalism as an extension of the Gymnasium API, and used it to build on a large variety of widely used benchmark environments as novel co-design problems. In addition, we built a number of new environments from scratch specifically to stress-test the interplay between the embodiment and policy optimization, native and targeted to co-design. Finally, our experiments tested representative co-design methods and analyzed their performance.
The appendix complements these results with full environment specifications and additional presets (Appendix A), baseline implementation details (Appendix B), detailed results including raw returns, normalization anchors, controls, two additional baselines, compute statistics, per-preset learning curves and final designs (Appendix C), an exploratory study in which an LLM given privileged task knowledge fails as an outer-loop designer (Appendix D), and a detailed illustration of the API (Appendix E).
Unified benchmarks helped carry RL from struggling on basic tasks to superhuman performance. We hope Co-Design Gym can play the same role for co-design, advancing towards its broader promise: optimizing every dimension of a problem jointly to solve it as effectively as possible.

\section*{Acknowledgements and Funding Statement}

The authors would like to acknowledge and thank Austin Yang, Hantao Zhong, Andy Zhou, Allan Zhao, and Ferran Hernandez Caralt for their insightful discussions. We would further like to thank Taylor Whatley and Ruth Lu for assistance with reviewing code and preliminary testing.

This project is funded by the Advanced Research + Invention Agency (ARIA) as part of its Robot Dexterity programme under grant numbers SMRB-P01-P20 and SMRB-P01-P14.

\appendix
\refstepcounter{section}
\section*{Appendix \thesection: Environments}\label{sec:envs}

Unless stated otherwise, the tasks described below use MuJoCo~\citep{todorov_mujoco_2012} and MuJoCo Warp~\citep{google-deepmind_mujoco_warp_2025, macklin_warp_2022} as the physics engine.

\begin{table}[htbp]
\centering
\footnotesize
\setlength{\tabcolsep}{3pt}
\renewcommand{\arraystretch}{1.0}
\begin{tabularx}{\linewidth}{@{}l c >{\raggedright\arraybackslash}X >{\raggedright\arraybackslash}X >{\raggedright\arraybackslash}p{2.2cm} >{\raggedright\arraybackslash}p{2.0cm}@{}}
\toprule
Family & Presets & Design space & A design is rejected if & Return per episode & Episode \\
\midrule
\multicolumn{6}{@{}l}{\textit{Native}} \\
BallCatcher & 3 & 45 continuous link offsets in $[0,1]$ & out of bounds or non-finite & $[0, 10]$ & $\le$800 steps of 0.05\,s \\
SolarCleaner & 3 & 48 continuous leg and arm link offsets in $[0,1]$ & out of bounds or non-finite & $\le 10$ & per-goal step budgets, 0.04\,s steps \\
TruckUnload & 2 & 6 / 9 continuous column and arm lengths per robot & out of bounds or non-finite & $\le 10$ & 3{,}000 / 7{,}500 steps of 0.04\,s \\
NeroGrasp & 13 & two triangle meshes (index and thumb fingertips) & invalid topology, volume below 90\% of the base, wall thinner than 0.3\,mm, or too small & $\ge 0$; at most 23.2 if the object stays below 0.2\,m & 100 steps of 0.1\,s \\
RotateInHand & 1 & 245 pad heights in $[0,1]$ and 3 finger offsets in $[-1,1]$ & out of bounds or non-finite & up to 10 per goal reached & $\le$375 steps of 0.04\,s \\
HandPickUp & 3 & as RotateInHand (248 values) & out of bounds or non-finite & Cube: up to 9 per goal reached, plus 1; Hammers: $\le 10$ & $\le$375 / 175 steps of 0.04\,s \\
SoftWalker & 15 & material and actuation of each cell: 432 (2D) or 2{,}880 (3D) mixed continuous and integer values & out of bounds, empty or disconnected body, or actuator without a fibre & forward progress minus actuation work & 240 steps of 0.01\,s \\
\midrule
\multicolumn{6}{@{}l}{\textit{Reframed}} \\
LQStructure & 3 & 18 continuous masses, stiffnesses and dampings; 12 binary actuator and sensor placements & no actuator; LQG: no sensor; QP: total mass above 8.25 or more than 3 actuators & $\le 0$ & 200 steps of 0.05\,s \\
Microgrid & 5 & 3 continuous capacities (solar, battery, generator) & out of bounds or non-finite & $\le 720$ / 2{,}880 (grid-connected), $\le 0$ (off-grid) & 24 / 72 steps of 1\,h \\
Network & 3 & number of base stations (integer) and each one's position and power (continuous) & station count, position or power out of range, or total power above the budget & $[-100, 100]$ & 100 steps \\
Racing & 2 & 5 continuous: tyre grip and stiffness, drive force, weight distribution & out of bounds or non-finite & distance along the track, at most one lap & 1{,}500 / 1{,}000 steps of 0.01\,s \\
Warehouse & 3 & 98 / 254 / 438 binary shelf placements & too few shelves, or too few free cells for the robots & $[0, 1000]$ & 500 steps \\
Pok\'emon & 23 & team composition: categorical and integer, up to 258 values & rejected by the Showdown team validator & battle: $[-28, 28]$; League: up to $+140$ & one battle, or up to five \\
\midrule
\multicolumn{6}{@{}l}{\textit{Extended}} \\
MuJoCo ports & 7 & 6--36 continuous link offsets in $[-1,1]$ & out of bounds or non-finite & unbounded & 1{,}000 steps \\
\bottomrule
\end{tabularx}
\caption{Specification of every environment family (a full version is in the readthedocs). A rejected design is reported at \texttt{reset} and the copy cannot be stepped until it is reset with a valid design. Evaluation uses held-out environment seeds and refuses design changes during an episode; Racing evaluation episodes also run for up to 10{,}000 steps. Normalization anchors are listed in Table~\ref{table:anchors}.}
\label{table:env-spec}
\end{table}

\begin{wrapfigure}{r}{0.45\textwidth}
\vspace{-1.75em}
\begin{center}
    \includegraphics[width=0.45\textwidth]{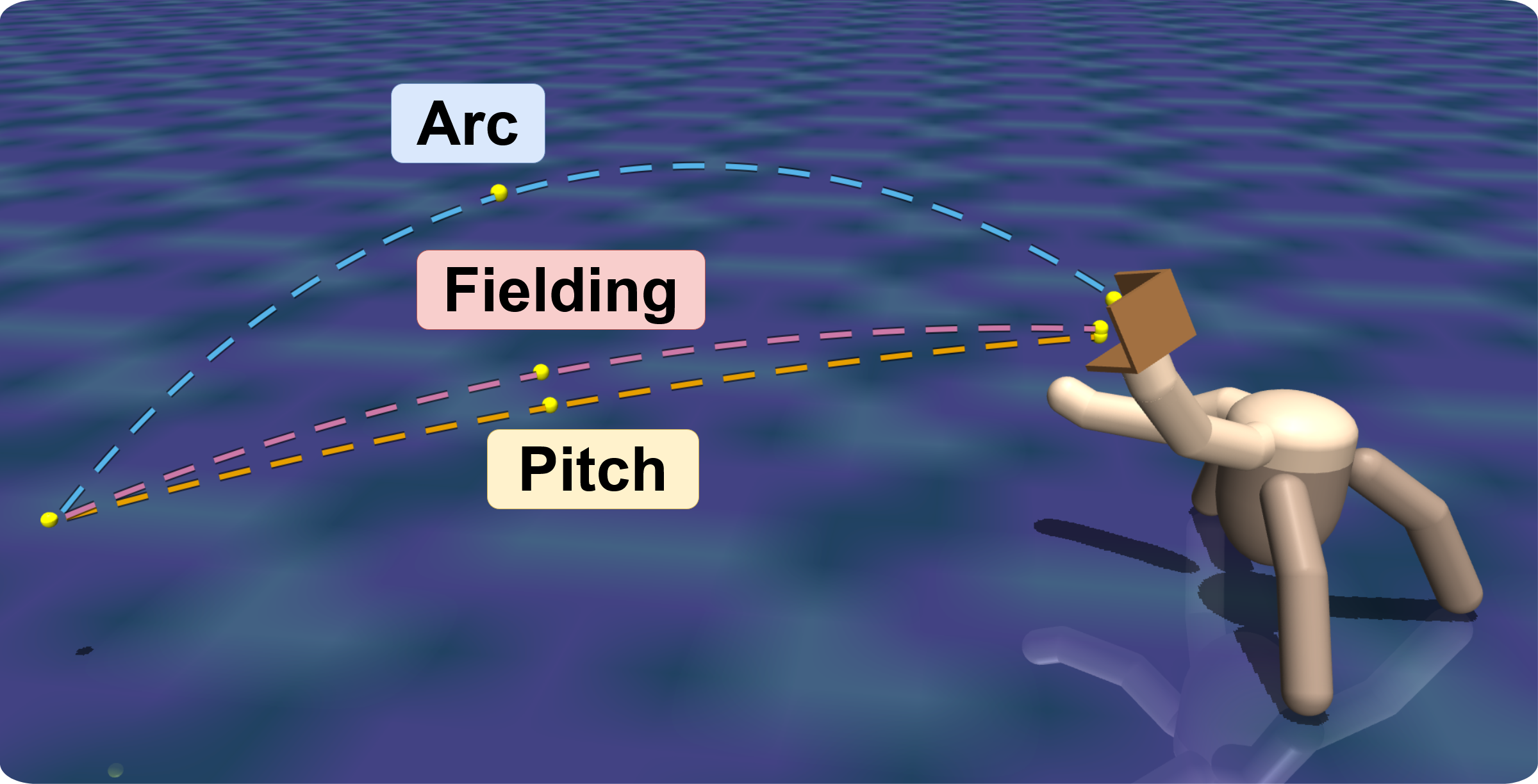}
    \label{fig:envs:ballcatcher}
    \vspace{-1.7em}
    \caption{Ball Catching environment.}
    \vspace{-1.9em}
\end{center}
\end{wrapfigure}
\textbf{Ball Catching.} In this environment a robot inspired by Rocky from Project Hail Mary~\citep{weir2021projecthailmary, lordmiller2026projecthailmary} is tasked with catching a ball that is thrown towards it. At the end of the arm a "glove" is attached, in which the ball must be caught. The source of the ball is fixed but the target and speed are randomised. We provide three presets. 1) BallCatcherPitch in which the ball is thrown fast, flat, and low. 2) BallCatcherArc in which the ball is thrown in a slower, high arc descending towards the quadruped. 3) BallCatcherFielding in which the ball is thrown with varied speed and direction toward a broad 3D catch region requiring versatility. In these tasks, the control optimization loop is training a policy for the arm-equipped quadruped, and the design problem is the configuration of the arm. Partial rewards are obtained for catching and dropping the ball, but full reward (10) is accomplished when catching, retaining the ball, and with all three legs touching the ground. This final condition is required to prevent unstable jumping policies.

\begin{wrapfigure}{r}{0.73\textwidth}
\vspace{-10px}
\vspace{-0.5em}
\begin{center}
    \includegraphics[width=0.73\textwidth]{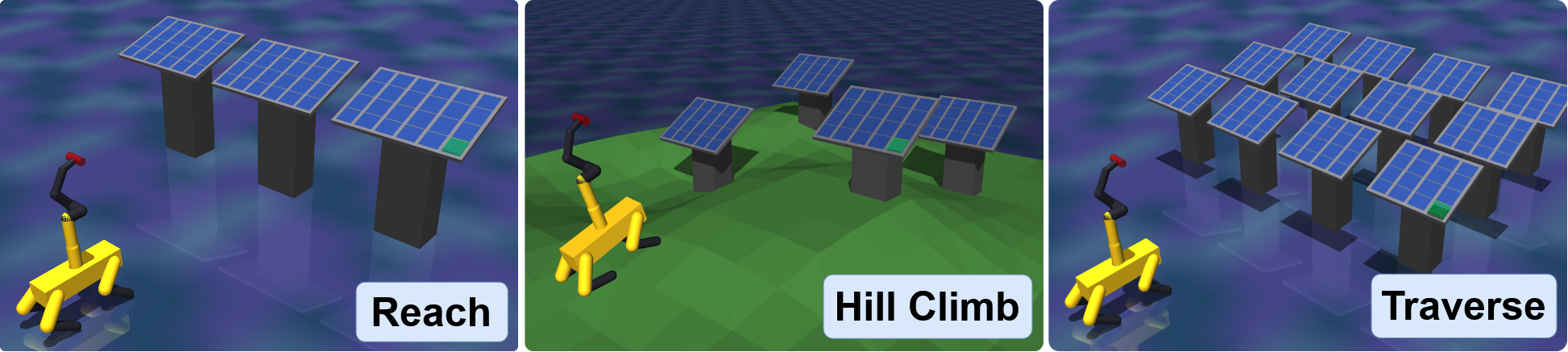}
    \vspace{-1.5em}
    \caption{Solar Panel Cleaning environment.}
    \vspace{-1.5em}
\end{center}
\end{wrapfigure}
\textbf{Solar Panel Cleaning.} In this environment a quadruped is equipped with an arm and tasked with carefully cleaning an assortment of solar panels. At the end of the arm is a small roller which cleans a solar panel cell when in contact. Since panels are delicate, no other part of the robot may come in contact with the panel cells. We provide three presets. 1) SolarCleanerReach in which three panels are mounted higher up than the base configuration of the robot and laid out in a line. 2) SolarCleanerHillClimb in which four panels are distributed across a rocky and uneven terrain. 3) SolarCleanerTraverse in which a 3x4 grid of solar panels is tightly aligned forming narrow aisles for robot traversal. In these tasks, the control optimization loop is training a policy to control the Ant robot and arm. The design problem is to modify the legs and the arms of the robot to enable stable and careful locomotion alongside precise arm control.

\newpage

\begin{wrapfigure}{r}{0.5\textwidth}
\begin{center}
    \includegraphics[width=0.5\textwidth]{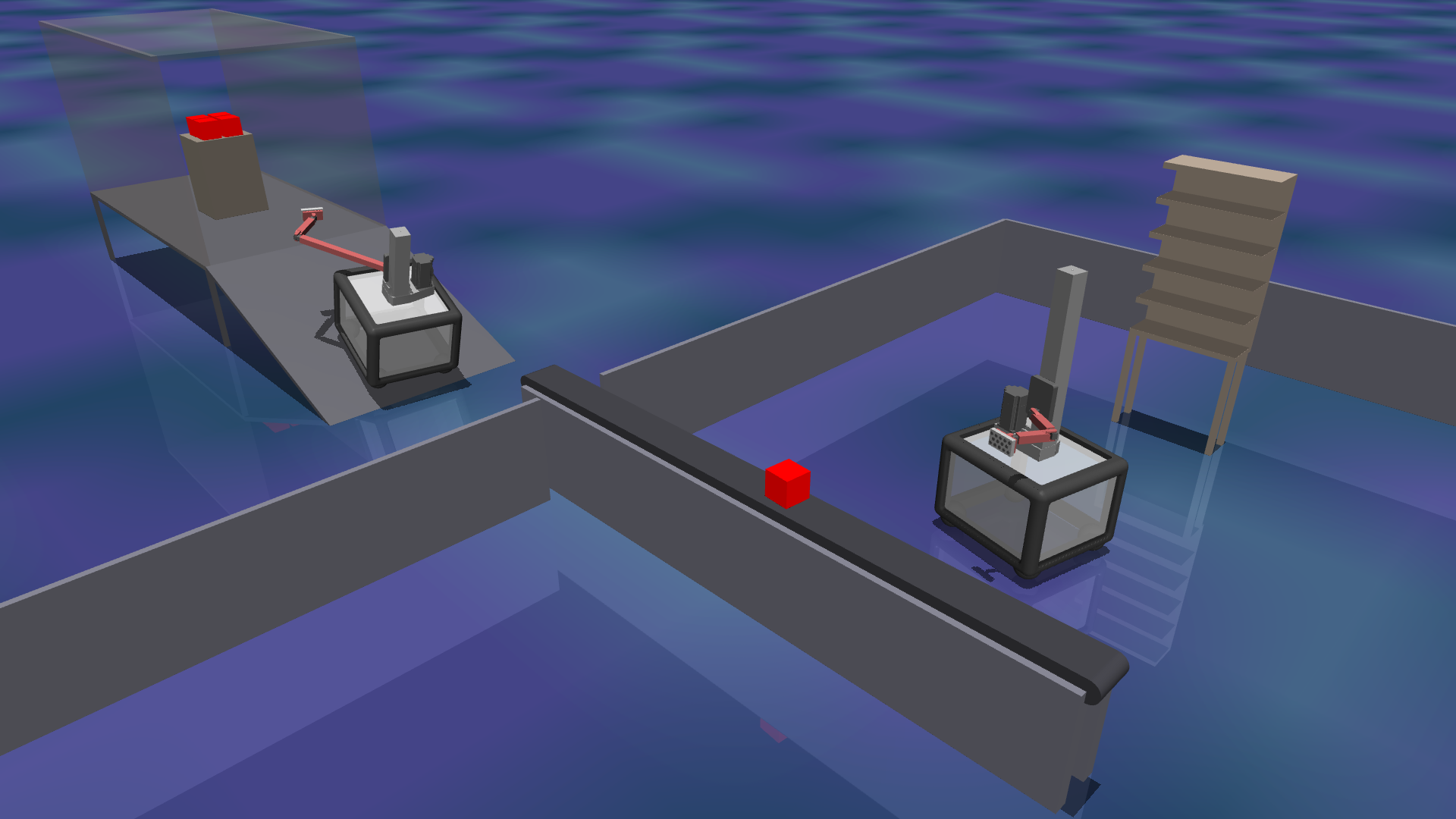}
    \label{fig:envs:truckunloading}
    \vspace{-1.5em}
    \caption{Truck Unloading environment.}
    \vspace{-1.5em}
\end{center}
\end{wrapfigure}
\textbf{Truck Unloading.} In this task mobile manipulator robots are tasked with unloading a truck packed with boxes. One mobile manipulator collects boxes from the truck and places them on a conveyor belt, around which other robots are placed. These must collect different boxes from the conveyor (keyed by color) and place them in a specified area. We provide two presets. 1) TruckUnloadSingle in which there is a single mobile manipulator collecting the red boxes from the conveyor belt, and placing them on a shelf. 2) TruckUnloadMulti in which there is an additional mobile manipulator collecting the blue boxes from the conveyor, and stack these boxes into an area behind them. In these tasks, the control optimization loop is training a single policy to control all of the robots. The design problem is to modify the robot arms on the mobile manipulators to enable dexterous and quick unloading and packing of boxes.

\begin{wrapfigure}{r}{0.6\textwidth}
\vspace{-1.5em}
\begin{center}
    \includegraphics[width=0.6\textwidth]{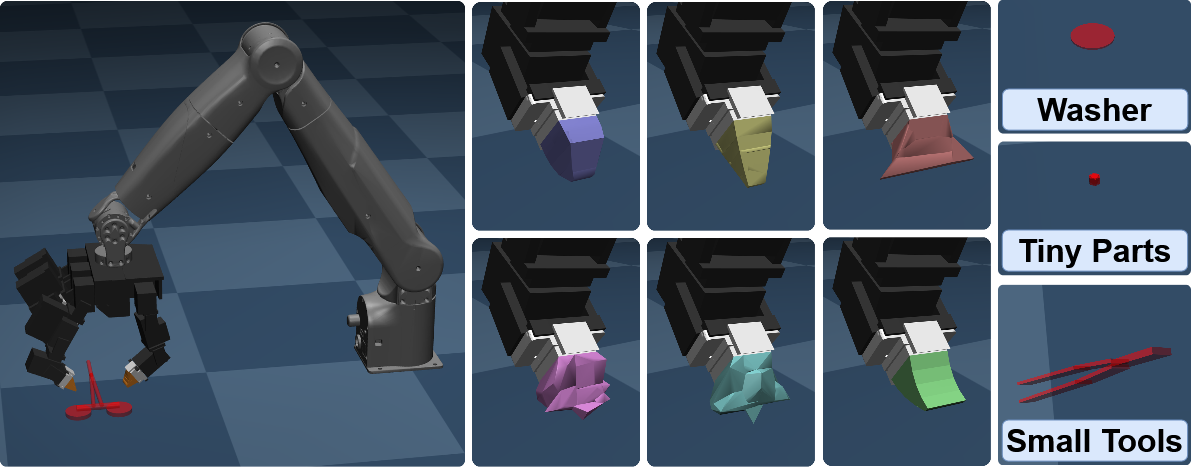}
    \label{fig:envs:nerograsp}
    \vspace{-1.5em}
    \caption{Nero-LEAP Grasping environment.}
    \vspace{-1.5em}
\end{center}
\end{wrapfigure}
\textbf{Nero-LEAP Grasping.} A Nero robot arm with a LEAP hand is tasked with picking up small objects from a flat surface. The LEAP hand has fingertips affixed. The objects are randomized from a set of ten divided into three categories. A success in this task is considered a sustained, controlled, and stable lift of the object. We provide four main presets. 1) NeroGraspSmall in which the object is randomly selected from the `small` category. 2) NeroGraspTiny in which the object is randomly selected from the `tiny` category. 3) NeroGraspThin in which the object is a thin washer whose radius is randomly selected. 4) NeroGraspAll in which the object is randomly selected from all ten available objects. There are additional presets for each object, which are further described in Appendix~\ref{appx:additional-presets}. The trajectory of the Nero arm is fixed, and the control optimization loop is training a policy for the LEAP hand. The design problem in this task is to modify the meshes of the LEAP fingertips on the thumb and index, to enable a stable and sustained lift. The base vertices of the mesh are fixed to the robot, but the remainder of the mesh can be fully replaced by the user.

\begin{wrapfigure}{r}{0.73\textwidth}
\vspace{-1.0em}
\begin{center}
    \includegraphics[width=0.73\textwidth]{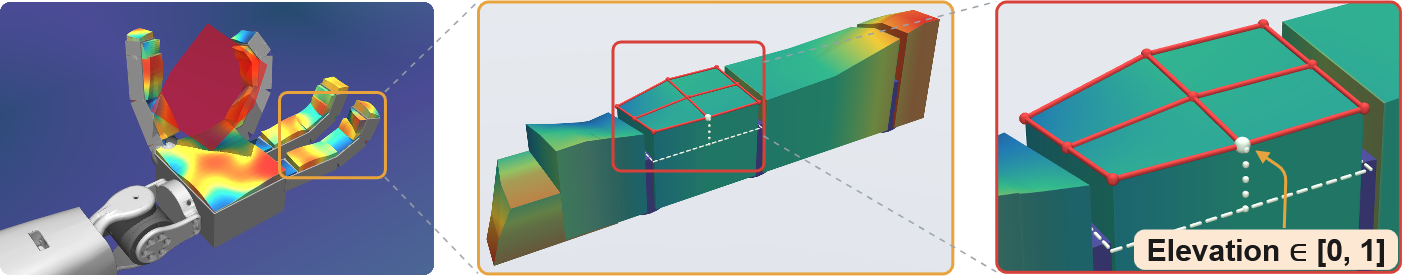}
    \vspace{-1.5em}
    \caption{MorphHand Rotation environment.}
    \vspace{-1.5em}
\end{center}
\end{wrapfigure}
\textbf{MorphHand Rotation.} In this environment an anthropomorphic robot hand is tasked with reorienting and rotating an object in its palm to a target orientation relative to the palm. We provide a single preset in which an 8\,cm cube is used, the goal tolerance is 0.1\,rad and new targets are generated 1 radian away from a met target. The control optimization loop trains a policy for the hand and wrist, while the design problem is the modify the surface of the hand through height-field displacements.  

\newpage

\begin{wrapfigure}{r}{0.65\textwidth}
\vspace{-0.5em}
\begin{center}
    \includegraphics[width=0.65\textwidth]{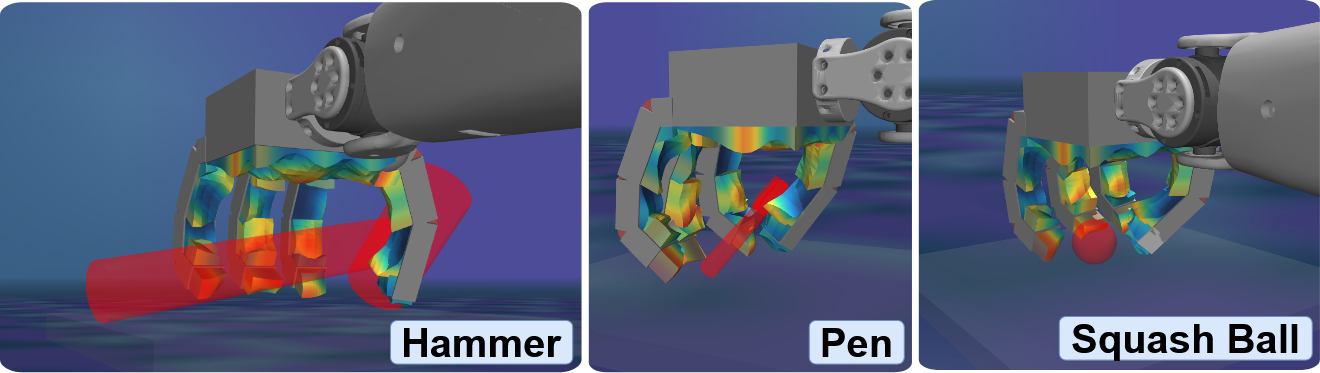}
    \label{fig:envs:handpickup}
    \vspace{-1.5em}
    \caption{MorphHand Pick-Up environment.}
    \vspace{-1.5em}
\end{center}
\end{wrapfigure}
\textbf{MorphHand Pick-Up.} In this environment an anthropomorphic hand is tasked to pick up an object from a flat surface and place it in a certain position and orientation. We provide three presets. 1) MorphHandPickUp-Cube in which the hand must grasp an 8\,cm cube. 2) MorphHandPickUp-SquashBall in which the hand must grasp a 4\,cm ball. 3) MorphHandPickUp-Hammer in which the hand must grasp a hammer. The control optimization loop is training a policy to control the hand and wrist, and the design problem is the modify the surface of the hand through height-field displacements and finger configuration parameters (spacing and width).  

\begin{wrapfigure}{l}{0.45\textwidth}
\begin{center}
    \includegraphics[width=0.45\textwidth]{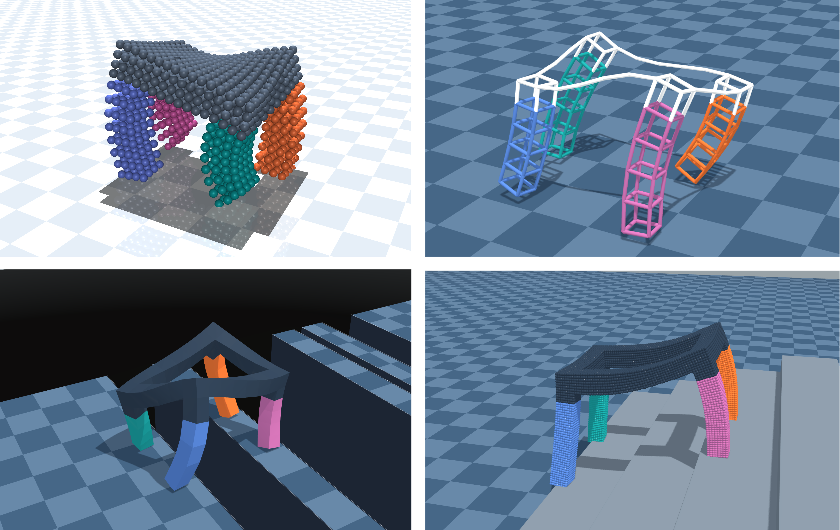}
    \label{fig:envs:softwalkers}
    \vspace{-1.5em}
    \caption{Soft Walkers environment.}
    \vspace{-1.5em}
\end{center}
\end{wrapfigure}
\textbf{Soft Walkers.} In this task a soft robot is tasked with efficiently traversing a given domain. We provide three terrain presets: 1) a flat terrain; 2) a terrain with equidistant stairs; 3) a terrain with gaps and holes. For each of these three terrains, we provide five versions of the problem each supported by GPU-accelerated implicit Euler integration: 1) in 2D using a Spring-solver~\citep{Bhatia2022EvolutionGA}; 2) in 2D using an Material Point Method (MPM)-solver~\citep{spielberg2023advanced}; 3) in 3D using an MPM-solver~\citep{spielberg2023advanced}; 4) in 3D using a Beam-solver~\citep{cheney2014unshackling}; 5) in 3D using an Finite Element Method (FEM)-solver~\citep{ma2021diffaqua}. This leads to a total of 15 unique presets. In these tasks, the control optimization loop trains a policy to control the actuation of the soft robot walker. The design problem specifies the shape, mass, actuation, and material properties of the robot. This task was inspired by \cite{Bhatia2022EvolutionGA} and \cite{cheney2014unshackling}, and is built on NVIDIA Warp~\cite{macklin_warp_2022}.

\begin{wrapfigure}{r}{0.44\textwidth}
\begin{center}
    \includegraphics[width=0.44\textwidth]{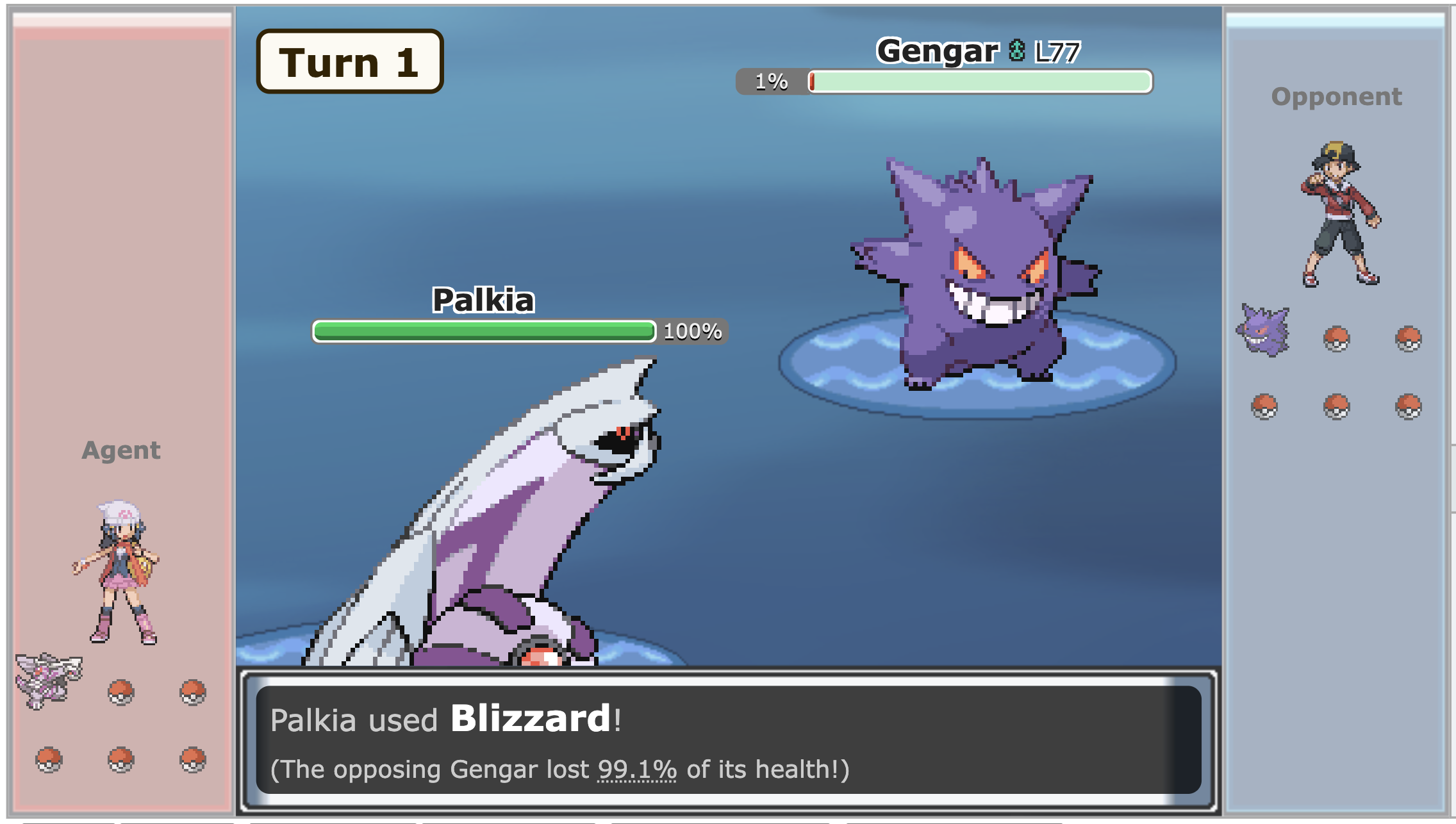}
    \label{fig:envs:pokenv}
    \vspace{-1.5em}
    \caption{The Pokemon Showdown co-design environment.}
    \vspace{-1.5em}
\end{center}
\end{wrapfigure}
\textbf{Pokemon Showdown.} In this task a Pokemon battle is conducted using a team of six Pokemon tasked with winning the battle or sequence of battles against an opponent trainer. We provide four main presets. 1) PokenvRandom in which the opponent's Pokemon team is randomly sampled and the opponent agent takes random moves in a Pokemon Generation 4 Single Battle. 2) PokenvMaxPower in which the opponent's Pokemon team is randomly sampled and the opponent agent chooses moves with the highest base power in a Pokemon Generation 6 Single Battle. 3) PokenvHeuristic in which the opponent's Pokemon team is randomly sampled and the opponent agent plays according to the SimpleHeuristicPlayer~\citep{hsahovic_poke_env_2019} in a Pokemon Generation 9 Single Battle.
4) PokenvLeague in which the opponent is a sequence of 5-opponents representing the Pokemon League, an elite set of trainers from the video games. This preset has a number of variants, which we discuss in more detail in Appendix~\ref{appx:additional-presets}. In this task, the control optimization loop is tasked with finding the best policy to play with a given Pokemon team, and the design problem is to build the optimal Pokemon team. This task is built on Pokemon Showdown~\citep{smogon_pokemon_showdown_2011} and Poke-env~\citep{hsahovic_poke_env_2019}.

\begin{wrapfigure}{l}{0.45\textwidth}
\begin{center}
    \includegraphics[width=0.45\textwidth]{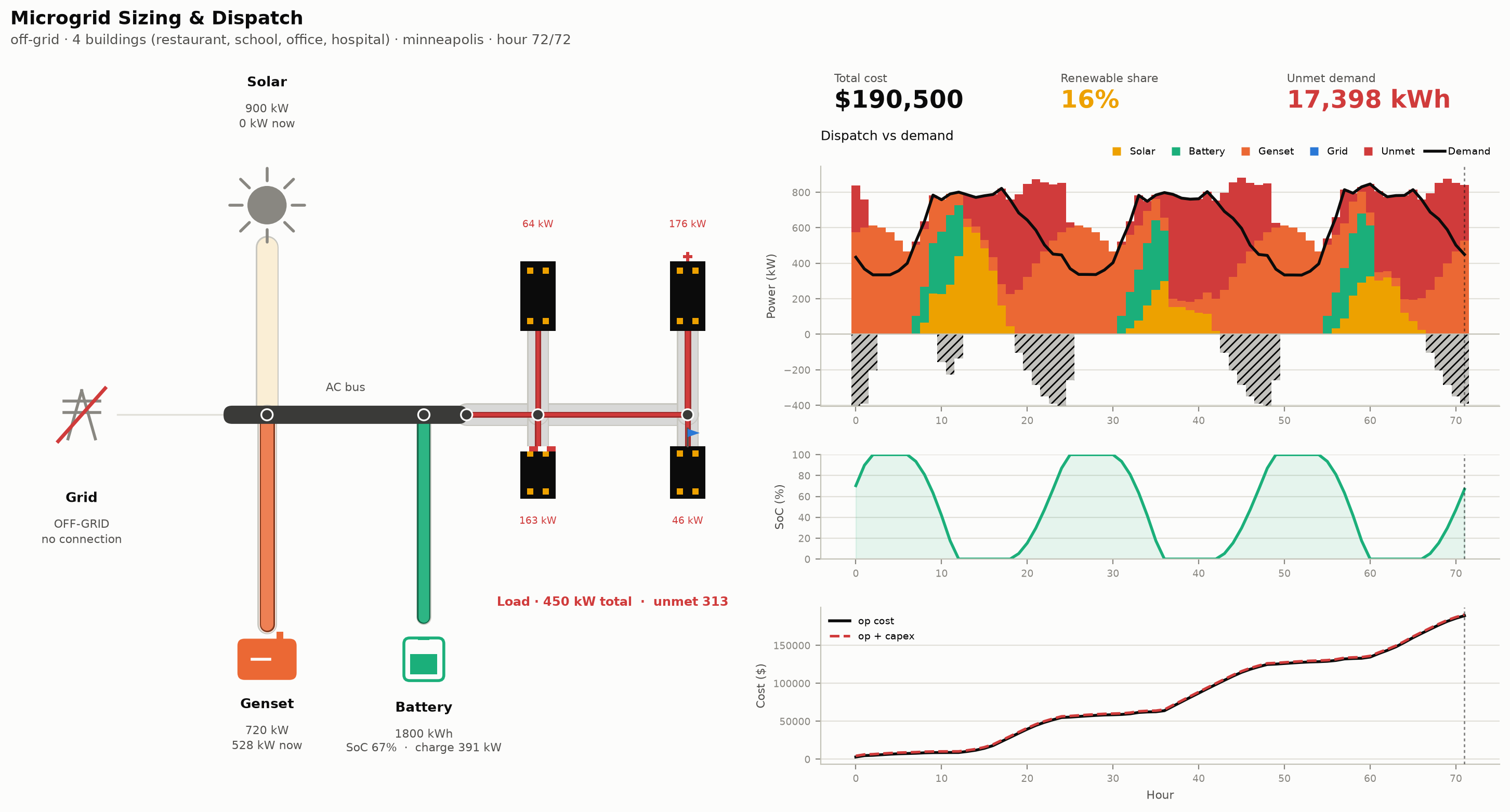}
    \label{fig:envs:microgrid}
    \vspace{-1.5em}
    \caption{Electrical Microgrid environment.}
    \vspace{-1.5em}
\end{center}
\end{wrapfigure}

\newpage

\textbf{Electrical Microgrid.} In this budgeted optimization task an energy Microgrid is supplying power to a small collection of buildings, each with their own power demands and schedules which are randomised. A success is considered when there is no unmet demand and there was little unnecessary spending over a given time frame. We provide three presets where the Microgrid is connected to a larger grid. 1) MicrogridRandom (24 hours) in which there is a single random building connected to the grid. 2) MicrogridForecast (24 hours) in which there is a single random building connected to the grid and observations are augmented with noisy 12 hour forecasts. 3) MicrogridCampus (24 hours) in which there are several buildings. We provide a further two presets where the Microgrid is not connected to a larger grid. 1) MicrogridOffGridSingle (72 hours) in which there is a single random building connected to the grid. 2) MicrogridOffGridCampus (72 hours) in which there are several buildings connected the grid. In these tasks, the control optimization loop is training a policy for energy flow, and the design problem specifies solar capacity, battery size, and diesel generator capacities. Each design choice incurs a cost which is paid in the task reward at the first step, forming a budgeted co-optimization problem. This task is built on \texttt{python-microgrid}~\citep{Henri2020pymgridAO}.

\begin{wrapfigure}{r}{0.45\textwidth}
\vspace{-1.75em}
\begin{center}
    \includegraphics[width=0.45\textwidth]{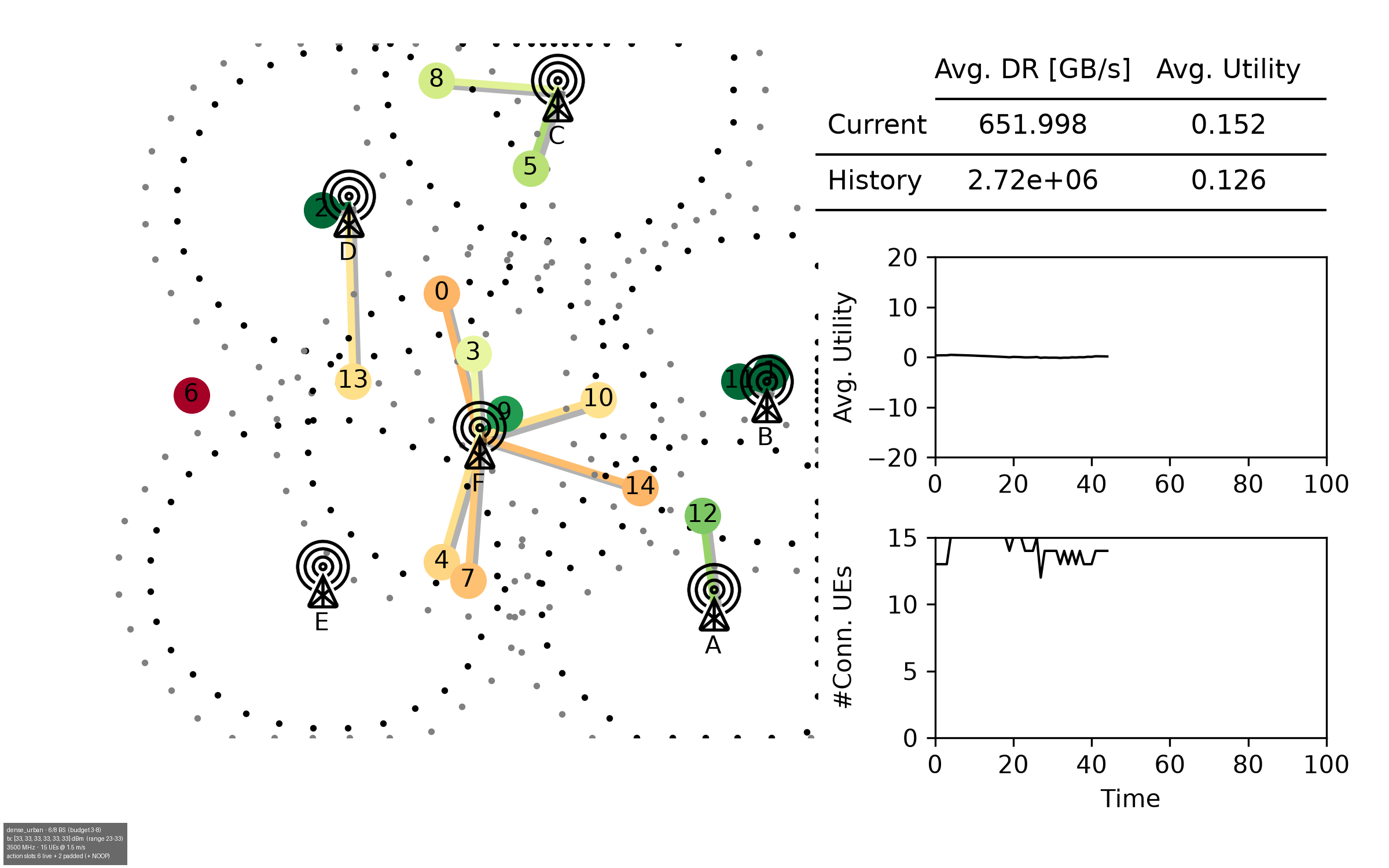}
    \label{fig:envs:mobile}
    \vspace{-1.5em}
    \caption{Wireless Network environment.}
    \vspace{-1.5em}
\end{center}
\end{wrapfigure}

\textbf{Wireless Network.} In this budgeted optimization task a set of wireless stations are covering a high-traffic area providing service to mobile-phone users. Success in this task is directly correlated with the strength and regularity of the signal provided across all users. We provide three presets. 1) NetworkUrban in which pedestrian low-velocity users are moving in random movement patterns. 2) NetworkSuburban in which there is a mix of velocities with some vehicular users in random movement patterns. 3) NetworkMotorway in which users are travelling at high velocities, and terrain requires sparse coverage. In these tasks, the control optimization loop chooses which base station each mobile user is provided connection to as they move around the map. The design problem specifies how many base stations to deploy (under a maximum), where they are placed, and what transmission power capacity each station is given. This task is built on \citet{schneider2022mobileenv}.

\begin{wrapfigure}{l}{0.4\textwidth}
\begin{center}
    \includegraphics[width=0.4\textwidth]{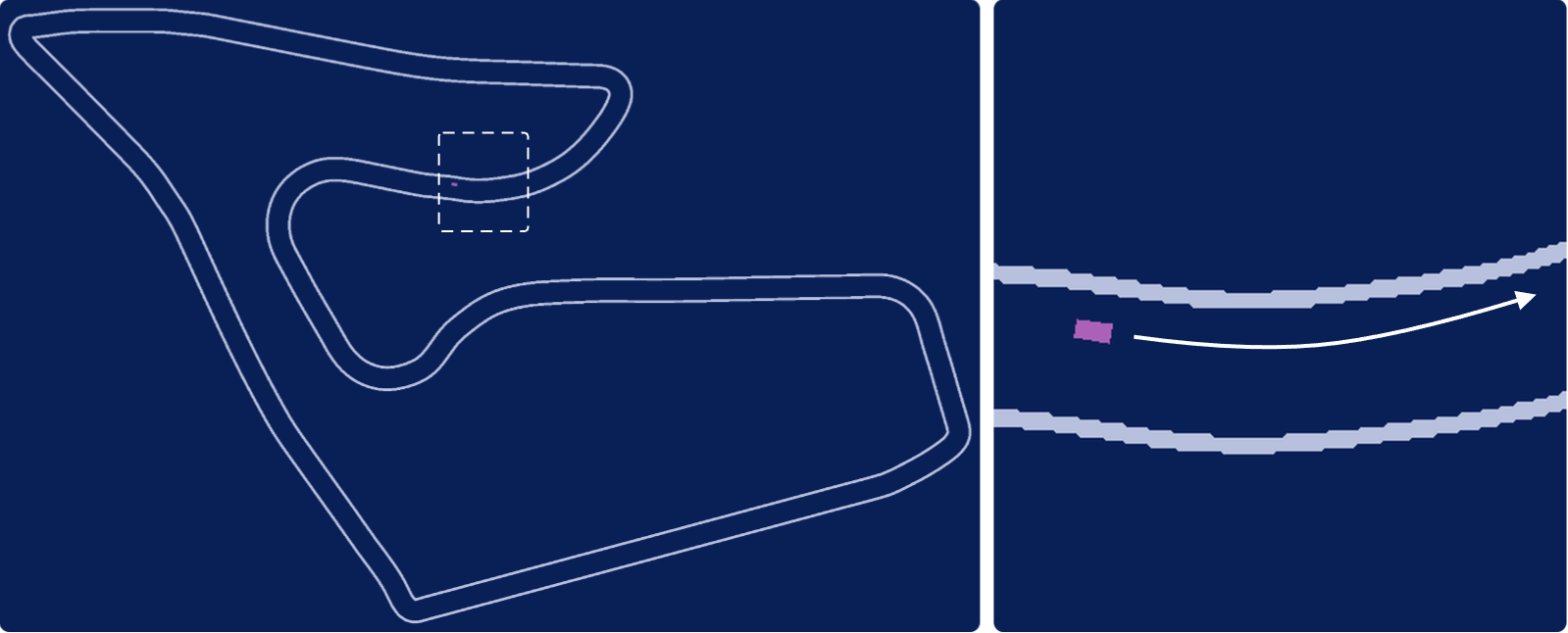}
    \label{fig:envs:carracing}
    \vspace{-1.5em}
    \caption{Car Racing environment.}
    \vspace{-1.5em}
\end{center}
\end{wrapfigure}

\textbf{Car Racing.} In this task a scaled racing car is driving around a track, trying to complete laps in the lowest possible time. Each car is modelled with load sensitivity, yaw-inertia, and friction trade-offs. We provide two presets. 1) RacingLevine in which the car is racing around the Levine track preset. 2) RacingSpielberg in which the car is racing around the Spielberg track preset. In these tasks, the control optimization loop is training a policy for the car and the design problem specifies the vehicle setup. This configuration includes tyre grip, cornering stiffness, drive power, and more which all possess trade-offs. This task is built on F1Tenth ~\citep{okelly2020f1tenth}.

\newpage

\begin{wrapfigure}{r}{0.3\textwidth}
\begin{center}
    \includegraphics[width=0.3\textwidth]{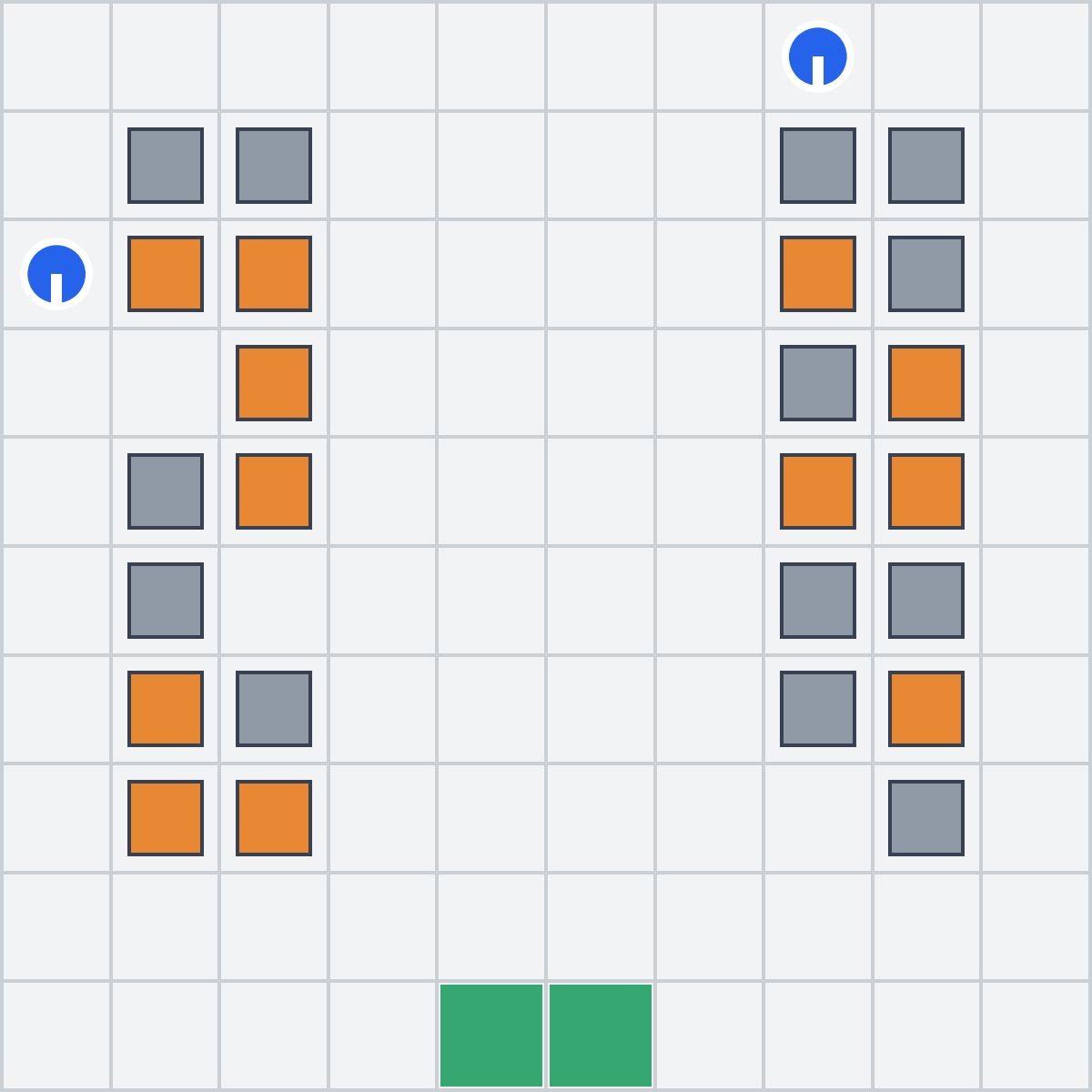}
    \label{fig:envs:robotwarehouse}
    \vspace{-1.5em}
    \caption{Robotic Warehouse environment.}
    \vspace{-1.5em}
\end{center}
\end{wrapfigure}

\textbf{Robotic Warehouse.} In this task a warehouse is equipped with multiple robotic agents with the goal of providing the highest throughput of request completion. These requests are deliveries to goals, and they are randomized for each environment within a certain max load. We provide three presets. 1) WarehouseSmall in which there is a 10x10 warehouse floor, two robots, and $\geq$25 shelves. 2) WarehouseMedium in which there is a 16x16 warehouse floor, four robots, and $\geq$115 shelves. 3) WarehouseCongested in which there is a 20x22 warehouse floor, eight robots, and $\geq$264 shelves. In these tasks, the control optimization loop is training a policy for the fleet of robots to jointly navigate them to retrieve requested shelves and deliver them to goals. The design problem specifies the layout and placement of the shelves on a fixed floor trading off route length and congestion. This task is built on Jumanji~\citep{bonnet2024jumanji} and RWARE~\citep{semitable_multi_robot_warehouse_rware_2020}.

\begin{wrapfigure}{l}{0.4\textwidth}
\vspace{-1.75em}
\begin{center}
    \includegraphics[width=0.4\textwidth]{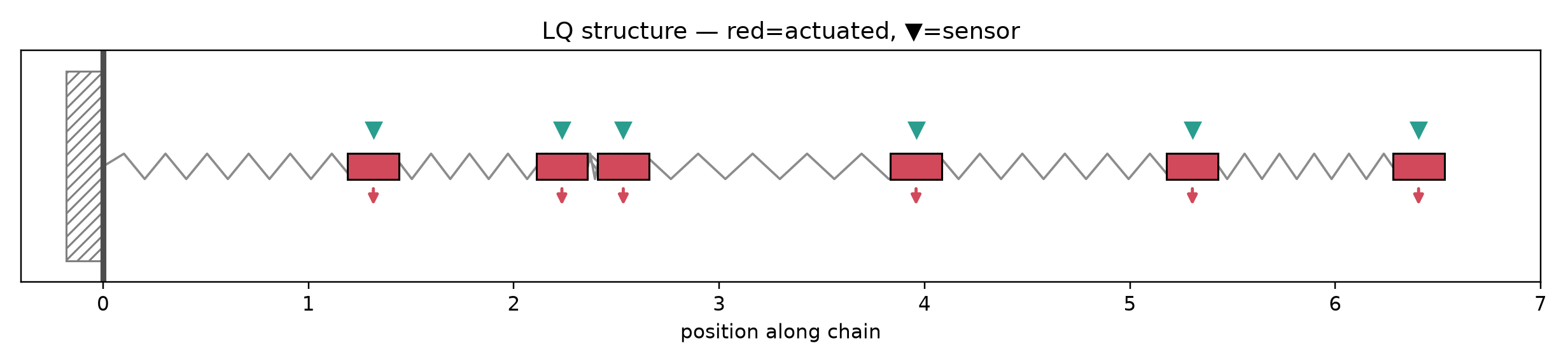}
    \label{fig:envs:lqstructure}
    \vspace{-1.5em}
    \caption{Linear-Quadratic Control environment.}
    \vspace{-1.5em}
\end{center}
\end{wrapfigure}

\textbf{Linear-Quadratic Control.} In this task a mass-spring-damper chain is regulated against environmental disturbances. We provide three presets: 1) continuous design with unconstrained, full-state control; 2)
actuator and sensor placement with control under process and measurement noise; and 3) actuator placement under a mass and actuator budget with saturated control inputs. In these tasks, the control optimization loop trains a
policy to regulate the structure against the disturbances, while the design problem specifies the masses, stiffnesses, damping, and the actuator/sensor placements. Because the three presets span continuous, combinatorial, and
budget-constrained design, this environment is designed to serve as a lightweight sanity-checker exercising the distinct facets of a co-design algorithm.

\subsection{Additional Environment Presets}
\label{appx:additional-presets}

Alongside the main presets described and presented in Section~\ref{sec:gym}. There are additional environment configurations exposed by the gym for the Nero-LEAP Grasping and Pokemon Showdown environments. \\

\subsubsection{Nero-LEAP Grasping}

The NeroGrasp class of environment presets contains ten objects. These are scissors, tweezers, spatula, stick, screw, long screw, headless screw, t-nut, lock nut, and washer. Each object is categorised in the main presets, but can be launched on its own as well. This exposes nine additional presets. \\

\begin{center}
\renewcommand{\arraystretch}{1.35}
\setlength{\tabcolsep}{12pt}

\begin{tabular}{|p{5cm}|p{5cm}|}
\hline
\multicolumn{2}{|c|}{\textbf{NeroGrasp Additional Presets}} \\
\hline
\textbf{Object Name} & \textbf{Preset Name} \\
\hline
Scissors        & NeroGraspScissors   \\
Tweezers        & NeroGraspTweezers   \\
Spatula         & NeroGraspSpatula    \\
Stick           & NeroGraspStick      \\
Screw           & NeroGraspScrew      \\
Long Screw      & NeroGraspScrewLong  \\
Headless Screw  & NeroGraspSetScrew   \\
T-Nut           & NeroGraspTNut       \\
Lock Nut        & NeroGraspLockNut    \\
\hline
\end{tabular}
\end{center}

\subsubsection{Pokemon Showdown}

The Pokenv class of environment presets exposes the PokenvLeague preset. This environment is the most challenging of the Pokemon presets, and involves fighting 5 designed opponents in sequence, mimicking the Pokemon League from the video games. Over the nine generations of Pokemon video games, there are nine iterations of the Pokemon League, who are typically challenged by the player towards the end of the main storyline. In this environment preset, we therefore instantiate all nine of these leagues with their teams taken directly from the video games, each controlled by the SimpleHeuristicPlayer~\citep{hsahovic_poke_env_2019}. Each league exposes two presets. One where the agent's team is ``healed'' and reset to full strength between battles, and one where it is not. The second class of presets is substantially harder, and requires the co-design learner to design a team and play such that it can survive 5 successive opponents. In addition to these 18 presets, we also expose two additional presets which randomly select the opponent league from all nine generations.

\begin{center}
  \renewcommand{\arraystretch}{1.3}
  \setlength{\tabcolsep}{10pt}
  \begin{tabular}{@{}llcc@{}}
  \toprule
  \multicolumn{4}{c}{\textbf{PokenvLeague Presets}} \\
  \midrule
  & & \multicolumn{2}{c}{\textbf{Team status reset?}} \\
  \cmidrule(l){3-4}
  \textbf{Opponent league} & \textbf{Base name} & \textbf{Yes} & \textbf{No ($+$\texttt{H})} \\
  \midrule
  Generation $X$, $X \in \{1,\dots,9\}$ & \texttt{PokenvLeague-G$X$}   & \texttt{-G$X$}   & \texttt{-G$X$H}   \\
  Random generation                     & \texttt{PokenvLeague-Rand}  & \texttt{-Rand}   & \texttt{-RandH}   \\
  \bottomrule
  \end{tabular}
  \end{center}

\section{Baseline Optimization Methods}\label{sec:baselines}

\begin{wrapfigure}{r}{0.4\textwidth}

\begin{center}
    \vspace{-1.75em}
    \includegraphics[width=0.4\textwidth]{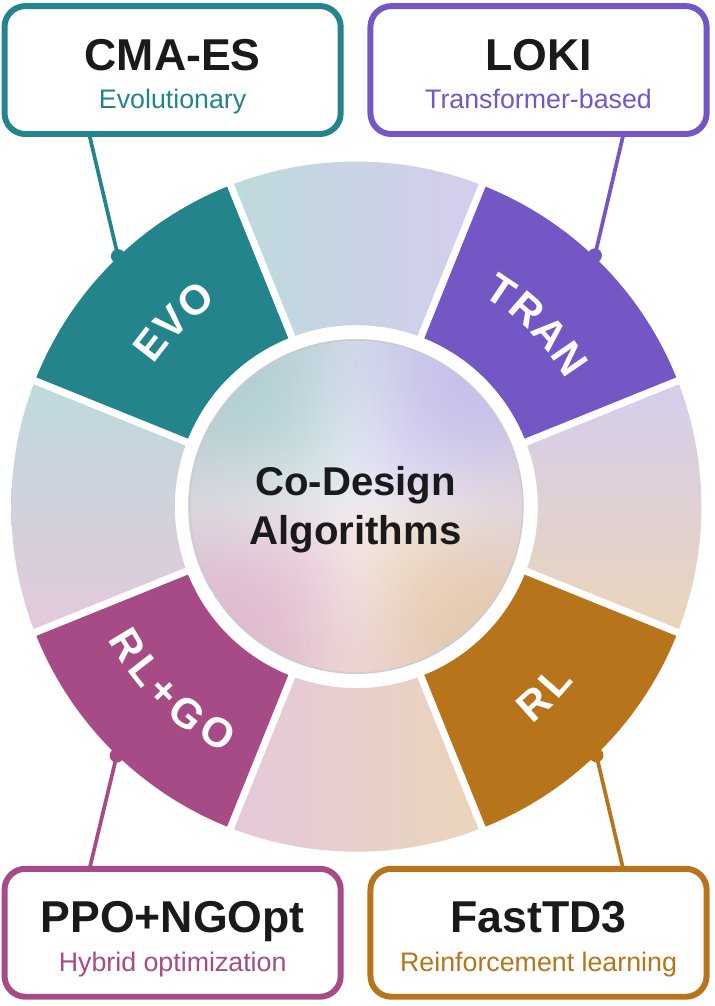}
    \label{fig:baselines:algos_overview}
    \vspace{-1.5em}
    \caption{An overview of the space of co-design algorithms highlighting our chosen main baselines.}
    \vspace{-1.5em}
\end{center}
\end{wrapfigure}

\textbf{Baseline 1: CMA-ES}. The CMA-ES evolutionary global optimizer is deployed for co-design by sampling embodiments and controller parameters from the algorithm. This puts both the controller and the embodiment into the same joint search space from which the algorithm samples. The controller parameters sampled are used in a neural-network controller and the embodiment values are passed directly to the environment. At evaluation boundaries (every 5\% of training), the current best controller-embodiment pair is evaluated in multiple parallel evaluation environments on held-out seeds.

\textbf{Baseline 2: FastTD3}. FastTD3 is an off-policy state-of-the-art \gls{acr:rl} algorithm predominantly designed for continuous control tasks, such as robot control. To instantiate FastTD3 for co-design, we expand its action space to include both the embodiment parameters and the control actions. At the first step of a FastTD3 rollout, the chosen embodiment values are used to instantiate/reset the environment and the control actions are discarded. In subsequent steps, the embodiment values are discarded and the control actions are used to step through the environment. The replay buffer is populated with the corrected samples (after setting either the embodiment values or control actions to 0) so that the policy learns from which actions were actually taken. At evaluation boundaries (every 5\% of training), the current policy generates a new embodiment and control actions for multiple parallel evaluation environments on held-out seeds.

\textbf{Baseline 3: PPO and NGOpt}. In this baseline PPO and NGOpt are used in alternation to perform co-design. PPO is used as normal to train the controller. After each PPO update iteration, NGOpt samples one embodiment per training environment. Samples from NGOpt are scored using the mean return obtained by that embodiment in the rollouts in which it was used. At evaluation boundaries (every 5\% of training), the current best embodiment is used to instantiate multiple parallel evaluation environments on held-out seeds and is controlled with the current PPO policy.

\textbf{Baseline 4: LOKI~\citep{jeon2026convergent}}. In this baseline algorithm, we generalise the approach of~\citet{jeon2026convergent} to arbitrary embodiment parametrisations. For each set of embodiment parameters $\embody$, we construct a naive tokenization scheme, in place of the robot-centric tokenization scheme from the original work. Following this, we randomly sample $500,000$ valid embodiments and pre-train the Transformer VAE on these embodiments following the original training recipe. Afterwards, we run the online training loop of~\citet{jeon2026convergent} with 6 clusters (as opposed to the original 40) to ensure comparability with other baselines, using PPO to train Transformer controllers. At evaluation boundaries (every 5\% of online training), the controller from the best cluster is selected and the top performing embodiments from that cluster are used to instantiate multiple parallel evaluation environments on held-out seeds.

\subsection*{Additional baselines}

\textbf{LA-MCTS~\citep{wang2020lamcts}.} LA-MCTS searches the joint vector space of the embodiment parameters, bounded by the design space, concatenated with the weights of a controller. It learns a tree of search-space partitions, splitting the samples seen so far with k-means clustering on their rewards and an RBF-kernel SVM, selects a leaf with an upper confidence bound, and samples uniformly inside the leaf's region. We use the authors' implementation with its default settings (exploration constant 0.1, leaf size 10, random sampler) and three adaptations for our setting: every iteration proposes one batch of 128 candidates that is scored in a single parallel rollout; the root keeps only the 2{,}048 best samples, so that fitting the partition remains tractable for genomes with thousands of dimensions; and the exploration constant is scaled by the magnitude of the best reward, so that the exploration bonus stays positive when rewards are negative. A candidate's fitness is the mean undiscounted return over the training worlds assigned to it, and at evaluation boundaries the best candidate so far is evaluated on held-out seeds.

\textbf{SPSA~\citep{spall1992spsa}.} Simultaneous-perturbation stochastic approximation estimates the gradient of the return from pairs of symmetrically perturbed evaluations and also searches the joint vector space of the embodiment parameters, bounded by the design space, concatenated with the weights of a controller. Each step draws 64 independent Rademacher perturbations, scores all 128 resulting $\pm$ points in one rollout, and moves along the average of the 64 gradient estimates (gradient-averaged SPSA). The gain sequences follow Nevergrad's SPSA ($c_k = c/(k+1)^{0.101}$, $a_k = a/(k+1+A)^{0.602}$, $c = 0.1$, $A = 10$) in Nevergrad's standardized parameter space. Because a fixed step gain $a$ barely moves a policy at our reward scales, $a$ is calibrated from the first five gradient estimates so that a step moves the parameters by 0.1 standardized units on average, and each step is capped at its scheduled size, guarding against the occasional very large gradient estimate under sparse rewards.

\textbf{Controls.} Two controls isolate the two halves of co-design. Both reuse the PPO+NGOpt implementation and its per-preset configuration. \emph{Policy only} (PPO, fixed design) runs the PPO updates of PPO+NGOpt with the design search switched off: the policy is trained on the preset's base embodiment throughout, with the same PPO settings and network, and is evaluated on the base embodiment. \emph{Design only} (NGOpt, random policy) runs only the NGOpt design rounds of PPO+NGOpt under a frozen policy that is never trained. The policy has the PPO+NGOpt architecture with random weights; its action layer uses standard PyTorch initialization, because PPO's default output gain of 0.01 would make an untrained policy output nearly constant actions. Each round scores candidate embodiments by their mean return under the frozen policy, and evaluation uses the best embodiment found so far under the same frozen policy.

\section{Detailed results}
\label{sec:detailed-results}

This appendix reports the raw returns behind the normalized scores of Table~\ref{table:baselines}, the normalization anchors, policy-only and design-only controls, two additional baselines, and the compute and environment interactions of every method. All tables and figures are generated from the same runs (5 seeds per preset and method).

Tables~\ref{table:full-results-raw} and~\ref{table:full-results-norm} give the raw and normalized final evaluation returns of the four baselines on all 35 presets.

\subsection{Normalized scores}
\label{sec:normalization}

We follow \citet{Mnih2013PlayingAW} and report $100 \times (R - R_{\text{rand}}) / (R_{\text{ref}} - R_{\text{rand}})$ for a final evaluation return $R$. The low anchor $R_{\text{rand}}$ is the mean return of a uniform-random policy on the preset's base embodiment, measured with the same evaluation code as the baselines (5 seeds $\times$ 128 episodes). As no human reference exists for co-design tasks, the high anchor $R_{\text{ref}}$ is the preset's theoretical maximum episode return, derived from its reward function, episode length and terminal bonuses. Where this maximum is not finite (the MuJoCo locomotion presets and SoftWalker, whose forward-velocity rewards are unbounded), $R_{\text{ref}}$ is a cap derived from the environment's physics: the episode length times the maximum forward reward implied by the actuator, joint and design-space limits, with zero control cost. No observed result is used to set either anchor. Several references are loose upper bounds (e.g.\ Warehouse, NeroGrasp, the hand-manipulation cube presets, Swimmer and SoftWalker), so normalized scores on these presets are conservative. For NeroGrasp, where the code does not bound how high a held object can rise, the reference assumes that the object stays at or below 0.2\,m, below the highest palm position of the scripted lift (0.234\,m): $100$ steps $\times$ (0.2 + 0.02 contact) + 59 $\times$ 0.003 proximity + 1 for success $\approx 23.2$. Table~\ref{table:anchors} lists both anchors for every preset.

\begin{table}[htbp]
\centering
\footnotesize
\setlength{\tabcolsep}{3pt}
\begin{tabular}[t]{@{}lrr@{}}
\toprule
Preset & \shortstack[r]{Random\\agent (0\%)} & \shortstack[r]{Reference\\(100\%)} \\
\midrule
\multicolumn{3}{@{}l}{\textit{Native}} \\
BallCatcher-Pitch & 0.0826 & 10.0 \\
BallCatcher-Arc & 0.0794 & 10.0 \\
BallCatcher-Fielding & 0.0894 & 10.0 \\
NeroGraspAll & 0.0846 & 23.2 \\
HandPickUpCube & $-$0.862 & 170 \\
HandPickUpHammer & $-$0.194 & 10.0 \\
RotateInHandCube & $-$4.47 & 376 \\
SoftWalkerBeam3D & $-$2.80 & 60.0 \\
SolarCleanerReach & $-$0.00114 & 10.0 \\
SolarCleanerTraverse & $-$0.000280 & 10.0 \\
TruckUnloadSingle & 0.0392 & 10.0 \\
\midrule
\multicolumn{3}{@{}l}{\textit{Extended}} \\
HalfCheetah & $-$295 & 22{,}800 \\
Hopper & 18.3 & 15{,}800 \\
Walker2d & 2.77 & 21{,}400 \\
Swimmer & 1.25 & 3{,}400 \\
Ant & $-$52.0 & 20{,}900 \\
Humanoid & 123 & 21{,}400 \\
HumanoidStandup & 33{,}464 & 648{,}000 \\
\bottomrule
\end{tabular}
\qquad
\begin{tabular}[t]{@{}lrr@{}}
\toprule
Preset & \shortstack[r]{Random\\agent (0\%)} & \shortstack[r]{Reference\\(100\%)} \\
\midrule
\multicolumn{3}{@{}l}{\textit{Reframed}} \\
\multicolumn{3}{@{}l}{Racing} \\
\hspace{0.8em}Spielberg & 1.75 & 344 \\
\hspace{0.8em}Levine & 1.47 & 64.2 \\
\multicolumn{3}{@{}l}{Warehouse} \\
\hspace{0.8em}Small & 0.00625 & 1{,}000 \\
\hspace{0.8em}Medium & 0 & 1{,}000 \\
\hspace{0.8em}Congested & 0.00313 & 1{,}000 \\
\multicolumn{3}{@{}l}{Network} \\
\hspace{0.8em}Urban & $-$73.9 & 100 \\
\hspace{0.8em}Suburban & 4.05 & 100 \\
\hspace{0.8em}Motorway & 15.7 & 100 \\
\multicolumn{3}{@{}l}{Microgrid} \\
\hspace{0.8em}Random & $-$53{,}173 & 720 \\
\hspace{0.8em}Forecast & $-$53{,}173 & 720 \\
\hspace{0.8em}Campus & $-$211{,}339 & 2{,}880 \\
\hspace{0.8em}OffGridSingle & $-$68{,}952 & 0 \\
\hspace{0.8em}OffGridCampus & $-$262{,}181 & 0 \\
\multicolumn{3}{@{}l}{Pokenv} \\
\hspace{0.8em}Random & $-$11.5 & 28.0 \\
\hspace{0.8em}MaxPower & $-$14.4 & 28.0 \\
\hspace{0.8em}Heuristic & $-$16.7 & 28.0 \\
LQStructure (LQR) & $-$28.2 & 0 \\
\bottomrule
\end{tabular}

\caption{Normalization anchors: return of a uniform-random policy on the base embodiment (0\%) and reference return (100\%): the theoretical maximum episode return, or a physics-based cap where the return is unbounded (Appendix~\ref{sec:normalization}).}
\label{table:anchors}
\end{table}

\begin{table}[htbp]
\centering
\footnotesize
\setlength{\tabcolsep}{3pt}
\begin{tabular}{@{}lrrrr@{}}
\toprule
Preset & \shortstack[r]{CMA-\\ES} & \shortstack[r]{Fast\\TD3} & \shortstack[r]{PPO+\\NGOpt} & LOKI \\
\midrule
\multicolumn{5}{@{}l}{\textit{Native}} \\
BallCatcher-Pitch & 0.296 $\pm$ 0.12 & \textbf{3.42 $\pm$ 1.6} & 0.435 $\pm$ 0.33 & 0.0633 $\pm$ 0.0021 \\
BallCatcher-Arc & 0.316 $\pm$ 0.11 & \textbf{4.15 $\pm$ 0.65} & 1.59 $\pm$ 0.23 & 0.0875 $\pm$ 0.014 \\
BallCatcher-Fielding & 0.143 $\pm$ 0.045 & \textbf{3.15 $\pm$ 2.5} & 0.461 $\pm$ 0.40 & 0.0739 $\pm$ 0.0037 \\
NeroGraspAll & \textbf{2.34 $\pm$ 0.43} & $-$250 $\pm$ 0 & 0.147 $\pm$ 0.18 & 0.0899 $\pm$ 0.019 \\
HandPickUpCube & \textbf{2.11 $\pm$ 0.43} & 1.25 $\pm$ 1.4 & 0.561 $\pm$ 0.017 & 0.127 $\pm$ 0.14 \\
HandPickUpHammer & \textbf{2.61 $\pm$ 0.19} & 1.04 $\pm$ 1.7 & 0.752 $\pm$ 0.026 & 0.305 $\pm$ 0.090 \\
RotateInHandCube & 0.244 $\pm$ 0.081 & \textbf{1.81 $\pm$ 2.0} & 1.58 $\pm$ 0.70 & $-$2.89 $\pm$ 0.51 \\
SoftWalkerBeam3D & $-$0.00720 $\pm$ 0.0023 & $-$0.0481 $\pm$ 0.054 & \textbf{0.0445 $\pm$ 0.048} & $-$1.24 $\pm$ 0.17 \\
SolarCleanerReach & 0.0249 $\pm$ 0.0015 & 0.00639 $\pm$ 0.0018 & \textbf{0.186 $\pm$ 0.27} & 0.00613 $\pm$ 0.0021 \\
SolarCleanerTraverse & \textbf{0.0185 $\pm$ 0.016} & 0.00431 $\pm$ 0.0033 & 0.000430 $\pm$ 0.0013 & 0.00689 $\pm$ 0.00033 \\
TruckUnloadSingle & 0.840 $\pm$ 0.39 & \textbf{2.26 $\pm$ 2.4} & 0.790 $\pm$ 0.097 & 0.506 $\pm$ 0.0076 \\
\midrule
\multicolumn{5}{@{}l}{\textit{Reframed}} \\
\multicolumn{5}{@{}l}{Racing} \\
\hspace{0.8em}Spielberg & 6.77 $\pm$ 2.0 & \textbf{76.9 $\pm$ 39} & 64.1 $\pm$ 8.8 & 38.4 $\pm$ 1.8 \\
\hspace{0.8em}Levine & 6.07 $\pm$ 1.0 & \textbf{48.3 $\pm$ 12} & 39.6 $\pm$ 10 & 17.3 $\pm$ 2.0 \\
\multicolumn{5}{@{}l}{Warehouse} \\
\hspace{0.8em}Small & 0.00313 $\pm$ 0.0038 & 0 $\pm$ 0 & \textbf{12.6 $\pm$ 2.4} & 0.738 $\pm$ 0.44 \\
\hspace{0.8em}Medium & 0 $\pm$ 0 & 0 $\pm$ 0 & \textbf{0.0172 $\pm$ 0.013} & 0 $\pm$ 0 \\
\hspace{0.8em}Congested & 0 $\pm$ 0 & $-$50.0 $\pm$ 100 & \textbf{0.00313 $\pm$ 0.0063} & 0 $\pm$ 0 \\
\multicolumn{5}{@{}l}{Network} \\
\hspace{0.8em}Urban & $-$87.2 $\pm$ 2.0 & $-$100.0 $\pm$ 0.0016 & $-$81.4 $\pm$ 5.6 & \textbf{$-$76.2 $\pm$ 1.1} \\
\hspace{0.8em}Suburban & $-$15.8 $\pm$ 2.4 & $-$92.0 $\pm$ 1.7 & \textbf{$-$6.99 $\pm$ 3.2} & $-$30.3 $\pm$ 1.4 \\
\hspace{0.8em}Motorway & 39.5 $\pm$ 2.5 & $-$83.0 $\pm$ 28 & \textbf{57.4 $\pm$ 2.6} & 46.5 $\pm$ 0.32 \\
\multicolumn{5}{@{}l}{Microgrid} \\
\hspace{0.8em}Random & $-$2{,}578 $\pm$ 549 & \textbf{$-$1{,}286 $\pm$ 259} & $-$19{,}698 $\pm$ 1{,}023 & $-$25{,}107 $\pm$ 130 \\
\hspace{0.8em}Forecast & $-$4{,}958 $\pm$ 285 & \textbf{$-$1{,}124 $\pm$ 114} & $-$22{,}178 $\pm$ 195 & $-$25{,}432 $\pm$ 246 \\
\hspace{0.8em}Campus & $-$14{,}121 $\pm$ 1{,}442 & \textbf{$-$5{,}793 $\pm$ 449} & $-$92{,}551 $\pm$ 11{,}476 & $-$100{,}053 $\pm$ 752 \\
\hspace{0.8em}OffGridSingle & $-$16{,}576 $\pm$ 5{,}792 & \textbf{$-$13{,}917 $\pm$ 8{,}719} & $-$81{,}464 $\pm$ 1{,}249 & $-$19{,}909 $\pm$ 967 \\
\hspace{0.8em}OffGridCampus & \textbf{$-$38{,}235 $\pm$ 2{,}479} & $-$81{,}419 $\pm$ 50{,}542 & $-$260{,}851 $\pm$ 4{,}386 & $-$75{,}925 $\pm$ 3{,}200 \\
\multicolumn{5}{@{}l}{Pokenv} \\
\hspace{0.8em}Random & $-$1.23 $\pm$ 4.2 & 3.81 $\pm$ 4.0 & 7.05 $\pm$ 4.0 & \textbf{8.18 $\pm$ 3.1} \\
\hspace{0.8em}MaxPower & $-$1.56 $\pm$ 7.3 & $-$3.83 $\pm$ 4.2 & $-$0.315 $\pm$ 3.3 & \textbf{3.13 $\pm$ 2.7} \\
\hspace{0.8em}Heuristic & $-$14.0 $\pm$ 4.5 & $-$8.49 $\pm$ 3.9 & \textbf{$-$3.64 $\pm$ 5.0} & $-$8.49 $\pm$ 3.2 \\
LQStructure (LQR) & $-$4.97 $\pm$ 0.47 & \textbf{$-$2.52 $\pm$ 0.36} & $-$3.69 $\pm$ 0.18 & $-$5.13 $\pm$ 0.19 \\
\midrule
\multicolumn{5}{@{}l}{\textit{Extended}} \\
HalfCheetah & \textbf{8{,}033 $\pm$ 442} & 3{,}317 $\pm$ 53 & 4{,}782 $\pm$ 145 & 3{,}400 $\pm$ 203 \\
Hopper & 1{,}261 $\pm$ 72 & 897 $\pm$ 184 & 1{,}588 $\pm$ 891 & \textbf{1{,}601 $\pm$ 213} \\
Walker2d & \textbf{3{,}428 $\pm$ 1{,}564} & 3{,}087 $\pm$ 736 & 3{,}257 $\pm$ 272 & 1{,}429 $\pm$ 177 \\
Swimmer & \textbf{405 $\pm$ 0.82} & 83.2 $\pm$ 1.3 & 70.4 $\pm$ 27 & 53.1 $\pm$ 12 \\
Ant & 3{,}726 $\pm$ 469 & 3{,}526 $\pm$ 170 & \textbf{4{,}708 $\pm$ 284} & 36.9 $\pm$ 5.7 \\
Humanoid & 730 $\pm$ 15 & \textbf{3{,}610 $\pm$ 1{,}654} & 684 $\pm$ 57 & 379 $\pm$ 14 \\
HumanoidStandup & \textbf{226{,}124 $\pm$ 23{,}094} & 73{,}522 $\pm$ 27{,}979 & 127{,}672 $\pm$ 15{,}381 & 79{,}271 $\pm$ 2{,}493 \\
\bottomrule
\end{tabular}

\caption{Final evaluation return of each baseline (mean $\pm$ std over 5 seeds; higher is better) under a 4-hour wall-clock budget (LOKI: 3 hours online after 1 hour of preparation). Bold: best normalized mean (Table~\ref{table:full-results-norm}).}
\label{table:full-results-raw}
\end{table}

\begin{table}[htbp]
\centering
\footnotesize
\setlength{\tabcolsep}{3pt}
\begin{tabular}{@{}lrrrr@{}}
\toprule
Preset & \shortstack[r]{CMA-\\ES} & \shortstack[r]{Fast\\TD3} & \shortstack[r]{PPO+\\NGOpt} & LOKI \\
\midrule
\multicolumn{5}{@{}l}{\textit{Native}} \\
BallCatcher-Pitch & 2.1 $\pm$ 1.2 & \textbf{33.6 $\pm$ 16.3} & 3.6 $\pm$ 3.3 & $-$0.2 $\pm$ 0.0 \\
BallCatcher-Arc & 2.4 $\pm$ 1.1 & \textbf{41.0 $\pm$ 6.5} & 15.2 $\pm$ 2.3 & 0.1 $\pm$ 0.1 \\
BallCatcher-Fielding & 0.5 $\pm$ 0.5 & \textbf{30.9 $\pm$ 25.0} & 3.8 $\pm$ 4.0 & $-$0.2 $\pm$ 0.0 \\
NeroGraspAll & \textbf{9.8 $\pm$ 1.9} & $-$1081.9 $\pm$ 0.0 & 0.3 $\pm$ 0.8 & 0.0 $\pm$ 0.1 \\
HandPickUpCube & \textbf{1.7 $\pm$ 0.3} & 1.2 $\pm$ 0.8 & 0.8 $\pm$ 0.0 & 0.6 $\pm$ 0.1 \\
HandPickUpHammer & \textbf{27.5 $\pm$ 1.9} & 12.1 $\pm$ 16.4 & 9.3 $\pm$ 0.3 & 4.9 $\pm$ 0.9 \\
RotateInHandCube & 1.2 $\pm$ 0.0 & \textbf{1.6 $\pm$ 0.5} & 1.6 $\pm$ 0.2 & 0.4 $\pm$ 0.1 \\
SoftWalkerBeam3D & 4.4 $\pm$ 0.0 & 4.4 $\pm$ 0.1 & \textbf{4.5 $\pm$ 0.1} & 2.5 $\pm$ 0.3 \\
SolarCleanerReach & 0.3 $\pm$ 0.0 & 0.1 $\pm$ 0.0 & \textbf{1.9 $\pm$ 2.7} & 0.1 $\pm$ 0.0 \\
SolarCleanerTraverse & \textbf{0.2 $\pm$ 0.2} & 0.0 $\pm$ 0.0 & 0.0 $\pm$ 0.0 & 0.1 $\pm$ 0.0 \\
TruckUnloadSingle & 8.0 $\pm$ 3.9 & \textbf{22.3 $\pm$ 23.6} & 7.5 $\pm$ 1.0 & 4.7 $\pm$ 0.1 \\
\midrule
\multicolumn{5}{@{}l}{\textit{Reframed}} \\
\multicolumn{5}{@{}l}{Racing} \\
\hspace{0.8em}Spielberg & 1.5 $\pm$ 0.6 & \textbf{21.9 $\pm$ 11.4} & 18.2 $\pm$ 2.6 & 10.7 $\pm$ 0.5 \\
\hspace{0.8em}Levine & 7.3 $\pm$ 1.6 & \textbf{74.6 $\pm$ 18.5} & 60.9 $\pm$ 16.5 & 25.3 $\pm$ 3.1 \\
\multicolumn{5}{@{}l}{Warehouse} \\
\hspace{0.8em}Small & $-$0.0 $\pm$ 0.0 & $-$0.0 $\pm$ 0.0 & \textbf{1.3 $\pm$ 0.2} & 0.1 $\pm$ 0.0 \\
\hspace{0.8em}Medium & 0.0 $\pm$ 0.0 & 0.0 $\pm$ 0.0 & \textbf{0.0 $\pm$ 0.0} & 0.0 $\pm$ 0.0 \\
\hspace{0.8em}Congested & $-$0.0 $\pm$ 0.0 & $-$5.0 $\pm$ 10.0 & \textbf{0.0 $\pm$ 0.0} & $-$0.0 $\pm$ 0.0 \\
\multicolumn{5}{@{}l}{Network} \\
\hspace{0.8em}Urban & $-$7.6 $\pm$ 1.1 & $-$15.0 $\pm$ 0.0 & $-$4.3 $\pm$ 3.2 & \textbf{$-$1.3 $\pm$ 0.6} \\
\hspace{0.8em}Suburban & $-$20.7 $\pm$ 2.5 & $-$100.1 $\pm$ 1.8 & \textbf{$-$11.5 $\pm$ 3.4} & $-$35.8 $\pm$ 1.4 \\
\hspace{0.8em}Motorway & 28.2 $\pm$ 3.0 & $-$117.2 $\pm$ 33.5 & \textbf{49.4 $\pm$ 3.1} & 36.5 $\pm$ 0.4 \\
\multicolumn{5}{@{}l}{Microgrid} \\
\hspace{0.8em}Random & 93.9 $\pm$ 1.0 & \textbf{96.3 $\pm$ 0.5} & 62.1 $\pm$ 1.9 & 52.1 $\pm$ 0.2 \\
\hspace{0.8em}Forecast & 89.5 $\pm$ 0.5 & \textbf{96.6 $\pm$ 0.2} & 57.5 $\pm$ 0.4 & 51.5 $\pm$ 0.5 \\
\hspace{0.8em}Campus & 92.1 $\pm$ 0.7 & \textbf{96.0 $\pm$ 0.2} & 55.5 $\pm$ 5.4 & 51.9 $\pm$ 0.4 \\
\hspace{0.8em}OffGridSingle & 76.0 $\pm$ 8.4 & \textbf{79.8 $\pm$ 12.6} & $-$18.1 $\pm$ 1.8 & 71.1 $\pm$ 1.4 \\
\hspace{0.8em}OffGridCampus & \textbf{85.4 $\pm$ 0.9} & 68.9 $\pm$ 19.3 & 0.5 $\pm$ 1.7 & 71.0 $\pm$ 1.2 \\
\multicolumn{5}{@{}l}{Pokenv} \\
\hspace{0.8em}Random & 26.0 $\pm$ 10.6 & 38.8 $\pm$ 10.1 & 47.0 $\pm$ 10.0 & \textbf{49.8 $\pm$ 7.9} \\
\hspace{0.8em}MaxPower & 30.2 $\pm$ 17.1 & 24.9 $\pm$ 10.0 & 33.2 $\pm$ 7.9 & \textbf{41.3 $\pm$ 6.3} \\
\hspace{0.8em}Heuristic & 5.9 $\pm$ 10.2 & 18.3 $\pm$ 8.8 & \textbf{29.2 $\pm$ 11.1} & 18.3 $\pm$ 7.2 \\
LQStructure (LQR) & 82.3 $\pm$ 1.7 & \textbf{91.0 $\pm$ 1.3} & 86.9 $\pm$ 0.6 & 81.8 $\pm$ 0.7 \\
\midrule
\multicolumn{5}{@{}l}{\textit{Extended}} \\
HalfCheetah & \textbf{36.1 $\pm$ 1.9} & 15.6 $\pm$ 0.2 & 22.0 $\pm$ 0.6 & 16.0 $\pm$ 0.9 \\
Hopper & 7.9 $\pm$ 0.5 & 5.6 $\pm$ 1.2 & 9.9 $\pm$ 5.6 & \textbf{10.0 $\pm$ 1.3} \\
Walker2d & \textbf{16.0 $\pm$ 7.3} & 14.4 $\pm$ 3.4 & 15.2 $\pm$ 1.3 & 6.7 $\pm$ 0.8 \\
Swimmer & \textbf{11.9 $\pm$ 0.0} & 2.4 $\pm$ 0.0 & 2.0 $\pm$ 0.8 & 1.5 $\pm$ 0.4 \\
Ant & 18.0 $\pm$ 2.2 & 17.1 $\pm$ 0.8 & \textbf{22.7 $\pm$ 1.4} & 0.4 $\pm$ 0.0 \\
Humanoid & 2.9 $\pm$ 0.1 & \textbf{16.4 $\pm$ 7.8} & 2.6 $\pm$ 0.3 & 1.2 $\pm$ 0.1 \\
HumanoidStandup & \textbf{31.4 $\pm$ 3.8} & 6.5 $\pm$ 4.6 & 15.3 $\pm$ 2.5 & 7.5 $\pm$ 0.4 \\
\bottomrule
\end{tabular}

\caption{Normalized score (\%) of each baseline (mean $\pm$ std over 5 seeds; 0 = uniform-random policy, 100 = reference return of Table~\ref{table:anchors}) under a 4-hour wall-clock budget (LOKI: 3 hours online after 1 hour of preparation). Bold: best mean.}
\label{table:full-results-norm}
\end{table}

\subsection{The benefit of co-optimization}
\label{sec:controls}

Two controls under the same budget separate the contributions of policy and embodiment (Tables~\ref{table:controls-full-raw} and~\ref{table:controls-full-norm}): PPO on the fixed base embodiment (policy only) and NGOpt design search under a frozen random policy (design only). A random policy earns little task reward on many presets (e.g.\ most robotic control tasks) but can act meaningfully on others (e.g.\ Pok\'emon battles), so we run design-only search on a subset of 13 presets. Design search alone falls short of the best co-design baseline on all 13, so policy learning is indispensable. Conversely, the best co-design baseline beats fixed-design PPO on 13 of 22 presets, by up to 32.5 points (MicrogridRandom), so the embodiment matters. Yet fixed-design PPO beats every co-design baseline on the other 9, and PPO+NGOpt, which shares its PPO learner, improves on it by more than one standard deviation on only 3. Current co-design methods thus do not reliably turn design freedom into better performance under a fixed budget, which is precisely the gap the benchmark is designed to expose.

Controls were run on the 22 presets listed in Tables~\ref{table:controls-full-raw} and~\ref{table:controls-full-norm}. Design-only search (NGOpt under a frozen random policy) was not run on presets whose task reward requires behaviour a random policy never achieves (e.g. lifting, grasping or delivering an object).

\begin{table}[htbp]
\centering
\footnotesize
\setlength{\tabcolsep}{3pt}
\begin{tabular}{@{}lrrrr@{}}
\toprule
Preset & \shortstack[r]{Random\\agent} & \shortstack[r]{Policy\\only} & \shortstack[r]{Design\\only} & \shortstack[r]{Best\\co-design} \\
\midrule
\multicolumn{5}{@{}l}{\textit{Native}} \\
BallCatcher-Pitch & 0.0826 & \textbf{6.06 $\pm$ 0.40} & 0.103 $\pm$ 0.019 & 3.42 $\pm$ 1.6 (F) \\
BallCatcher-Arc & 0.0794 & 2.77 $\pm$ 0.098 & 0.105 $\pm$ 0.029 & \textbf{4.15 $\pm$ 0.65 (F)} \\
BallCatcher-Fielding & 0.0894 & \textbf{5.19 $\pm$ 0.51} & 0.0816 $\pm$ 0.0096 & 3.15 $\pm$ 2.5 (F) \\
NeroGraspAll & 0.0846 & 0.203 $\pm$ 0.17 & -- & \textbf{2.34 $\pm$ 0.43 (C)} \\
HandPickUpCube & $-$0.862 & 1.81 $\pm$ 1.1 & -- & \textbf{2.11 $\pm$ 0.43 (C)} \\
HandPickUpHammer & $-$0.194 & 0.777 $\pm$ 0.0051 & -- & \textbf{2.61 $\pm$ 0.19 (C)} \\
RotateInHandCube & $-$4.47 & \textbf{4.94 $\pm$ 0.46} & -- & 1.81 $\pm$ 2.0 (F) \\
SoftWalkerBeam3D & $-$2.80 & $-$0.125 $\pm$ 0.083 & $-$0.0378 $\pm$ 0.064 & \textbf{0.0445 $\pm$ 0.048 (P)} \\
SolarCleanerReach & $-$0.00114 & \textbf{1.75 $\pm$ 0.072} & -- & 0.186 $\pm$ 0.27 (P) \\
TruckUnloadSingle & 0.0392 & 0.0905 $\pm$ 0.00030 & -- & \textbf{2.26 $\pm$ 2.4 (F)} \\
\midrule
\multicolumn{5}{@{}l}{\textit{Reframed}} \\
Racing (Spielberg) & 1.75 & \textbf{154 $\pm$ 26} & 0.388 $\pm$ 0.38 & 76.9 $\pm$ 39 (F) \\
Warehouse (Small) & 0.00625 & \textbf{22.3 $\pm$ 1.0} & -- & 12.6 $\pm$ 2.4 (P) \\
Network (Urban) & $-$73.9 & $-$85.3 $\pm$ 3.3 & $-$78.4 $\pm$ 5.0 & \textbf{$-$76.2 $\pm$ 1.1 (L)} \\
Microgrid (Random) & $-$53{,}173 & $-$18{,}795 $\pm$ 690 & $-$13{,}563 $\pm$ 5{,}535 & \textbf{$-$1{,}286 $\pm$ 259 (F)} \\
Pokenv (Random) & $-$11.5 & 6.69 $\pm$ 4.2 & $-$4.21 $\pm$ 5.5 & \textbf{8.18 $\pm$ 3.1 (L)} \\
LQStructure (LQR) & $-$28.2 & $-$3.75 $\pm$ 0.20 & $-$11.1 $\pm$ 1.9 & \textbf{$-$2.52 $\pm$ 0.36 (F)} \\
\midrule
\multicolumn{5}{@{}l}{\textit{Extended}} \\
HalfCheetah & $-$295 & 4{,}413 $\pm$ 1{,}183 & 660 $\pm$ 355 & \textbf{8{,}033 $\pm$ 442 (C)} \\
Hopper & 18.3 & \textbf{2{,}688 $\pm$ 802} & 394 $\pm$ 269 & 1{,}601 $\pm$ 213 (L) \\
Walker2d & 2.77 & \textbf{6{,}483 $\pm$ 1{,}309} & -- & 3{,}428 $\pm$ 1{,}564 (C) \\
Ant & $-$52.0 & \textbf{6{,}436 $\pm$ 246} & 579 $\pm$ 52 & 4{,}708 $\pm$ 284 (P) \\
Humanoid & 123 & 686 $\pm$ 51 & 279 $\pm$ 26 & \textbf{3{,}610 $\pm$ 1{,}654 (F)} \\
HumanoidStandup & 33{,}464 & 99{,}553 $\pm$ 14{,}090 & -- & \textbf{226{,}124 $\pm$ 23{,}094 (C)} \\
\bottomrule
\end{tabular}

\caption{Controls versus the best co-design baseline: final evaluation return (mean $\pm$ std over 5 seeds), with the uniform-random policy's return. Policy only: PPO on the fixed base embodiment. Design only: NGOpt under a frozen random policy (--: not run). Best co-design: best of the four baselines (C: CMA-ES, F: FastTD3, P: PPO+NGOpt, L: LOKI). Bold: best normalized mean of the three (Table~\ref{table:controls-full-norm}).}
\label{table:controls-full-raw}
\end{table}

\begin{table}[htbp]
\centering
\footnotesize
\setlength{\tabcolsep}{3pt}
\begin{tabular}{@{}lrrr@{}}
\toprule
Preset & \shortstack[r]{Policy\\only} & \shortstack[r]{Design\\only} & \shortstack[r]{Best\\co-design} \\
\midrule
\multicolumn{4}{@{}l}{\textit{Native}} \\
BallCatcher-Pitch & \textbf{60.3 $\pm$ 4.1} & 0.2 $\pm$ 0.2 & 33.6 $\pm$ 16.3 (F) \\
BallCatcher-Arc & 27.1 $\pm$ 1.0 & 0.3 $\pm$ 0.3 & \textbf{41.0 $\pm$ 6.5 (F)} \\
BallCatcher-Fielding & \textbf{51.4 $\pm$ 5.1} & $-$0.1 $\pm$ 0.1 & 30.9 $\pm$ 25.0 (F) \\
NeroGraspAll & 0.5 $\pm$ 0.8 & -- & \textbf{9.8 $\pm$ 1.9 (C)} \\
HandPickUpCube & 1.6 $\pm$ 0.6 & -- & \textbf{1.7 $\pm$ 0.3 (C)} \\
HandPickUpHammer & 9.5 $\pm$ 0.1 & -- & \textbf{27.5 $\pm$ 1.9 (C)} \\
RotateInHandCube & \textbf{2.5 $\pm$ 0.1} & -- & 1.6 $\pm$ 0.5 (F) \\
SoftWalkerBeam3D & 4.3 $\pm$ 0.1 & 4.4 $\pm$ 0.1 & \textbf{4.5 $\pm$ 0.1 (P)} \\
SolarCleanerReach & \textbf{17.5 $\pm$ 0.7} & -- & 1.9 $\pm$ 2.7 (P) \\
TruckUnloadSingle & 0.5 $\pm$ 0.0 & -- & \textbf{22.3 $\pm$ 23.6 (F)} \\
\midrule
\multicolumn{4}{@{}l}{\textit{Reframed}} \\
Racing (Spielberg) & \textbf{44.4 $\pm$ 7.7} & $-$0.4 $\pm$ 0.1 & 21.9 $\pm$ 11.4 (F) \\
Warehouse (Small) & \textbf{2.2 $\pm$ 0.1} & -- & 1.3 $\pm$ 0.2 (P) \\
Network (Urban) & $-$6.5 $\pm$ 1.9 & $-$2.6 $\pm$ 2.9 & \textbf{$-$1.3 $\pm$ 0.6 (L)} \\
Microgrid (Random) & 63.8 $\pm$ 1.3 & 73.5 $\pm$ 10.3 & \textbf{96.3 $\pm$ 0.5 (F)} \\
Pokenv (Random) & 46.1 $\pm$ 10.6 & 18.5 $\pm$ 13.8 & \textbf{49.8 $\pm$ 7.9 (L)} \\
LQStructure (LQR) & 86.7 $\pm$ 0.7 & 60.5 $\pm$ 6.6 & \textbf{91.0 $\pm$ 1.3 (F)} \\
\midrule
\multicolumn{4}{@{}l}{\textit{Extended}} \\
HalfCheetah & 20.4 $\pm$ 5.1 & 4.1 $\pm$ 1.5 & \textbf{36.1 $\pm$ 1.9 (C)} \\
Hopper & \textbf{16.9 $\pm$ 5.1} & 2.4 $\pm$ 1.7 & 10.0 $\pm$ 1.3 (L) \\
Walker2d & \textbf{30.3 $\pm$ 6.1} & -- & 16.0 $\pm$ 7.3 (C) \\
Ant & \textbf{31.0 $\pm$ 1.2} & 3.0 $\pm$ 0.2 & 22.7 $\pm$ 1.4 (P) \\
Humanoid & 2.6 $\pm$ 0.2 & 0.7 $\pm$ 0.1 & \textbf{16.4 $\pm$ 7.8 (F)} \\
HumanoidStandup & 10.8 $\pm$ 2.3 & -- & \textbf{31.4 $\pm$ 3.8 (C)} \\
\bottomrule
\end{tabular}

\caption{Controls versus the best co-design baseline: normalized score (\%, mean $\pm$ std over 5 seeds). Columns and markers as in Table~\ref{table:controls-full-raw}. Bold: best of the three.}
\label{table:controls-full-norm}
\end{table}

\subsection{Additional baselines: LA-MCTS and SPSA}
\label{sec:extra-baselines}

To broaden the set of optimizer families, we additionally ran LA-MCTS~\citep{wang2020lamcts}, a learned space-partitioning tree search, and SPSA~\citep{spall1992spsa}, a stochastic finite-difference gradient method, on 20 presets. Both search the same joint vector of embodiment parameters and controller weights as CMA-ES, with the same evaluation protocol. Tables~\ref{table:six-method-raw} and~\ref{table:six-method-norm} compare all six methods. LA-MCTS and SPSA lead only on NetworkUrban, where all other baselines score below the random policy, so the conclusion that no method dominates is unchanged.

\begin{table}[htbp]
\centering
\footnotesize
\setlength{\tabcolsep}{3pt}
\begin{tabular}{@{}lrrrrrr@{}}
\toprule
Preset & \shortstack[r]{CMA-\\ES} & \shortstack[r]{Fast\\TD3} & \shortstack[r]{PPO+\\NGOpt} & LOKI & \shortstack[r]{LA-\\MCTS} & SPSA \\
\midrule
\multicolumn{7}{@{}l}{\textit{Native}} \\
BallCatcher-Pitch & 0.296 & \textbf{3.42} & 0.435 & 0.0633 & 0.0841 & 0.348 \\
NeroGraspAll & \textbf{2.34} & $-$250 & 0.147 & 0.0899 & 0.889 & 1.47 \\
HandPickUpCube & \textbf{2.11} & 1.25 & 0.561 & 0.127 & 0.911 & 1.52 \\
RotateInHandCube & 0.244 & \textbf{1.81} & 1.58 & $-$2.89 & $-$0.0827 & 0.186 \\
SoftWalkerBeam3D & $-$0.00720 & $-$0.0481 & \textbf{0.0445} & $-$1.24 & $-$0.814 & $-$0.00824 \\
SolarCleanerReach & 0.0249 & 0.00639 & \textbf{0.186} & 0.00613 & 0.0189 & 0.0172 \\
TruckUnloadSingle & 0.840 & \textbf{2.26} & 0.790 & 0.506 & 0.407 & 0.509 \\
\midrule
\multicolumn{7}{@{}l}{\textit{Reframed}} \\
Racing (Spielberg) & 6.77 & \textbf{76.9} & 64.1 & 38.4 & 10.4 & 60.2 \\
Warehouse (Small) & 0.00313 & 0 & \textbf{12.6} & 0.738 & 0.00156 & 0.372 \\
Network (Urban) & $-$87.2 & $-$100.0 & $-$81.4 & $-$76.2 & \textbf{$-$61.2} & $-$62.1 \\
Microgrid (Random) & $-$2{,}578 & \textbf{$-$1{,}286} & $-$19{,}698 & $-$25{,}107 & $-$7{,}840 & $-$3{,}630 \\
Pokenv (Random) & $-$1.23 & 3.81 & 7.05 & \textbf{8.18} & $-$4.27 & $-$0.232 \\
LQStructure (LQR) & $-$4.97 & \textbf{$-$2.52} & $-$3.69 & $-$5.13 & $-$31.3 & $-$4.12 \\
\midrule
\multicolumn{7}{@{}l}{\textit{Extended}} \\
HalfCheetah & \textbf{8{,}033} & 3{,}317 & 4{,}782 & 3{,}400 & 2{,}388 & 6{,}986 \\
Hopper & 1{,}261 & 897 & 1{,}588 & \textbf{1{,}601} & 1{,}083 & 1{,}041 \\
Walker2d & \textbf{3{,}428} & 3{,}087 & 3{,}257 & 1{,}429 & 767 & 1{,}704 \\
Swimmer & \textbf{405} & 83.2 & 70.4 & 53.1 & 320 & 87.5 \\
Ant & 3{,}726 & 3{,}526 & \textbf{4{,}708} & 36.9 & $-$5.60 & 3{,}373 \\
Humanoid & 730 & \textbf{3{,}610} & 684 & 379 & 403 & 895 \\
HumanoidStandup & \textbf{226{,}124} & 73{,}522 & 127{,}672 & 79{,}271 & 96{,}923 & 155{,}449 \\
\bottomrule
\end{tabular}

\caption{All six baselines on the 20 presets with LA-MCTS and SPSA results: final evaluation return (mean over 5 seeds; the seed spread is shown in Table~\ref{table:six-method-norm}). Bold: best normalized mean.}
\label{table:six-method-raw}
\end{table}

\begin{table}[htbp]
\centering
\footnotesize
\setlength{\tabcolsep}{3pt}
\begin{tabular}{@{}lrrrrrr@{}}
\toprule
Preset & \shortstack[r]{CMA-\\ES} & \shortstack[r]{Fast\\TD3} & \shortstack[r]{PPO+\\NGOpt} & LOKI & \shortstack[r]{LA-\\MCTS} & SPSA \\
\midrule
\multicolumn{7}{@{}l}{\textit{Native}} \\
BallCatcher-Pitch & 2.1 $\pm$ 1.2 & \textbf{33.6 $\pm$ 16.3} & 3.6 $\pm$ 3.3 & $-$0.2 $\pm$ 0.0 & 0.0 $\pm$ 0.1 & 2.7 $\pm$ 2.8 \\
NeroGraspAll & \textbf{9.8 $\pm$ 1.9} & $-$1081.9 $\pm$ 0.0 & 0.3 $\pm$ 0.8 & 0.0 $\pm$ 0.1 & 3.5 $\pm$ 0.7 & 6.0 $\pm$ 4.4 \\
HandPickUpCube & \textbf{1.7 $\pm$ 0.3} & 1.2 $\pm$ 0.8 & 0.8 $\pm$ 0.0 & 0.6 $\pm$ 0.1 & 1.0 $\pm$ 0.2 & 1.4 $\pm$ 0.3 \\
RotateInHandCube & 1.2 $\pm$ 0.0 & \textbf{1.6 $\pm$ 0.5} & 1.6 $\pm$ 0.2 & 0.4 $\pm$ 0.1 & 1.2 $\pm$ 0.0 & 1.2 $\pm$ 0.0 \\
SoftWalkerBeam3D & 4.4 $\pm$ 0.0 & 4.4 $\pm$ 0.1 & \textbf{4.5 $\pm$ 0.1} & 2.5 $\pm$ 0.3 & 3.2 $\pm$ 0.3 & 4.4 $\pm$ 0.0 \\
SolarCleanerReach & 0.3 $\pm$ 0.0 & 0.1 $\pm$ 0.0 & \textbf{1.9 $\pm$ 2.7} & 0.1 $\pm$ 0.0 & 0.2 $\pm$ 0.0 & 0.2 $\pm$ 0.0 \\
TruckUnloadSingle & 8.0 $\pm$ 3.9 & \textbf{22.3 $\pm$ 23.6} & 7.5 $\pm$ 1.0 & 4.7 $\pm$ 0.1 & 3.7 $\pm$ 0.3 & 4.7 $\pm$ 0.5 \\
\midrule
\multicolumn{7}{@{}l}{\textit{Reframed}} \\
Racing (Spielberg) & 1.5 $\pm$ 0.6 & \textbf{21.9 $\pm$ 11.4} & 18.2 $\pm$ 2.6 & 10.7 $\pm$ 0.5 & 2.5 $\pm$ 0.5 & 17.1 $\pm$ 2.4 \\
Warehouse (Small) & $-$0.0 $\pm$ 0.0 & $-$0.0 $\pm$ 0.0 & \textbf{1.3 $\pm$ 0.2} & 0.1 $\pm$ 0.0 & $-$0.0 $\pm$ 0.0 & 0.0 $\pm$ 0.0 \\
Network (Urban) & $-$7.6 $\pm$ 1.1 & $-$15.0 $\pm$ 0.0 & $-$4.3 $\pm$ 3.2 & $-$1.3 $\pm$ 0.6 & \textbf{7.3 $\pm$ 1.4} & 6.8 $\pm$ 0.7 \\
Microgrid (Random) & 93.9 $\pm$ 1.0 & \textbf{96.3 $\pm$ 0.5} & 62.1 $\pm$ 1.9 & 52.1 $\pm$ 0.2 & 84.1 $\pm$ 1.1 & 91.9 $\pm$ 1.9 \\
Pokenv (Random) & 26.0 $\pm$ 10.6 & 38.8 $\pm$ 10.1 & 47.0 $\pm$ 10.0 & \textbf{49.8 $\pm$ 7.9} & 18.3 $\pm$ 26.2 & 28.6 $\pm$ 9.7 \\
LQStructure (LQR) & 82.3 $\pm$ 1.7 & \textbf{91.0 $\pm$ 1.3} & 86.9 $\pm$ 0.6 & 81.8 $\pm$ 0.7 & $-$11.2 $\pm$ 27.3 & 85.4 $\pm$ 1.0 \\
\midrule
\multicolumn{7}{@{}l}{\textit{Extended}} \\
HalfCheetah & \textbf{36.1 $\pm$ 1.9} & 15.6 $\pm$ 0.2 & 22.0 $\pm$ 0.6 & 16.0 $\pm$ 0.9 & 11.6 $\pm$ 2.2 & 31.5 $\pm$ 6.5 \\
Hopper & 7.9 $\pm$ 0.5 & 5.6 $\pm$ 1.2 & 9.9 $\pm$ 5.6 & \textbf{10.0 $\pm$ 1.3} & 6.7 $\pm$ 0.1 & 6.5 $\pm$ 0.2 \\
Walker2d & \textbf{16.0 $\pm$ 7.3} & 14.4 $\pm$ 3.4 & 15.2 $\pm$ 1.3 & 6.7 $\pm$ 0.8 & 3.6 $\pm$ 1.1 & 7.9 $\pm$ 4.8 \\
Swimmer & \textbf{11.9 $\pm$ 0.0} & 2.4 $\pm$ 0.0 & 2.0 $\pm$ 0.8 & 1.5 $\pm$ 0.4 & 9.4 $\pm$ 0.3 & 2.5 $\pm$ 0.0 \\
Ant & 18.0 $\pm$ 2.2 & 17.1 $\pm$ 0.8 & \textbf{22.7 $\pm$ 1.4} & 0.4 $\pm$ 0.0 & 0.2 $\pm$ 0.0 & 16.3 $\pm$ 2.6 \\
Humanoid & 2.9 $\pm$ 0.1 & \textbf{16.4 $\pm$ 7.8} & 2.6 $\pm$ 0.3 & 1.2 $\pm$ 0.1 & 1.3 $\pm$ 0.0 & 3.6 $\pm$ 0.9 \\
HumanoidStandup & \textbf{31.4 $\pm$ 3.8} & 6.5 $\pm$ 4.6 & 15.3 $\pm$ 2.5 & 7.5 $\pm$ 0.4 & 10.3 $\pm$ 0.9 & 19.9 $\pm$ 1.0 \\
\bottomrule
\end{tabular}

\caption{All six baselines on the 20 presets with LA-MCTS and SPSA results: normalized score (\%) (mean $\pm$ std over 5 seeds). Bold: best normalized mean.}
\label{table:six-method-norm}
\end{table}

\subsection{Compute and environment interactions}

\begin{figure}[p]
  \centering
  \includegraphics[width=\linewidth]{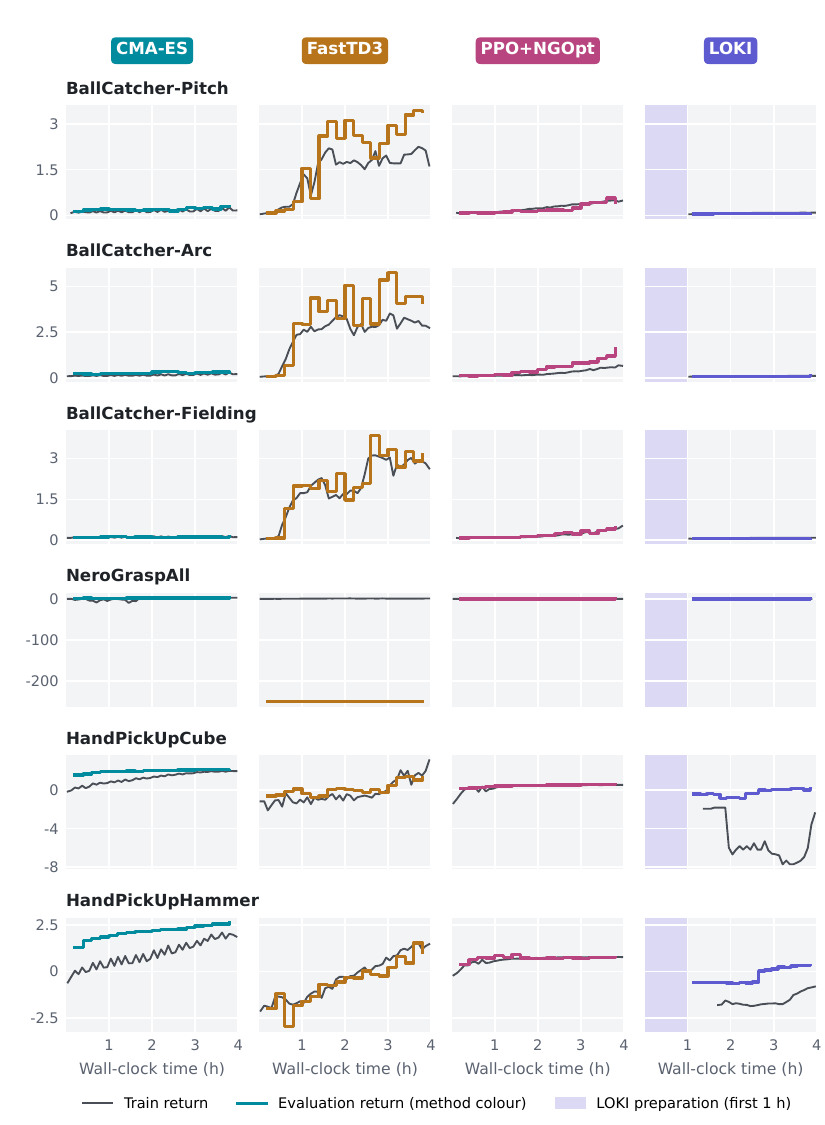}
  \caption{\textbf{Train vs.\ eval return, Native presets (1/2)} (mean over 5 seeds, each panel sharing its row's y-axis. Train return averages over the designs being trained or explored and eval return uses the selected design (each seed's latest evaluation carried forward), so the gap measures each method's cost of design exploration. LOKI's curves start after its 1-hour preparation (shaded).)}
  \label{fig:train-eval-native}
\end{figure}

\begin{figure}[p]
  \centering
  \includegraphics[width=\linewidth]{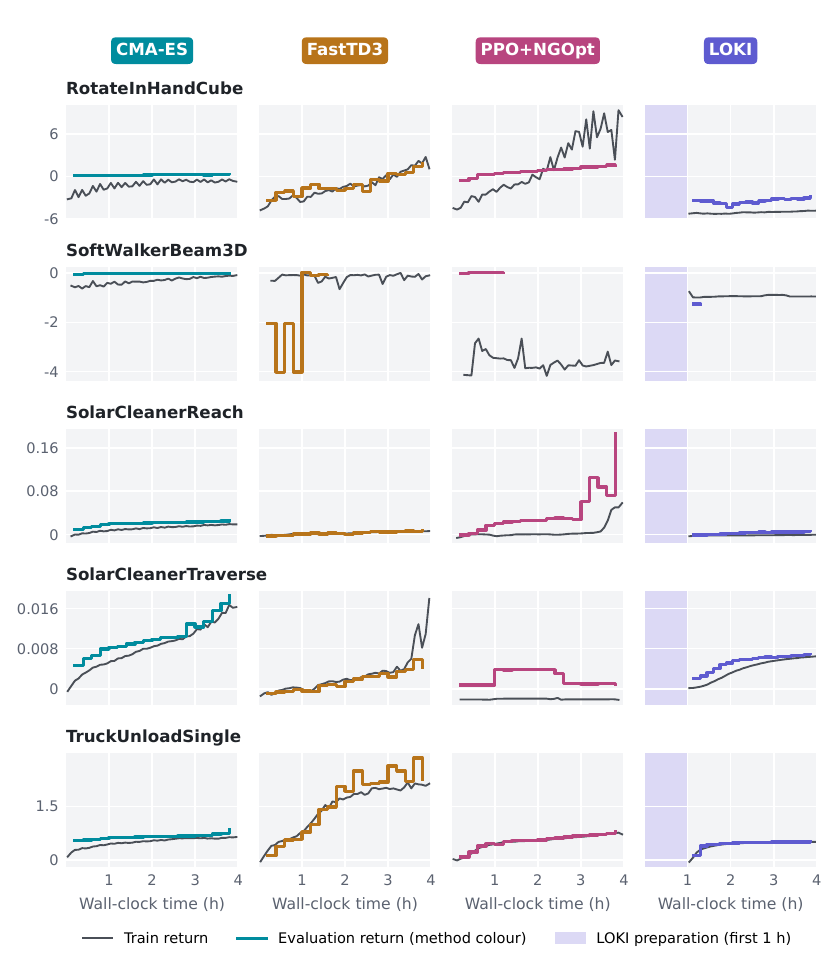}
  \caption{\textbf{Train vs.\ eval return, Native presets (2/2)} (continued; see Figure~\ref{fig:train-eval-native}). On SoftWalker-Beam3D many evaluations did not complete within the budget, so its eval curves end early.}
  \label{fig:train-eval-native-2}
\end{figure}

\begin{figure}[p]
  \centering
  \includegraphics[width=\linewidth]{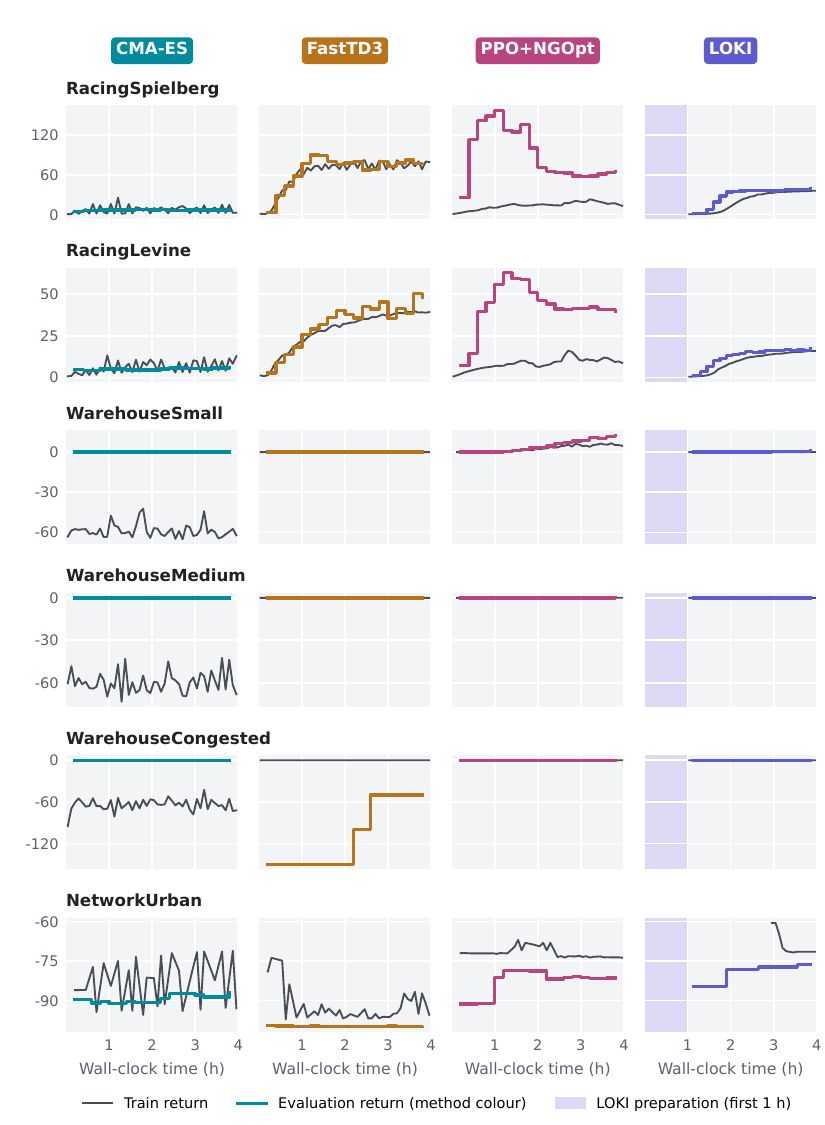}
  \caption{\textbf{Train vs.\ eval return, Reframed presets (1/3)} (mean over 5 seeds, each panel sharing its row's y-axis. Train return averages over the designs being trained or explored and eval return uses the selected design (each seed's latest evaluation carried forward), so the gap measures each method's cost of design exploration. LOKI's curves start after its 1-hour preparation (shaded).) For CMA-ES on the Warehouse presets the train return also includes the penalty for rejected designs.}
  \label{fig:train-eval-reframed}
\end{figure}

\begin{figure}[p]
  \centering
  \includegraphics[width=\linewidth]{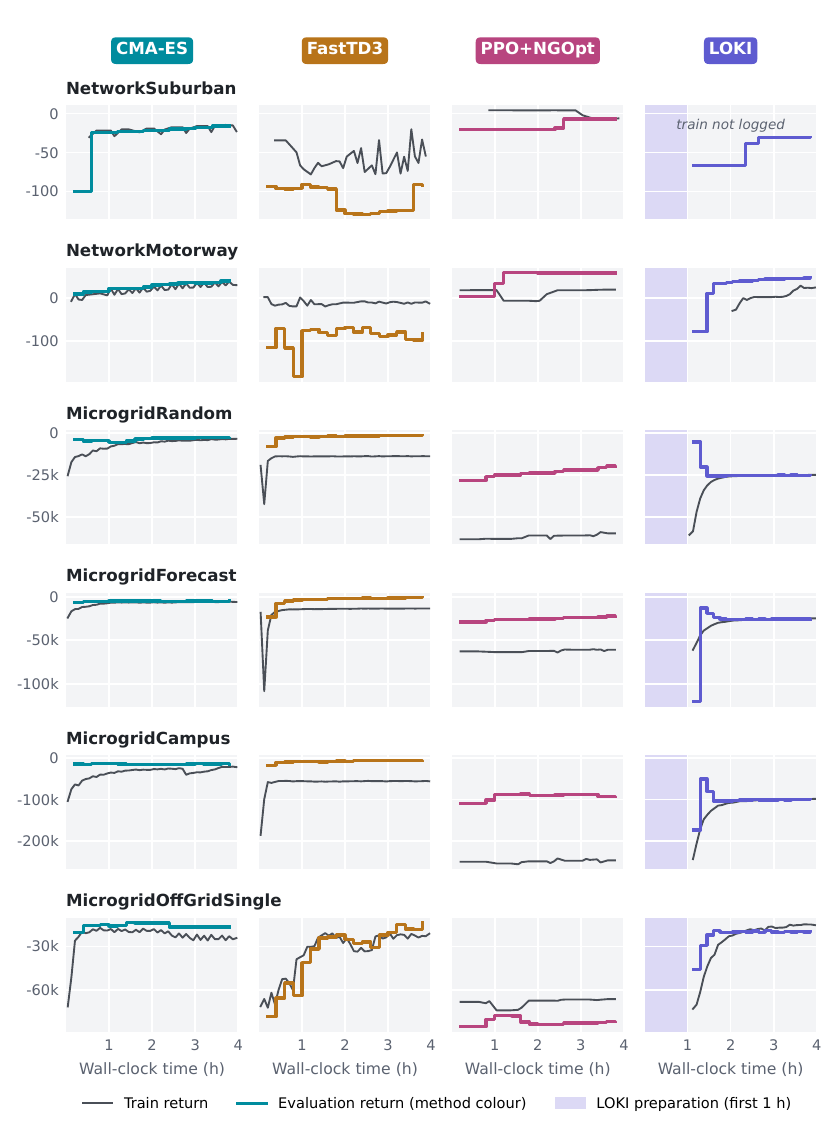}
  \caption{\textbf{Train vs.\ eval return, Reframed presets (2/3)} (continued; see Figure~\ref{fig:train-eval-reframed}). LOKI does not log a train return on NetworkSuburban.}
  \label{fig:train-eval-reframed-2}
\end{figure}

\begin{figure}[p]
  \centering
  \includegraphics[width=\linewidth]{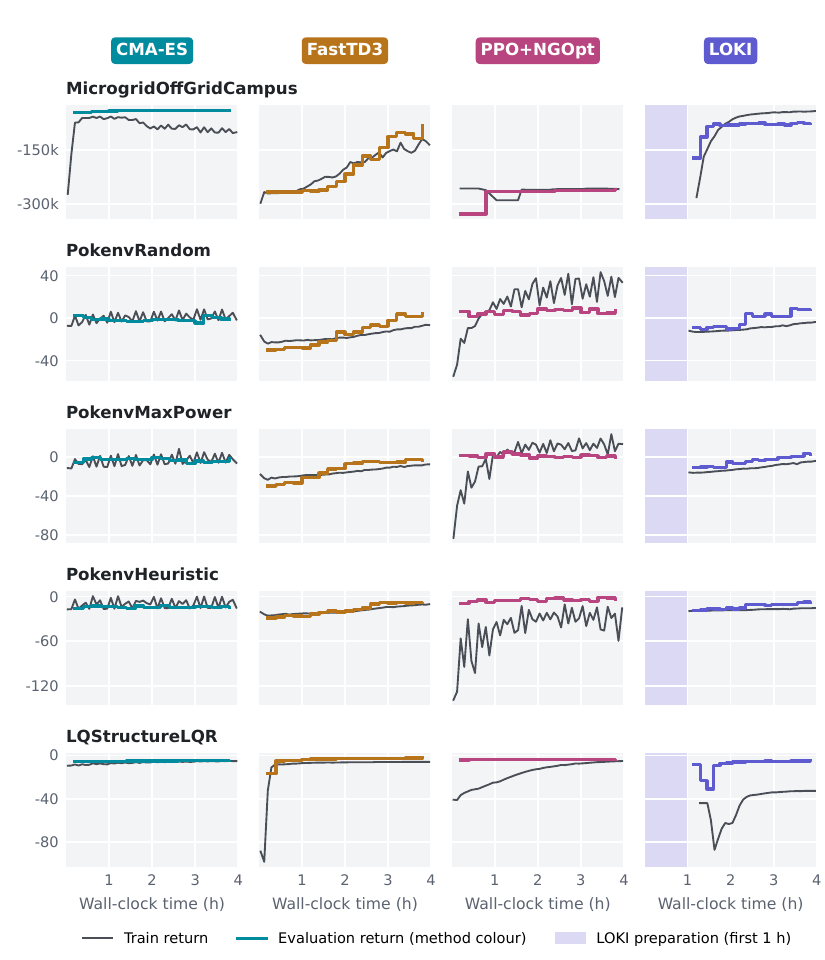}
  \caption{\textbf{Train vs.\ eval return, Reframed presets (3/3)} (continued; see Figure~\ref{fig:train-eval-reframed}).}
  \label{fig:train-eval-reframed-3}
\end{figure}

\begin{figure}[p]
  \centering
  \includegraphics[width=\linewidth]{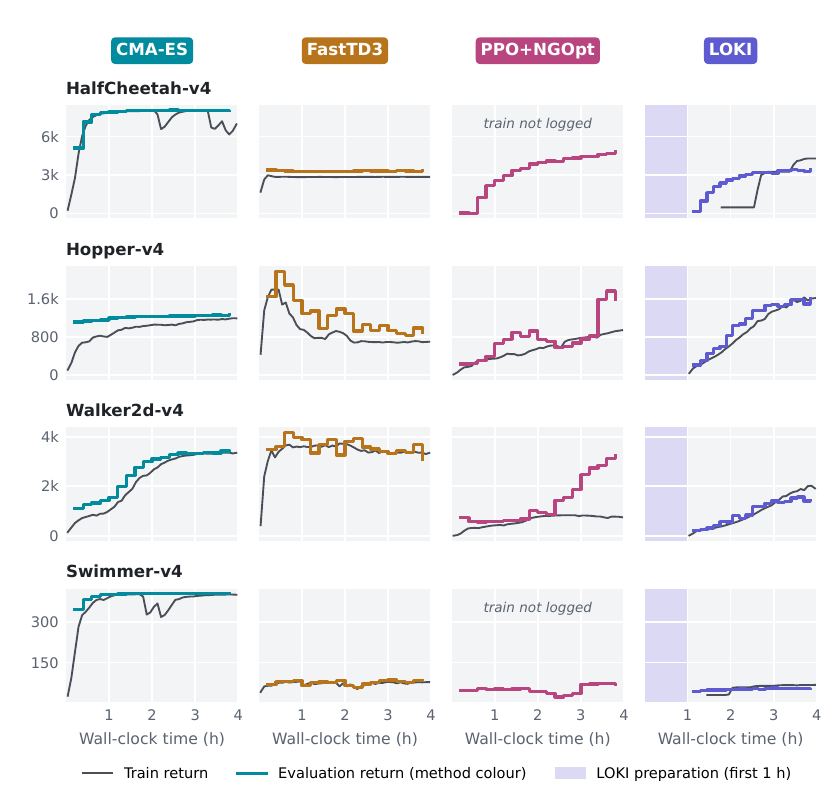}
  \caption{\textbf{Train vs.\ eval return, Extended presets (1/2)} (mean over 5 seeds, each panel sharing its row's y-axis. Train return averages over the designs being trained or explored and eval return uses the selected design (each seed's latest evaluation carried forward), so the gap measures each method's cost of design exploration. LOKI's curves start after its 1-hour preparation (shaded).) PPO+NGOpt does not log a train return on HalfCheetah and Swimmer.}
  \label{fig:train-eval-extended-1}
\end{figure}

\begin{figure}[p]
  \centering
  \includegraphics[width=\linewidth]{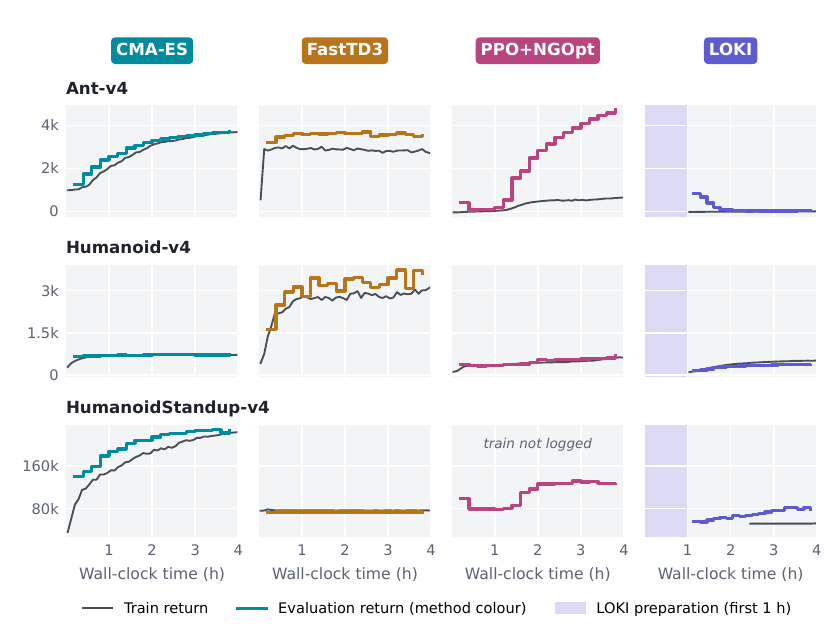}
  \caption{\textbf{Train vs.\ eval return, Extended presets (2/2)} (continued; see Figure~\ref{fig:train-eval-extended-1}). PPO+NGOpt does not log a train return on HumanoidStandup.}
  \label{fig:train-eval-extended}
\end{figure}

Table~\ref{table:compute} summarizes, for each baseline over all 35 presets, the number of environment steps taken within the budget, the resulting throughput, and the number of trainable parameters. Within the same wall-clock budget, the methods differ by more than an order of magnitude in environment interactions (median 1.2 billion steps per run for CMA-ES versus 24 million for LOKI); learning curves over time are shown in Figures~\ref{fig:train-eval-native}--\ref{fig:train-eval-extended}.

\begin{table}[htbp]
\centering
\footnotesize
\setlength{\tabcolsep}{3pt}
\begin{tabular}{@{}llrrrr@{}}
\toprule
& & \multicolumn{2}{c}{Env.\ steps / run (M)} & & \\
\cmidrule(lr){3-4}
Method & Budget & median & range & \shortstack[r]{Steps / s\\(median)} & \shortstack[r]{Parameters\\(range)} \\
\midrule
CMA-ES & 4 h & 1{,}176 & 0.19--10{,}553 & 81{,}666 & 390--32{,}702 \\
FastTD3 & 4 h & 209 & 0.064--2{,}679 & 14{,}530 & 1{,}567{,}056--5{,}047{,}080 \\
PPO+NGOpt & 4 h & 249 & 0.15--6{,}640 & 17{,}301 & 9{,}735--134{,}761 \\
LOKI & 3 h + 1 h prep. & 24 & 0.011--130 & 2{,}197 & 20{,}070{,}140--21{,}252{,}625 \\
\bottomrule
\end{tabular}

\caption{Compute per baseline over the 35 presets of Table~\ref{table:baselines}: environment steps per run (millions; median and range over presets and seeds; for LOKI, summed over all cluster controllers and including candidate scoring and ranking), throughput (environment steps per second, median), and trainable parameters (range over presets; CMA-ES: controller and embodiment parameters; LOKI: all cluster controllers and the design VAE). Parameter counts cover 30 of the 35 presets.}
\label{table:compute}
\end{table}

\begin{figure}[htbp]
  \centering
  \includegraphics[width=\linewidth]{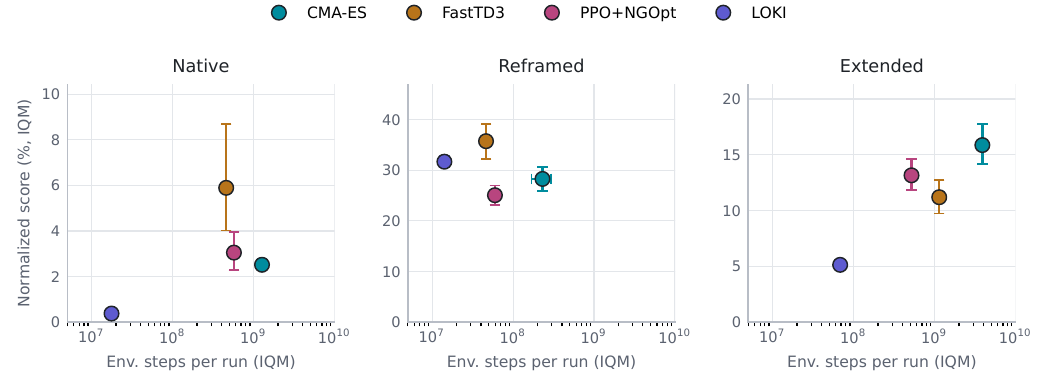}
  \caption{\textbf{Sample efficiency by category.} Interquartile mean (IQM) over runs of the environment steps each run took within its budget (log scale) and of its normalized score; whiskers are stratified bootstrap 95\% confidence intervals.}
  \label{fig:sample-efficiency}
\end{figure}

\textbf{Sample efficiency, design validity and stability.} Tables~\ref{table:efficiency-steps} and~\ref{table:efficiency-validity} break these statistics down per preset: the environment steps each method takes within the budget and the steps it needs to get 90\% of the way from its first to its final evaluation, the fraction of evaluated designs rejected as invalid, and the drop from the best to the final normalized score, which measures how well a method retains its best solution. Figure~\ref{fig:sample-efficiency} relates each method's score to its environment steps by category. How efficiently the baselines turn environment steps into score depends on the domain: FastTD3 leads on the native and reframed presets, CMA-ES leads on the extended locomotion tasks while taking the most steps, and LOKI reaches the second-highest IQM on the reframed presets with the fewest steps. Compared at a matched number of environment steps (the fewest taken by any method on a preset) rather than at the end of the wall-clock budget, the leading method changes on 22 of 35 presets.

\begin{table}[htbp]
\centering
\footnotesize
\setlength{\tabcolsep}{3pt}
\begin{tabular}{@{}lrrrrrrrr@{}}
\toprule
 & \multicolumn{2}{c}{CMA-ES} & \multicolumn{2}{c}{FastTD3} & \multicolumn{2}{c}{PPO+NGOpt} & \multicolumn{2}{c}{LOKI} \\
\cmidrule(lr){2-3}\cmidrule(lr){4-5}\cmidrule(lr){6-7}\cmidrule(lr){8-9}
Preset & Steps & 90\% & Steps & 90\% & Steps & 90\% & Steps & 90\% \\
\midrule
\multicolumn{9}{@{}l}{\textit{Native}} \\
BallCatcher-Pitch & 1{,}898 & 1{,}054 & 828 & 347 & 1{,}201 & 677 & 28 & 14 \\
BallCatcher-Arc & 2{,}466 & 1{,}358 & 813 & 286 & 1{,}123 & 1{,}001 & 28 & 15 \\
BallCatcher-Fielding & 2{,}385 & 1{,}269 & 836 & 233 & 1{,}083 & 636 & 29 & 7.5 \\
NeroGraspAll & 3.2 & 1.2 & 7.3 & 0.28 & 3.6 & 2.6 & 6.6 & 3.9 \\
HandPickUpCube & 1{,}284 & 661 & 642 & 434 & 806 & 467 & 3.6 & 1.9 \\
HandPickUpHammer & 540 & 421 & 728 & 398 & 634 & 106 & 3.9 & 2.6 \\
RotateInHandCube & 413 & 176 & 560 & 413 & 343 & 253 & 3.0 & 1.4 \\
SoftWalkerBeam3D & 0.21 & 0.0096 & 0.068 & 0.010 & 0.15 & 0.010 & 0.013 & 0 \\
SolarCleanerReach & 3{,}556 & 2{,}344 & 184 & 97 & 1{,}331 & 755 & 24 & 16 \\
SolarCleanerTraverse & 1{,}839 & 1{,}110 & 82 & 61 & 10 & 5.4 & 26 & 17 \\
TruckUnloadSingle & 680 & 260 & 230 & 103 & 110 & 91 & 81 & 23 \\
\midrule
\multicolumn{9}{@{}l}{\textit{Reframed}} \\
\multicolumn{9}{@{}l}{Racing} \\
\hspace{0.8em}Spielberg & 6{,}314 & 2{,}250 & 354 & 133 & 6{,}398 & 734 & 63 & 24 \\
\hspace{0.8em}Levine & 3{,}206 & 1{,}964 & 558 & 311 & 5{,}051 & 799 & 75 & 36 \\
\multicolumn{9}{@{}l}{Warehouse} \\
\hspace{0.8em}Small & 8{,}789 & 2{,}245 & 225 & 13 & 654 & 569 & 12 & 10 \\
\hspace{0.8em}Medium & 4{,}769 & 223 & 216 & 19 & 393 & 307 & 5.3 & 0 \\
\hspace{0.8em}Congested & 2{,}554 & 100 & 185 & 56 & 232 & 45 & 2.2 & 0 \\
\multicolumn{9}{@{}l}{Network} \\
\hspace{0.8em}Urban & 1.5 & 0.84 & 1.2 & 0.31 & 1.4 & 0.46 & 2.9 & 1.9 \\
\hspace{0.8em}Suburban & 0.73 & 0.12 & 0.82 & 0.39 & 0.61 & 0.29 & 1.7 & 0.80 \\
\hspace{0.8em}Motorway & 2.4 & 1.9 & 2.0 & 0.44 & 1.4 & 0.39 & 4.5 & 1.4 \\
\multicolumn{9}{@{}l}{Microgrid} \\
\hspace{0.8em}Random & 14 & 6.8 & 13 & 4.4 & 1.4 & 1.1 & 29 & 4.0 \\
\hspace{0.8em}Forecast & 12 & 3.2 & 11 & 3.8 & 1.4 & 1.1 & 19 & 1.6 \\
\hspace{0.8em}Campus & 14 & 4.7 & 13 & 4.1 & 1.4 & 0.53 & 21 & 2.6 \\
\hspace{0.8em}OffGridSingle & 16 & 6.2 & 15 & 5.6 & 1.6 & 0.48 & 25 & 3.5 \\
\hspace{0.8em}OffGridCampus & 17 & 6.3 & 16 & 11 & 1.7 & 0.27 & 24 & 3.5 \\
\multicolumn{9}{@{}l}{Pokenv} \\
\hspace{0.8em}Random & 12 & 3.0 & 7.7 & 5.8 & 5.2 & 0.94 & 7.1 & 4.5 \\
\hspace{0.8em}MaxPower & 13 & 5.1 & 8.0 & 4.1 & 5.0 & 1.2 & 9.1 & 6.2 \\
\hspace{0.8em}Heuristic & 11 & 3.8 & 7.3 & 5.0 & 5.0 & 1.2 & 6.2 & 3.9 \\
LQStructure (LQR) & 382 & 197 & 379 & 95 & 329 & 89 & 44 & 23 \\
\midrule
\multicolumn{9}{@{}l}{\textit{Extended}} \\
HalfCheetah & 10{,}314 & 1{,}715 & 2{,}660 & 542 & 786 & 522 & 79 & 36 \\
Hopper & 4{,}905 & 2{,}481 & 1{,}039 & 317 & 197 & 115 & 103 & 54 \\
Walker2d & 3{,}951 & 1{,}821 & 926 & 372 & 377 & 248 & 80 & 49 \\
Swimmer & 6{,}968 & 1{,}309 & 2{,}270 & 380 & 538 & 288 & 129 & 27 \\
Ant & 2{,}404 & 1{,}577 & 803 & 254 & 768 & 629 & 34 & 8.6 \\
Humanoid & 2{,}034 & 1{,}233 & 725 & 217 & 251 & 219 & 41 & 25 \\
HumanoidStandup & 2{,}002 & 1{,}081 & 1{,}152 & 147 & 649 & 417 & 42 & 28 \\
\bottomrule
\end{tabular}

\caption{Environment interactions and sample efficiency per preset (millions of environment steps, mean over 5 seeds). Steps: total steps within the budget. 90\%: steps at the first evaluation that is 90\% of the way from the seed's first to its final evaluation. Steps of PPO+NGOpt on the Pok\'emon presets are interpolated at a constant rate.}
\label{table:efficiency-steps}
\end{table}

\begin{table}[htbp]
\centering
\footnotesize
\setlength{\tabcolsep}{3pt}
\begin{tabular}{@{}lrrrrrrrr@{}}
\toprule
 & \multicolumn{2}{c}{CMA-ES} & \multicolumn{2}{c}{FastTD3} & \multicolumn{2}{c}{PPO+NGOpt} & \multicolumn{2}{c}{LOKI} \\
\cmidrule(lr){2-3}\cmidrule(lr){4-5}\cmidrule(lr){6-7}\cmidrule(lr){8-9}
Preset & Rej. & Drop & Rej. & Drop & Rej. & Drop & Rej. & Drop \\
\midrule
\multicolumn{9}{@{}l}{\textit{Native}} \\
BallCatcher-Pitch & 0 & 1.4 & 0 & 13 & 0 & 1.5 & 0 & 0.021 \\
BallCatcher-Arc & 0 & 1.1 & 0 & 28 & 0 & 0.54 & 0 & 0.069 \\
BallCatcher-Fielding & 0 & 0.28 & 0 & 19 & 0 & 0.81 & 0 & 0.081 \\
NeroGraspAll & 0 & 0.67 & 100 & 0 & 0 & 0.21 & 0 & 0.043 \\
HandPickUpCube & 0 & 0.043 & 0 & 0.40 & 0 & 0.00056 & 0 & 0.029 \\
HandPickUpHammer & 0 & $1.4\times10^{-15}$ & 0 & 6.6 & 0 & 2.8 & 0 & 0.28 \\
RotateInHandCube & 0 & 0.019 & 0 & 0.052 & 0 & 0.028 & 0 & 0.20 \\
SoftWalkerBeam3D & 0 & 0 & 21 & 0.19 & 0 & 0.0017 & 0 & 0.00070 \\
SolarCleanerReach & 0 & 0.000062 & 0 & 0.012 & 0 & 0.87 & 0 & 0.0038 \\
SolarCleanerTraverse & 0 & 0.00021 & 0 & 0.029 & 0 & 0.035 & 0 & 0.00013 \\
TruckUnloadSingle & 0 & 0.0033 & 0 & 7.8 & 0 & 0 & 0 & 0.044 \\
\midrule
\multicolumn{9}{@{}l}{\textit{Reframed}} \\
\multicolumn{9}{@{}l}{Racing} \\
\hspace{0.8em}Spielberg & -- & 0.35 & -- & 7.7 & -- & 42 & 0 & 0.27 \\
\hspace{0.8em}Levine & -- & 0.63 & -- & 9.6 & 0 & 37 & 0 & 1.1 \\
\multicolumn{9}{@{}l}{Warehouse} \\
\hspace{0.8em}Small & 0 & 0 & 0 & 0 & 0 & 0 & 0 & 0.00047 \\
\hspace{0.8em}Medium & 0 & 0 & 0 & 0 & 0 & 0.00016 & 0 & 0.00047 \\
\hspace{0.8em}Congested & 0 & 0 & 43 & 0 & 0 & 0.00063 & 0 & 0.00016 \\
\multicolumn{9}{@{}l}{Network} \\
\hspace{0.8em}Urban & 0 & 0.83 & 0 & 0.34 & 0 & 2.1 & 0 & 0.094 \\
\hspace{0.8em}Suburban & 0 & 0.14 & 9.4 & 7.9 & 0 & 0 & 0 & $1.4\times10^{-15}$ \\
\hspace{0.8em}Motorway & 0 & 0.37 & 5.9 & 55 & 0 & 0.79 & 0 & 0 \\
\multicolumn{9}{@{}l}{Microgrid} \\
\hspace{0.8em}Random & 0 & 0.56 & 0 & 0.29 & 0 & 0.029 & 0 & 37 \\
\hspace{0.8em}Forecast & 0 & 0.59 & 0 & 0.068 & 0 & 0.0081 & 0 & 23 \\
\hspace{0.8em}Campus & 0 & 1.1 & 0 & 0.62 & 0 & 3.9 & 0 & 25 \\
\hspace{0.8em}OffGridSingle & 0 & 4.3 & 0 & 10 & 0 & 9.6 & 0 & 2.7 \\
\hspace{0.8em}OffGridCampus & 0 & 0.0059 & 0 & 8.7 & 0 & 0.000021 & 0 & 2.5 \\
\multicolumn{9}{@{}l}{Pokenv} \\
\hspace{0.8em}Random & -- & 20 & 53 & 8.8 & -- & 17 & 0 & 5.2 \\
\hspace{0.8em}MaxPower & -- & 10 & 36 & 11 & -- & 20 & 0 & 3.9 \\
\hspace{0.8em}Heuristic & -- & 15 & 40 & 7.2 & -- & 12 & 0 & 7.8 \\
LQStructure (LQR) & 0 & 0.0091 & 0 & 0.11 & 0 & 0.13 & 0 & 0.35 \\
\midrule
\multicolumn{9}{@{}l}{\textit{Extended}} \\
HalfCheetah & 0 & 0.16 & 0 & 0.75 & 0 & 0 & 0 & 0.33 \\
Hopper & 0 & 0.030 & 0 & 12 & 0 & 4.2 & 0 & 0.81 \\
Walker2d & 0 & 0.16 & 0 & 6.9 & 0 & 0.72 & 0 & 1.2 \\
Swimmer & 0 & 0.00029 & 0 & 0.13 & 0 & 0.34 & 0 & 0.13 \\
Ant & 0 & 0.026 & 0 & 1.5 & 0 & 0 & 0 & 3.8 \\
Humanoid & 0 & 0.031 & 0 & 2.5 & 0 & 0.054 & 0 & 0.048 \\
HumanoidStandup & 0 & 0.48 & 0 & 0.26 & 0 & 1.2 & 0 & 0.88 \\
\bottomrule
\end{tabular}

\caption{Design validity and stability per preset (mean over 5 seeds). Rej.: percentage of evaluated designs rejected as invalid (--: not logged). Drop: best minus final normalized score (points); larger values mean a method found but did not retain a better solution.}
\label{table:efficiency-validity}
\end{table}

\subsection{LOKI with 40 clusters}
\label{sec:loki40}

Our LOKI baseline uses 6 clusters, whereas the original method uses 40~\citep{jeon2026convergent}. To test whether this choice limits LOKI, we reran it with 40 clusters and 40 cluster controllers on two presets at opposite ends of its results: MicrogridRandom, where it achieves its highest fraction of the theoretical maximum return, and SolarCleanerReach, where it achieves one of its lowest. The 40-cluster K-means is fit on the same VAE latents and seeds as the 6-cluster version; every other setting is unchanged (5 seeds, 3 hours online, 16{,}384 training worlds, now split into 409--410 worlds per cluster). With the same budget spread over 40 controllers, each controller receives roughly 7$\times$ fewer PPO updates. Table~\ref{table:loki40} shows that 40 clusters leave MicrogridRandom unchanged and raise the raw return on SolarCleanerReach about 2.5-fold, while both remain within 0.3 points of the 6-cluster normalized score, so the cluster count does not change LOKI's standing on either preset.

\begin{table}[htbp]
\centering
\footnotesize
\begin{tabular}{@{}lrrrr@{}}
\toprule
& \multicolumn{2}{c}{\textbf{LOKI, 6 clusters}} & \multicolumn{2}{c}{\textbf{LOKI, 40 clusters}} \\
\textbf{Preset} & Raw & Norm.\ (\%) & Raw & Norm.\ (\%) \\
\midrule
MicrogridRandom & $-$25{,}107 $\pm$ 130 & 52.1 $\pm$ 0.2 & $-$25{,}262 $\pm$ 202 & 51.8 $\pm$ 0.4 \\
SolarCleanerReach & 0.00613 $\pm$ 0.00207 & 0.07 $\pm$ 0.02 & 0.0155 $\pm$ 0.0007 & 0.17 $\pm$ 0.01 \\
\bottomrule
\end{tabular}
\caption{LOKI with 6 versus 40 clusters on its best and worst preset (final evaluation return and normalized score, mean $\pm$ std over 5 seeds; 3 hours online after 1 hour of preparation).}
\label{table:loki40}
\end{table}

\subsection{Learning curves and final designs}

\begin{figure}[htbp]
  \centering
  \includegraphics[width=\linewidth]{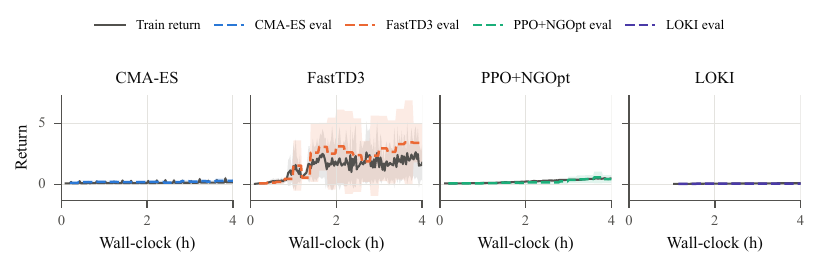}
  \caption{\textbf{BallCatcher-Pitch.} Only FastTD3, which emits the design as the first action of each episode, learns to catch
  (16.6\% full catches vs.\ $\leq$1.1\% for the others; final return 3.42 vs.\ $\leq$0.44). Its final return varies widely across seeds (1.35--5.98), so the
  joint design-and-policy agent is effective here but unreliable. None of the outer-loop methods reaches catching behaviour within the budget.}
  \label{fig:highlight-ballcatcher-pitch}
\end{figure}

\begin{figure}[htbp]
  \centering
  \includegraphics[width=\linewidth]{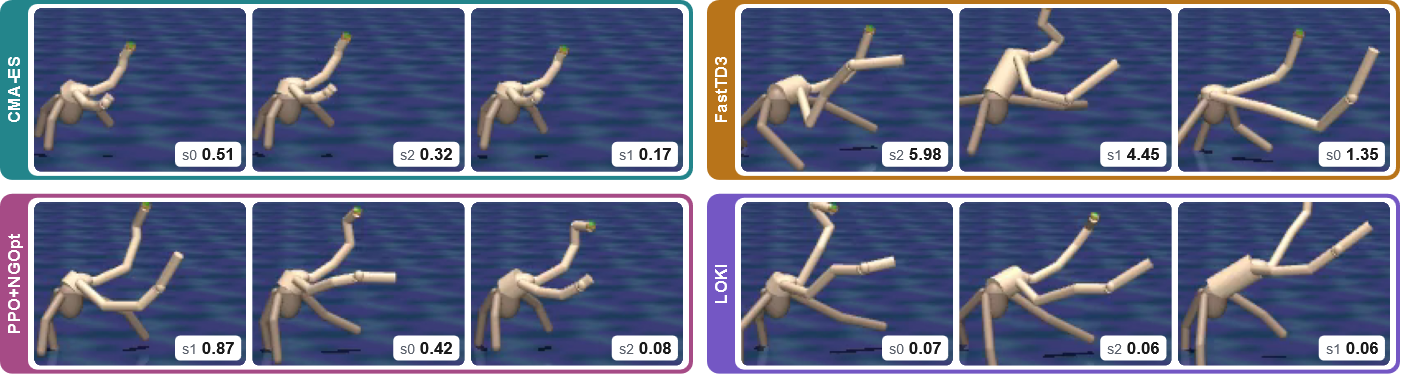}
  \caption{\textbf{Final designs on BallCatcher-Pitch}, one per run, taken from the first frame of the run's final evaluation rollout and
  shown at a common scale; seeds are sorted by final eval return (above each image) and the best is framed thick. The methods converge to
  visibly different bodies, and FastTD3 and LOKI also differ from seed to seed, whereas CMA-ES's designs look alike: similar returns can
  come from different morphologies.}
  \label{fig:designs-ballcatcher-pitch}
\end{figure}

\begin{figure}[htbp]
\vspace{-1.5em}
  \centering
  \includegraphics[width=\linewidth]{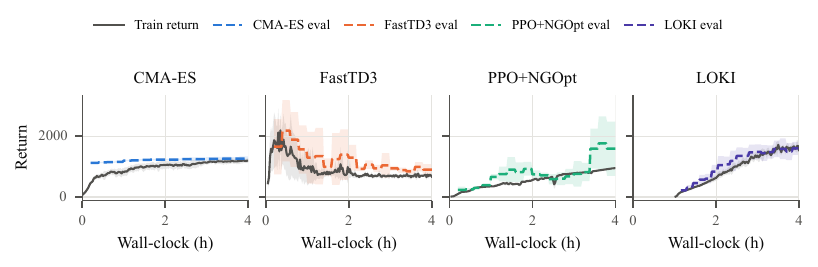}
  \caption{\textbf{Hopper.} The four families trace four different learning curves on the same task, and the leader changes six times in
  four hours. FastTD3 peaks early (2{,}800 at 0.6\,h) and decays to 897; PPO+NGOpt jumps late (761$\rightarrow$1{,}588 in the final hour,
  seeds 462--2{,}885); LOKI improves steadily throughout; CMA-ES plateaus at $\approx$1{,}260 almost immediately. A single
  budget checkpoint would rank the methods differently at almost every time.}
  \label{fig:highlight-hopper}
\end{figure}

\begin{figure}[htbp]
  \centering
  \includegraphics[width=\linewidth]{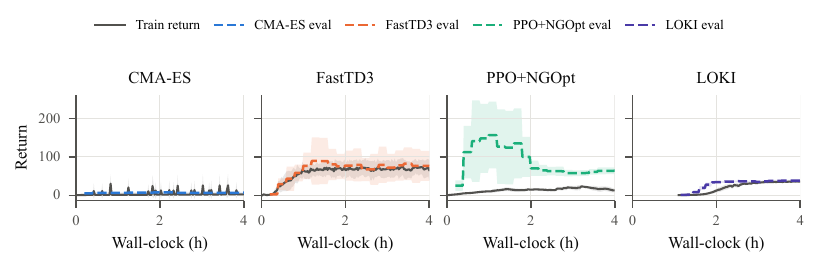}
  \caption{\textbf{RacingSpielberg.} PPO+NGOpt finds a strong vehicle setup within the first hour (eval 157 at 1\,h) and then loses it,
  ending at 64 after $\approx$245 design rounds: more outer-loop design updates are not monotonically better, and a good design found
  early is not retained. Its train return ($\approx$13) stays far below eval because most rounds
  train on exploratory setups. FastTD3 is stable at $\approx$77; CMA-ES never leaves $\approx$7.}
  \label{fig:highlight-racingspielberg}
\end{figure}

\begin{figure}[htbp]
  \centering
  \includegraphics[width=\linewidth]{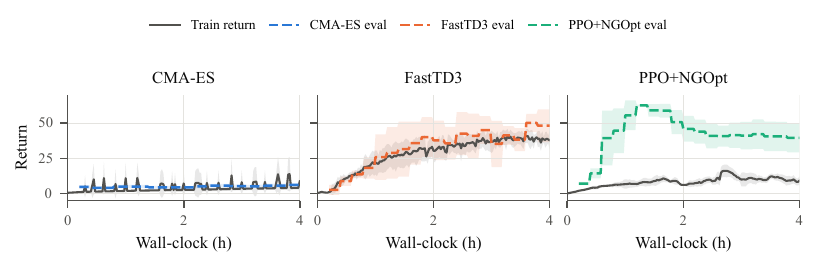}
  \caption{\textbf{RacingLevine.} The same pattern as on RacingSpielberg on a second track: PPO+NGOpt peaks within the first hour (63) and
  declines to 40 over $\approx$274 design rounds, while FastTD3 improves steadily, overtakes it and ends with the most completed laps
  (37\% vs.\ 20\%). Early design-search gains that are not retained recur across tracks, so they reflect the algorithm rather than
  a single environment.}
  \label{fig:highlight-racinglevine}
\end{figure}

\clearpage

\clearpage
\section{LLM Co-Design}
\label{appx:llm-ppo}

In this experiment, we use Gemma 4 31B as the design optimizer in a PPO+LLM co-design loop on the BallCatcher-Pitch preset. The algorithm mirrors the PPO+NGOpt baseline (Section~\ref{sec:expr}), with the LLM taking the place of NGOpt as the outer-loop design optimizer. The PPO controller is conditioned on the parameters of the embodiment it controls.

Placeholders of the form \texttt{\{VAR\}} are replaced with the corresponding values at query time.

\begin{figure}[p]
\vspace{-1em}
\begin{promptbox}
\footnotesize
\lstset{style=jsonstyle, basicstyle=\ttfamily\scriptsize, aboveskip=3pt, belowskip=3pt}

\noindent\textbf{SYSTEM INSTRUCTION:}
\noindent Propose exactly \texttt{\{NUM\_TEMPLATES\}} embodiment templates in one response. Return only a JSON object containing an \texttt{embodiments} array with exactly \texttt{\{NUM\_TEMPLATES\}} lists. Each list must contain exactly \texttt{\{EMBODIMENT\_DIM\}} numeric values in parameter-index order. Maximize the mean episodic return across random perturbations of each proposed template. Respect all parameter bounds and integer constraints. Copies of each template receive independent Gaussian perturbations scaled by the corresponding parameter ranges. Perturbed values are clipped to their bounds and integer-valued parameters are rounded. Use the supplied template-statistics history to explore the design space and improve robust designs. The reported statistics describe perturbed copies rather than the exact templates. Rejected copies receive the rejection penalty and are included when computing mean reward. The PPO controller changes after every round, so account for the round labels when comparing results. Return no code, explanation, or text outside the required JSON object.

\noindent\textbf{INPUT:} The user message is a JSON object with the following structure:
\begin{lstlisting}
{
  "task": "BallCatcher-Pitch",
  "description": {TASK_DESCRIPTION},
  "task_context": {
    "parameter_defaults": {PARAMETER_DEFAULTS},
    "parameter_columns": {PARAMETER_COLUMNS},
    "parameter_rows": {PARAMETER_ROWS},
    "parameter_encoding": {PARAMETER_ENCODING}
  },
  "round": {ROUND},
  "search_state": {
    "completed_rounds": {COMPLETED_ROUNDS},
    "completed_ppo_iterations": {COMPLETED_PPO_ITERATIONS},
    "population_size": {POPULATION_SIZE},
    "template_count": {NUM_TEMPLATES},
    "copies_per_template": {COPIES_PER_TEMPLATE},
    "remaining_seconds": {REMAINING_SECONDS},
    "best_observed_design": {BEST_EMBODIMENT},
    "best_observed_reward": {BEST_REWARD}
  },
  "base_embodiment": {BASE_EMBODIMENT},
  "perturbation_std_fraction": {PERTURBATION_STD},
  "template_statistics_columns": {STATISTICS_COLUMNS},
  "template_statistics_rows": {STATISTICS_ROWS},
  "history_embodiment_encoding": "Each historical embodiment cell contains
      [parameter_index, value] pairs; omitted indices equal base_embodiment.",
  "diagnostic_definitions": "Rates are fractions of evaluated copies with the
      event at least once during their first episode; rejected copies contribute
      zero events and zero length. catch_rate requires accumulated dwell and
      retention; goal_entry_rate includes brief contact with the reward region.
      Missing signals are omitted."
}
\end{lstlisting}

\noindent\textbf{TASK:} Propose \texttt{\{NUM\_TEMPLATES\}} new embodiment templates that are likely to maximize mean episodic return after independent perturbation. Use the task description, embodiment parameter semantics, current PPO training state, best result observed so far, historical template statistics, rejection statistics, and available episode diagnostics.

\smallskip
\noindent\textbf{OUTPUT REQUIREMENTS:} Return JSON only: no extra text, code fences, comments, or reasoning. Return one JSON object matching this schema:
\begin{lstlisting}
{
  "embodiments": [
    [<exactly {EMBODIMENT_DIM} numeric values>],
    ...
    [<exactly {EMBODIMENT_DIM} numeric values>]
  ]
}
\end{lstlisting}

\noindent\textbf{IMPORTANT INFORMATION:}
\begin{itemize}\setlength{\itemsep}{0pt}\setlength{\parskip}{0pt}
    \item  The output must contain exactly \texttt{\{NUM\_TEMPLATES\}} embodiment templates.
    \item Every template must contain exactly \texttt{\{EMBODIMENT\_DIM\}} values in parameter-index order.
    \item Every value must satisfy its supplied bounds and integer constraint.
    \item The optimization target is mean episodic return across the perturbed copies of each template.
    \item Rejected embodiments receive the rejection penalty and reduce the template's mean reward.
    \item Historical rewards come from different PPO training rounds; their round labels must therefore be considered.
    \item Only valid JSON matching the specified schema is accepted.
\end{itemize}
\end{promptbox}
\vspace{-1em}
\caption{Prompt used by the PPO+LLM outer loop on BallCatcher-Pitch. Placeholders of the form \texttt{\{VAR\}} are replaced with the corresponding values at query time.}
\label{fig:llm-prompt}
\end{figure}

This prompt conditions the LLM on far more information than any other baseline receives. It conveys details of the environment that the other baselines can only learn through interaction, and it provides context on each embodiment parameter, such that a model could anticipate the relationship between provided embodiment parameters and design. 

To ensure a fair comparison, PPO+LLM must train on as many embodiments per design round as the other baselines. Otherwise, any difference in performance could simply reflect a narrower exploration of the design space. PPO+NGOpt, its direct counterpart, samples one embodiment for each of the 16{,}384 parallel environments. Having the LLM generate that many designs directly, however, is infeasible under API rate limits and token costs. We therefore split design generation into two stages. In each design round, the LLM proposes 16 \emph{template} embodiments, and each template is perturbed with Gaussian noise (standard deviation 0.05) to yield 1{,}024 variants, giving 16{,}384 embodiments in total. The prompt informs the LLM of this perturbation scheme. For evaluation runs, both methods evaluate the single best-scoring design found so far.

\paragraph{Results.} Despite the informational advantage granted to the agentic designer, PPO+LLM performs poorly in all runs, reaching low final return and showing little learning progress. The embodiments proposed by the LLM were often impractical and over-engineered, such that, even when the controller did learn meaningful behaviour, the given design stalled learning progress.

\begin{figure}[h!]
    \vspace{-1.8em}
  \centering
  \includegraphics[width=\linewidth]{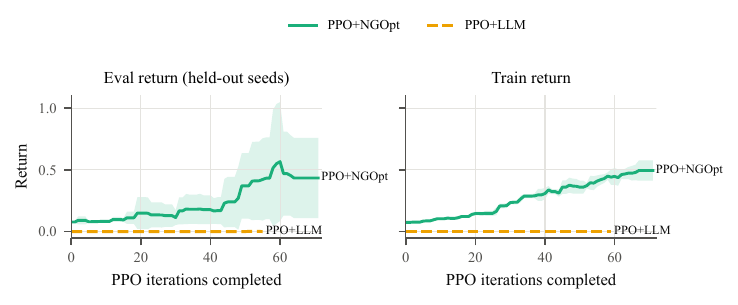}
  \vspace{-2.5em}
  \caption{\textbf{PPO+LLM vs.\ PPO+NGOpt on BallCatcher-Pitch}, plotted against PPO iterations because both algorithms run the same PPO
  inner loop. PPO+NGOpt's return (mean $\pm$ std) rises from $\approx$0.08 to
  0.43 at evaluation, whereas PPO+LLM stays at or near zero in training ($\leq$0.005) and at every evaluation ($\leq$0.02), over a
  comparable number of PPO iterations.}
  \vspace{-1.5em}
  \label{fig:llm-train-eval}
\end{figure}

\begin{figure}[htbp]
\vspace{-0em}
  \centering
  \includegraphics[width=\linewidth]{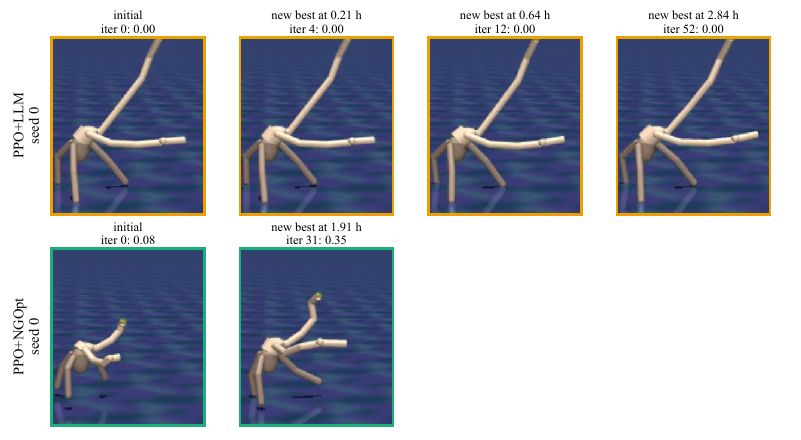}
  \vspace{-2em}
  \caption{\textbf{Every distinct evaluated design} (seed 0, illustrative) for PPO+LLM (amber frames) and PPO+NGOpt (aqua frames): the
  initial design and the design after each time a better-scoring candidate replaced it (PPO+LLM at 0.21, 0.64 and 2.84\,h; PPO+NGOpt once,
  at 1.91\,h), each taken from the first frame of the next evaluation video and shown at a common scale (title: PPO iteration and eval
  return of that evaluation).}
  \vspace{-2em}
  \label{fig:llm-designs}
\end{figure}

Figure~\ref{fig:llm-train-eval} compares training and evaluation return against the number of PPO iterations. Over a comparable number of iterations (68 and 59 for PPO+LLM, 68--71 for PPO+NGOpt), PPO+NGOpt improves from $\approx$0.08 to 0.43 eval return, while both PPO+LLM runs stay at or near zero throughout training and at every evaluation. Between 27\% and 36\% of LLM-designs generated were unusable. The best design was updated 3 and 4 times in the two PPO+LLM runs, and between 1 and 15 times across the PPO+NGOpt seeds. Figure~\ref{fig:llm-designs} shows every distinct evaluated design for seed 0 of each method.

\paragraph{Scope.} This is a deliberately low-budget, exploratory study: a single open-weight model (Gemma 4 31B), queried through Google's free API tier, with a single prompt design. It shows what an LLM designer achieves under the modest budgets that we aim for the whole benchmark to remain accessible to, and does not support conclusions about LLM-based design optimization in general.

\begin{table}[htbp]
\centering
\footnotesize
\setlength{\tabcolsep}{3pt}
\resizebox{\ifdim\width>\linewidth\linewidth\else\width\fi}{!}{%
\begin{tabular}{@{}llrrrr@{}}
\toprule
Run & Stopped at & PPO iters & Design rounds & Final eval & Peak eval \\
\midrule
PPO+LLM & 4 h & 64 & 65 & 0.01 $\pm$ 0.01 & 0.01 $\pm$ 0.01 \\
PPO+NGOpt & 4 h & 70 & 69 & 0.43 $\pm$ 0.33 & 0.59 $\pm$ 0.46 \\
\bottomrule
\end{tabular}}
\caption{PPO+LLM (Gemma 4 31B, 16 templates per round, perturbation std 0.05, 16{,}384 environments) on BallCatcher-Pitch, against the PPO+NGOpt baseline. }
\label{table:ppo-llm}
\end{table}

\clearpage
\section{API Illustration}

\begin{figure}[h!]
    \centering
    \includegraphics[width=0.7\linewidth]{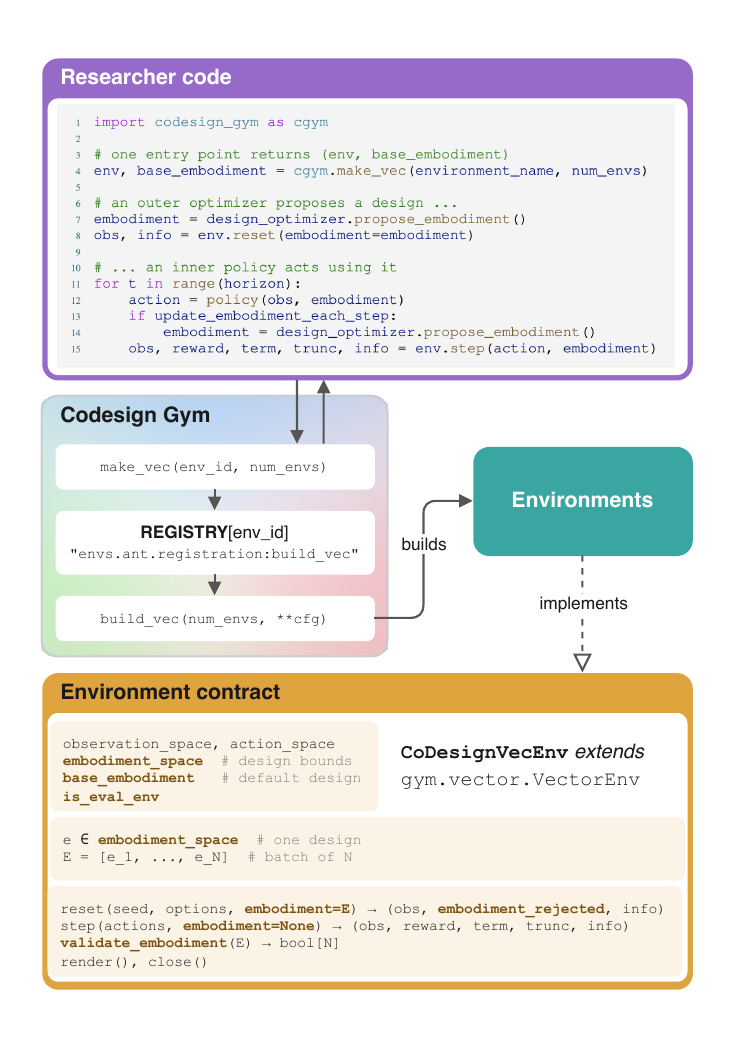}
    \caption{Co-design Gym API flowchart}
    \label{fig:api-detail}
\end{figure}

Figure~\ref{fig:api-detail} illustrates the inner workings of our API: It functions as an interface between co-design algorithm and the environments while minimizing deviations to the established Gymnasium structure, thereby ensuring broad compatibility across established frameworks.

\textbf{Two-level and single-level co-design.} The API does not prescribe how design and control optimization are organized. It supports two-level methods, in which an outer design optimizer proposes embodiments and an inner learner trains a controller for them, as in Evolution Gym~\citep{Bhatia2022EvolutionGA}, and single-level methods, in which one agent or optimizer chooses both the embodiment and the control, as in Transform2Act~\citep{yuan2021transform2act}. Our baselines cover both: PPO+NGOpt (and the PPO+LLM study) is two-level, while CMA-ES, LA-MCTS and SPSA search one joint vector of design and controller parameters, and FastTD3 emits the design as the first action of each episode (Appendix~\ref{sec:baselines}).

\newpage

\bibliography{ref}

\begin{thebibliography}{72}
\providecommand{\natexlab}[1]{#1}
\providecommand{\url}[1]{\texttt{#1}}
\expandafter\ifx\csname urlstyle\endcsname\relax
  \providecommand{\doi}[1]{doi: #1}\else
  \providecommand{\doi}{doi: \begingroup \urlstyle{rm}\Url}\fi

\bibitem[Balciunaite et~al.(2026)Balciunaite, Michelis, Filippi, Paniagua,
  Yasa, and Katzschmann]{balciunaite2026co}
Aiste Balciunaite, Mike~Y Michelis, Miriam Filippi, Pablo Paniagua, Oncay Yasa,
  and Robert~K Katzschmann.
\newblock Co-optimized volumetric muscle design for large dynamic deformations
  in versatile biohybrid robots.
\newblock \emph{Nature Communications}, 2026.

\bibitem[Berner et~al.(2019)Berner, Brockman, Chan, Cheung, Debiak, Dennison,
  Farhi, Fischer, Hashme, Hesse, J{\'o}zefowicz, Gray, Olsson, Pachocki,
  Petrov, de~Oliveira~Pinto, Raiman, Salimans, Schlatter, Schneider, Sidor,
  Sutskever, Tang, Wolski, and Zhang]{Berner2019Dota2W}
Christopher Berner, Greg Brockman, Brooke Chan, Vicki Cheung, Przemyslaw
  Debiak, Christy Dennison, David Farhi, Quirin Fischer, Shariq Hashme,
  Christopher Hesse, Rafal J{\'o}zefowicz, Scott Gray, Catherine Olsson,
  Jakub~W. Pachocki, Michael Petrov, Henrique~Pond{\'e} de~Oliveira~Pinto,
  Jonathan Raiman, Tim Salimans, Jeremy Schlatter, Jonas Schneider, Szymon
  Sidor, Ilya Sutskever, Jie Tang, Filip Wolski, and Susan Zhang.
\newblock Dota 2 with large scale deep reinforcement learning.
\newblock \emph{ArXiv}, abs/1912.06680, 2019.
\newblock URL \url{https://api.semanticscholar.org/CorpusID:209376771}.

\bibitem[Bhatia et~al.(2022)Bhatia, Jackson, Tian, Xu, and
  Matusik]{Bhatia2022EvolutionGA}
Jagdeep Bhatia, Holly Jackson, Yunsheng Tian, Jie Xu, and Wojciech Matusik.
\newblock Evolution gym: A large-scale benchmark for evolving soft robots.
\newblock In \emph{Neural Information Processing Systems}, 2022.
\newblock URL \url{https://api.semanticscholar.org/CorpusID:245011343}.

\bibitem[Bongard(2011)]{bongard2011morphological}
Josh Bongard.
\newblock Morphological change in machines accelerates the evolution of robust
  behavior.
\newblock \emph{Proceedings of the National Academy of Sciences}, 108\penalty0
  (4):\penalty0 1234--1239, 2011.

\bibitem[Bonnet et~al.(2024)Bonnet, Luo, Byrne, Surana, Abramowitz, Duckworth,
  Coyette, Midgley, Tegegn, Kalloniatis, Mahjoub, Macfarlane, Smit, Grinsztajn,
  Boige, Waters, Mimouni, Sob, de~Kock, Singh, Furelos-Blanco, Le, Pretorius,
  and Laterre]{bonnet2024jumanji}
Clément Bonnet, Daniel Luo, Donal Byrne, Shikha Surana, Sasha Abramowitz, Paul
  Duckworth, Vincent Coyette, Laurence~I. Midgley, Elshadai Tegegn, Tristan
  Kalloniatis, Omayma Mahjoub, Matthew Macfarlane, Andries~P. Smit, Nathan
  Grinsztajn, Raphael Boige, Cemlyn~N. Waters, Mohamed~A. Mimouni, Ulrich
  A.~Mbou Sob, Ruan de~Kock, Siddarth Singh, Daniel Furelos-Blanco, Victor Le,
  Arnu Pretorius, and Alexandre Laterre.
\newblock Jumanji: a diverse suite of scalable reinforcement learning
  environments in jax, 2024.
\newblock URL \url{https://arxiv.org/abs/2306.09884}.

\bibitem[Brockman et~al.(2016)Brockman, Cheung, Pettersson, Schneider,
  Schulman, Tang, and Zaremba]{brockman2016openai}
Greg Brockman, Vicki Cheung, Ludwig Pettersson, Jonas Schneider, John Schulman,
  Jie Tang, and Wojciech Zaremba.
\newblock Openai gym.
\newblock \emph{arXiv preprint arXiv:1606.01540}, 2016.

\bibitem[Censi(2015)]{censi2015mathematical}
Andrea Censi.
\newblock A mathematical theory of co-design.
\newblock \emph{arXiv preprint arXiv:1512.08055}, 2015.

\bibitem[Chen et~al.(2026)Chen, Xu, Wang, Wang, and Huang]{chen2026natural}
Changhe Chen, Xiaohao Xu, Xiangdong Wang, Jiaqi Wang, and Xiaonan Huang.
\newblock Natural selection via foundation models for soft robot evolution.
\newblock In \emph{2026 IEEE 9th International Conference on Soft Robotics
  (RoboSoft)}, pp.\  422--429. IEEE, 2026.

\bibitem[Cheney et~al.(2014{\natexlab{a}})Cheney, Clune, and
  Lipson]{cheney2014evolved}
Nicholas Cheney, Jeff Clune, and Hod Lipson.
\newblock Evolved electrophysiological soft robots.
\newblock In \emph{Artificial life conference proceedings}, pp.\  222--229. MIT
  Press, 2014{\natexlab{a}}.

\bibitem[Cheney et~al.(2014{\natexlab{b}})Cheney, MacCurdy, Clune, and
  Lipson]{cheney2014unshackling}
Nick Cheney, Robert MacCurdy, Jeff Clune, and Hod Lipson.
\newblock Unshackling evolution: evolving soft robots with multiple materials
  and a powerful generative encoding.
\newblock \emph{ACM SIGEVOlution}, 7\penalty0 (1):\penalty0 11--23,
  2014{\natexlab{b}}.

\bibitem[Christianos(2020)]{semitable_multi_robot_warehouse_rware_2020}
Filippos Christianos.
\newblock Multi-robot warehouse (rware), 2020.
\newblock URL \url{https://github.com/semitable/robotic-warehouse}.

\bibitem[Coros et~al.(2013)Coros, Thomaszewski, Noris, Sueda, Forberg, Sumner,
  Matusik, and Bickel]{coros2013computational}
Stelian Coros, Bernhard Thomaszewski, Gioacchino Noris, Shinjiro Sueda, Moira
  Forberg, Robert~W Sumner, Wojciech Matusik, and Bernd Bickel.
\newblock Computational design of mechanical characters.
\newblock \emph{ACM Transactions on Graphics (TOG)}, 32\penalty0 (4):\penalty0
  1--12, 2013.

\bibitem[Corucci et~al.(2016)Corucci, Cheney, Lipson, Laschi, and
  Bongard]{corucci2016evolving}
Francesco Corucci, Nick Cheney, Hod Lipson, Cecilia Laschi, and Josh Bongard.
\newblock Evolving swimming soft-bodied creatures.
\newblock In \emph{ALIFE XV, the fifteenth international conference on the
  synthesis and simulation of living systems, late breaking proceedings},
  volume~6, 2016.

\bibitem[Dai et~al.(2026)Dai, Wang, Ashley, and
  Schmidhuber]{dai2026stackelbergppo}
Yanning Dai, Yuhui Wang, Dylan~R. Ashley, and Jürgen Schmidhuber.
\newblock Efficient morphology--control co-design via stackelberg proximal
  policy optimization.
\newblock In \emph{International Conference on Learning Representations
  (ICLR)}, 2026.

\bibitem[De~Michell \& Gupta(1997)De~Michell and Gupta]{de1997hardware}
Giovanni De~Michell and Rajesh~K Gupta.
\newblock Hardware/software co-design.
\newblock \emph{Proceedings of the IEEE}, 85\penalty0 (3):\penalty0 349--365,
  1997.

\bibitem[DeepMind(2025)]{google-deepmind_mujoco_warp_2025}
Google DeepMind.
\newblock mujocowarp, 2025.
\newblock URL \url{https://github.com/google-deepmind/mujoco_warp}.
\newblock GitHub repository. GPU-optimized version of the MuJoCo physics
  simulator, designed for NVIDIA hardware.. Accessed 2026-09-12.

\bibitem[Degrave et~al.(2022)Degrave, Felici, Buchli, Neunert, Tracey,
  Carpanese, Ewalds, Hafner, Abdolmaleki, de~Las~Casas, Donner, Fritz,
  Galperti, Huber, Keeling, Tsimpoukelli, Kay, Merle, Moret, Noury, Pesamosca,
  Pfau, Sauter, Sommariva, Coda, Duval, Fasoli, Kohli, Kavukcuoglu, Hassabis,
  and Riedmiller]{Degrave2022MagneticCO}
Jonas Degrave, Federico Felici, Jonas Buchli, Michael Neunert, Brendan~D.
  Tracey, Francesco Carpanese, Timo Ewalds, Roland Hafner, Abbas Abdolmaleki,
  Diego de~Las~Casas, Craig Donner, Leslie Fritz, Cristian Galperti, Andrea
  Huber, James Keeling, Maria Tsimpoukelli, Jackie Kay, Antoine Merle, J-M.
  Moret, Seb Noury, Federico Pesamosca, D.~Pfau, Olivier Sauter, Cristian
  Sommariva, Stefano Coda, B.~P. Duval, Ambrogio Fasoli, Pushmeet Kohli, Koray
  Kavukcuoglu, Demis Hassabis, and Martin~A. Riedmiller.
\newblock Magnetic control of tokamak plasmas through deep reinforcement
  learning.
\newblock \emph{Nature}, 602:\penalty0 414 -- 419, 2022.
\newblock URL \url{https://api.semanticscholar.org/CorpusID:246904229}.

\bibitem[Dowski~Jr \& Cathey(1995)Dowski~Jr and Cathey]{dowski1995extended}
Edward~R Dowski~Jr and W~Thomas Cathey.
\newblock Extended depth of field through wave-front coding.
\newblock \emph{Applied optics}, 34\penalty0 (11):\penalty0 1859--1866, 1995.

\bibitem[Fay et~al.(2026)Fay, Djapri, Zorin, Clinton, Lahib, Su, Tolley, Yi,
  and Wang]{Fay2025HouseOD}
Kehlani Fay, Darin~Anthony Djapri, Anya Zorin, James Clinton, Ali~El Lahib, Hao
  Su, Michael~Thomas Tolley, Sha Yi, and Xiaolong Wang.
\newblock House of dextra: Cross-embodied co-design for dexterous hands.
\newblock In \emph{International Conference on Learning Representations
  (ICLR)}, 2026.
\newblock URL \url{https://api.semanticscholar.org/CorpusID:283466942}.

\bibitem[Gao et~al.(2026)Gao, Li, Shi, Li, Zhang, Figueroa, and
  Jayaraman]{gao2026vlmgineer}
George Gao, Tianyu Li, Junyao Shi, Yihan Li, Zizhe Zhang, Nadia Figueroa, and
  Dinesh Jayaraman.
\newblock Vlmgineer: Vision-language models as robotic toolsmiths.
\newblock In \emph{International Conference on Learning Representations},
  volume 2026, pp.\  91732--91763, 2026.

\bibitem[Geles et~al.(2026)Geles, Bauersfeld, Wulfmeier, and
  Scaramuzza]{geles2026superhumansafeagileracing}
Ismail Geles, Leonard Bauersfeld, Markus Wulfmeier, and Davide Scaramuzza.
\newblock Superhuman safe and agile racing through multi-agent reinforcement
  learning, 2026.
\newblock URL \url{https://arxiv.org/abs/2605.22748}.

\bibitem[Ha et~al.(2026)Ha, Liu, and Song]{ha2026transformer}
Huy Ha, C~Karen Liu, and Shuran Song.
\newblock Transformer transformer: A unified model for motion-conditioned robot
  co-design.
\newblock \emph{arXiv preprint arXiv:2607.25798}, 2026.

\bibitem[Ha et~al.(2017)Ha, Coros, Alspach, Kim, and Yamane]{ha2017joint}
Sehoon Ha, Stelian Coros, Alexander Alspach, Joohyung Kim, and Katsu Yamane.
\newblock Joint optimization of robot design and motion parameters using the
  implicit function theorem.
\newblock In \emph{Robotics: Science and systems}, volume~13, pp.\  10--15607,
  2017.

\bibitem[Hejna~III et~al.(2021)Hejna~III, Abbeel, and Pinto]{hejna2021task}
Donald~J Hejna~III, Pieter Abbeel, and Lerrel Pinto.
\newblock Task-agnostic morphology evolution.
\newblock \emph{arXiv preprint arXiv:2102.13100}, 2021.

\bibitem[Henri et~al.(2020)Henri, Levent, Halev, Alami, and
  Cordier]{Henri2020pymgridAO}
Gonzague Henri, Tanguy Levent, Avishai Halev, R{\'e}da Alami, and Philippe
  Cordier.
\newblock pymgrid: An open-source python microgrid simulator for applied
  artificial intelligence research.
\newblock \emph{ArXiv}, abs/2011.08004, 2020.
\newblock URL \url{https://api.semanticscholar.org/CorpusID:226964357}.

\bibitem[Hiller \& Lipson(2011)Hiller and Lipson]{hiller2011automatic}
Jonathan Hiller and Hod Lipson.
\newblock Automatic design and manufacture of soft robots.
\newblock \emph{IEEE Transactions on Robotics}, 28\penalty0 (2):\penalty0
  457--466, 2011.

\bibitem[Jeon et~al.(2026)Jeon, Eftekhar, Walsman, Zeng, Farhadi, and
  Krishna]{jeon2026convergent}
Hyeonseong Jeon, Ainaz Eftekhar, Aaron Walsman, Kuo-Hao Zeng, Ali Farhadi, and
  Ranjay Krishna.
\newblock Convergent functions, divergent forms.
\newblock \emph{Advances in Neural Information Processing Systems},
  38:\penalty0 46049--46076, 2026.

\bibitem[Kriegman et~al.(2020)Kriegman, Blackiston, Levin, and
  Bongard]{kriegman2020scalable}
Sam Kriegman, Douglas Blackiston, Michael Levin, and Josh Bongard.
\newblock A scalable pipeline for designing reconfigurable organisms.
\newblock \emph{Proceedings of the National Academy of Sciences}, 117\penalty0
  (4):\penalty0 1853--1859, 2020.

\bibitem[Lord \& Miller(2026)Lord and Miller]{lordmiller2026projecthailmary}
Phil Lord and Christopher Miller.
\newblock Project hail mary.
\newblock Film, March 2026.
\newblock Screenplay by Drew Goddard, based on the novel by Andy Weir. Starring
  Ryan Gosling.

\bibitem[Luyben \& Floudas(1994)Luyben and Floudas]{luyben1994analyzing}
ML~Luyben and CA~Floudas.
\newblock Analyzing the interaction of design and control---1. a multiobjective
  framework and application to binary distillation synthesis.
\newblock \emph{Computers \& chemical engineering}, 18\penalty0 (10):\penalty0
  933--969, 1994.

\bibitem[Ma et~al.(2021)Ma, Du, Zhang, Wu, Spielberg, Katzschmann, and
  Matusik]{ma2021diffaqua}
Pingchuan Ma, Tao Du, John~Z Zhang, Kui Wu, Andrew Spielberg, Robert~K
  Katzschmann, and Wojciech Matusik.
\newblock Diffaqua: A differentiable computational design pipeline for soft
  underwater swimmers with shape interpolation.
\newblock \emph{ACM Transactions on Graphics (TOG)}, 40\penalty0 (4):\penalty0
  1--14, 2021.

\bibitem[Macklin(2022)]{macklin_warp_2022}
Miles Macklin.
\newblock Warp: {A} {High}-performance {Python} {Framework} for {GPU}
  {Simulation} and {Graphics}, March 2022.
\newblock URL \url{https://github.com/nvidia/warp}.

\bibitem[Margolis et~al.(2021)Margolis, Chen, Paigwar, Fu, Kim, Kim, and
  Agrawal]{margolis2021learning}
Gabriel~B Margolis, Tao Chen, Kartik Paigwar, Xiang Fu, Donghyun Kim, Sangbae
  Kim, and Pulkit Agrawal.
\newblock Learning to jump from pixels.
\newblock \emph{arXiv preprint arXiv:2110.15344}, 2021.

\bibitem[Matthews et~al.(2023)Matthews, Spielberg, Rus, Kriegman, and
  Bongard]{matthews2023efficient}
David Matthews, Andrew Spielberg, Daniela Rus, Sam Kriegman, and Josh Bongard.
\newblock Efficient automatic design of robots.
\newblock \emph{Proceedings of the National Academy of Sciences}, 120\penalty0
  (41):\penalty0 e2305180120, 2023.

\bibitem[Milder et~al.(2012)Milder, Franchetti, Hoe, and
  P{\"u}schel]{milder2012computer}
Peter Milder, Franz Franchetti, James~C Hoe, and Markus P{\"u}schel.
\newblock Computer generation of hardware for linear digital signal processing
  transforms.
\newblock \emph{ACM Transactions on Design Automation of Electronic Systems
  (TODAES)}, 17\penalty0 (2):\penalty0 1--33, 2012.

\bibitem[Minaee et~al.(2024)Minaee, Mikolov, Nikzad, Chenaghlu, Socher,
  Amatriain, and Gao]{Minaee2024LargeLM}
Shervin Minaee, Tom{\'a}s Mikolov, Narjes Nikzad, Meysam~Asgari Chenaghlu,
  Richard Socher, Xavier Amatriain, and Jianfeng Gao.
\newblock Large language models: A survey.
\newblock \emph{ArXiv}, abs/2402.06196, 2024.
\newblock URL \url{https://api.semanticscholar.org/CorpusID:267617032}.

\bibitem[Mnih et~al.(2015)Mnih, Kavukcuoglu, Silver, Rusu, Veness, Bellemare,
  Graves, Riedmiller, Fidjeland, Ostrovski, Petersen, Beattie, Sadik,
  Antonoglou, King, Kumaran, Wierstra, Legg, and Hassabis]{Mnih2013PlayingAW}
Volodymyr Mnih, Koray Kavukcuoglu, David Silver, Andrei~A. Rusu, Joel Veness,
  Marc~G. Bellemare, Alex Graves, Martin Riedmiller, Andreas~K. Fidjeland,
  Georg Ostrovski, Stig Petersen, Charles Beattie, Amir Sadik, Ioannis
  Antonoglou, Helen King, Dharshan Kumaran, Daan Wierstra, Shane Legg, and
  Demis Hassabis.
\newblock Human-level control through deep reinforcement learning.
\newblock \emph{Nature}, 518\penalty0 (7540):\penalty0 529--533, February 2015.
\newblock ISSN 00280836.
\newblock URL \url{http://dx.doi.org/10.1038/nature14236}.

\bibitem[Morgan et~al.(2026)Morgan, Xie, Bass, Thomasson, Wang, Shahriari,
  Busa, Araromi, Aronov, Burgess, Dimitrov, Estrada, Hamdi, Kendig, Lee,
  Mur-Miranda, Sommers, Titchener, Wang, Wilson, Yeatman, Yirmibesoglu, Rizzi,
  Mozeika, Rojas, and Odhner]{Morgan_2026}
Andrew~S. Morgan, Gregory Xie, Capprin Bass, Rachel Thomasson, Chunpeng Wang,
  Erfan Shahriari, Harrison Busa, Oluwaseun Araromi, Joseph Aronov, Michael
  Burgess, Velin~D. Dimitrov, Matthew~A. Estrada, Faris Hamdi, Samuel Kendig,
  Taeyoon Lee, Jose~Oscar Mur-Miranda, Emma Sommers, Paul Titchener, Margaret
  Wang, Achu Wilson, Mark Yeatman, Osman~Dogan Yirmibesoglu, Alfred~A. Rizzi,
  Annan Mozeika, Nicolas Rojas, and Lael Odhner.
\newblock Athenazero: A low-inertia, bimanual robot for dynamic manipulation.
\newblock \emph{Science Robotics}, 11\penalty0 (118), September 2026.
\newblock ISSN 2470-9476.
\newblock \doi{10.1126/scirobotics.aee1868}.
\newblock URL \url{http://dx.doi.org/10.1126/scirobotics.aee1868}.

\bibitem[O'Kelly et~al.(2020)O'Kelly, Zheng, Karthik, and
  Mangharam]{okelly2020f1tenth}
Matthew O'Kelly, Hongrui Zheng, Dhruv Karthik, and Rahul Mangharam.
\newblock F1tenth: An open-source evaluation environment for continuous control
  and reinforcement learning.
\newblock In \emph{NeurIPS 2019 Competition and Demonstration Track}, pp.\
  77--89. PMLR, 2020.

\bibitem[Park \& Asada(1994)Park and Asada]{park1994concurrent}
Jahng-Hyon Park and Haruhiko Asada.
\newblock Concurrent design optimization of mechanical structure and control
  for high speed robots.
\newblock 1994.

\bibitem[Peng et~al.(2017)Peng, Berseth, Yin, and Van
  De~Panne]{peng2017deeploco}
Xue~Bin Peng, Glen Berseth, KangKang Yin, and Michiel Van De~Panne.
\newblock Deeploco: Dynamic locomotion skills using hierarchical deep
  reinforcement learning.
\newblock \emph{Acm transactions on graphics (tog)}, 36\penalty0 (4):\penalty0
  1--13, 2017.

\bibitem[Peng et~al.(2018)Peng, Abbeel, Levine, and Van~de
  Panne]{peng2018deepmimic}
Xue~Bin Peng, Pieter Abbeel, Sergey Levine, and Michiel Van~de Panne.
\newblock Deepmimic: Example-guided deep reinforcement learning of
  physics-based character skills.
\newblock \emph{ACM Transactions On Graphics (TOG)}, 37\penalty0 (4):\penalty0
  1--14, 2018.

\bibitem[Rapin \& Teytaud(2018)Rapin and Teytaud]{nevergrad}
J.~Rapin and O.~Teytaud.
\newblock {Nevergrad - A gradient-free optimization platform}.
\newblock \url{https://GitHub.com/FacebookResearch/Nevergrad}, 2018.

\bibitem[Sahovic(2019)]{hsahovic_poke_env_2019}
Haris Sahovic.
\newblock poke-env, 2019.
\newblock URL \url{https://github.com/hsahovic/poke-env}.
\newblock GitHub repository. Poke-env: Python Interface for Pokemon Showdown
  Bots. Accessed 2026-09-23.

\bibitem[Salama et~al.(1989)Salama, Ramaker, and Milman]{salama1989recent}
M~Salama, R~Ramaker, and M~Milman.
\newblock Recent experience in simultaneous control-structure optimization.
\newblock \emph{NASA, Langley Research Center, NASA (DOD Controls-Structures
  Interaction Technology 1989}, 1989.

\bibitem[Schneider et~al.(2022)Schneider, Werner, Khalili, Hecker, and
  Karl]{schneider2022mobileenv}
Stefan Schneider, Stefan Werner, Ramin Khalili, Artur Hecker, and Holger Karl.
\newblock mobile-env: An open platform for reinforcement learning in wireless
  mobile networks.
\newblock In \emph{Network Operations and Management Symposium (NOMS)}.
  IEEE/IFIP, 2022.

\bibitem[Schulman et~al.(2017)Schulman, Wolski, Dhariwal, Radford, and
  Klimov]{Schulman2017ProximalPO}
John~D. Schulman, Filip Wolski, Prafulla Dhariwal, Alec Radford, and Oleg
  Klimov.
\newblock Proximal policy optimization algorithms.
\newblock \emph{ArXiv}, abs/1707.06347, 2017.
\newblock URL \url{https://api.semanticscholar.org/CorpusID:28695052}.

\bibitem[Seo et~al.(2025)Seo, Sferrazza, Geng, Nauman, Yin, and
  Abbeel]{seo_fasttd3_2025}
Younggyo Seo, Carmelo Sferrazza, Haoran Geng, Michal Nauman, Zhao-Heng Yin, and
  Pieter Abbeel.
\newblock {FastTD3}: {Simple}, {Fast}, and {Capable} {Reinforcement} {Learning}
  for {Humanoid} {Control}.
\newblock 2025.
\newblock \doi{10.48550/ARXIV.2505.22642}.
\newblock URL \url{https://arxiv.org/abs/2505.22642}.
\newblock Version Number: 3.

\bibitem[Silver et~al.(2016)Silver, Huang, Maddison, Guez, Sifre, van~den
  Driessche, Schrittwieser, Antonoglou, Panneershelvam, Lanctot, Dieleman,
  Grewe, Nham, Kalchbrenner, Sutskever, Lillicrap, Leach, Kavukcuoglu, Graepel,
  and Hassabis]{Silver2016MasteringTG}
David Silver, Aja Huang, Chris~J. Maddison, Arthur Guez, L.~Sifre, George
  van~den Driessche, Julian Schrittwieser, Ioannis Antonoglou, Vedavyas
  Panneershelvam, Marc Lanctot, Sander Dieleman, Dominik Grewe, John Nham, Nal
  Kalchbrenner, Ilya Sutskever, Timothy~P. Lillicrap, Madeleine Leach, Koray
  Kavukcuoglu, Thore Graepel, and Demis Hassabis.
\newblock Mastering the game of go with deep neural networks and tree search.
\newblock \emph{Nature}, 529:\penalty0 484--489, 2016.
\newblock URL \url{https://api.semanticscholar.org/CorpusID:515925}.

\bibitem[Sims(1994)]{sims1994evolving}
Karl Sims.
\newblock Evolving virtual creatures.
\newblock In \emph{Proceedings of the 21st Annual Conference on Computer
  Graphics and Interactive Techniques (SIGGRAPH)}, pp.\  15--22. 1994.

\bibitem[Spall(1992)]{spall1992spsa}
James~C. Spall.
\newblock Multivariate stochastic approximation using a simultaneous
  perturbation gradient approximation.
\newblock \emph{IEEE Transactions on Automatic Control}, 37\penalty0
  (3):\penalty0 332--341, 1992.

\bibitem[Spielberg et~al.(2017)Spielberg, Araki, Sung, Tedrake, and
  Rus]{spielberg2017functional}
Andrew Spielberg, Brandon Araki, Cynthia Sung, Russ Tedrake, and Daniela Rus.
\newblock Functional co-optimization of articulated robots.
\newblock In \emph{2017 IEEE International Conference on Robotics and
  Automation (ICRA)}, pp.\  5035--5042. IEEE, 2017.

\bibitem[Spielberg et~al.(2019)Spielberg, Zhao, Hu, Du, Matusik, and
  Rus]{spielberg2019learning}
Andrew Spielberg, Allan Zhao, Yuanming Hu, Tao Du, Wojciech Matusik, and
  Daniela Rus.
\newblock Learning-in-the-loop optimization: End-to-end control and co-design
  of soft robots through learned deep latent representations.
\newblock \emph{Advances in Neural Information Processing Systems}, 32, 2019.

\bibitem[Spielberg et~al.(2023)Spielberg, Du, Hu, Rus, and
  Matusik]{spielberg2023advanced}
Andrew Spielberg, Tao Du, Yuanming Hu, Daniela Rus, and Wojciech Matusik.
\newblock Advanced soft robot modeling in chainqueen.
\newblock \emph{Robotica}, 41\penalty0 (1):\penalty0 74--104, 2023.

\bibitem[Todorov et~al.(2012)Todorov, Erez, and Tassa]{todorov_mujoco_2012}
Emanuel Todorov, Tom Erez, and Yuval Tassa.
\newblock {MuJoCo}: {A} physics engine for model-based control.
\newblock In \emph{2012 {IEEE}/{RSJ} {International} {Conference} on
  {Intelligent} {Robots} and {Systems}}, pp.\  5026--5033. IEEE, 2012.
\newblock \doi{10.1109/IROS.2012.6386109}.

\bibitem[Towers et~al.(2024)Towers, Kwiatkowski, Terry, Balis, Cola, Deleu,
  Goul{\~a}o, Kallinteris, Krimmel, Arjun, Perez-Vicente, Pierr{\'e},
  Schulhoff, Tai, Tan, and Younis]{Towers2024GymnasiumAS}
Mark Towers, Ariel Kwiatkowski, Jordan~K. Terry, John~U. Balis, Gian Marco~De
  Cola, Tristan Deleu, Manuel Goul{\~a}o, Andreas Kallinteris, Markus Krimmel,
  KG~Arjun, Rodrigo Perez-Vicente, Andrea Pierr{\'e}, Sander Schulhoff, Jun~Jet
  Tai, Hannah Tan, and Omar~G. Younis.
\newblock Gymnasium: A standard interface for reinforcement learning
  environments.
\newblock 2024.
\newblock URL \url{https://api.semanticscholar.org/CorpusID:290088643}.

\bibitem[University(2011)]{smogon_pokemon_showdown_2011}
Smogon University.
\newblock pokemon-showdown, 2011.
\newblock URL \url{https://github.com/smogon/pokemon-showdown}.
\newblock GitHub repository. Pokémon battle simulator.. Accessed 2026-09-12.

\bibitem[Ventrella(1994)]{ventrella1994explorations}
Jeffrey Ventrella.
\newblock Explorations in the emergence of morphology and locomotion behavior
  in animated characters.
\newblock 1994.

\bibitem[Wampler \& Popovi{\'c}(2009)Wampler and
  Popovi{\'c}]{wampler2009optimal}
Kevin Wampler and Zoran Popovi{\'c}.
\newblock Optimal gait and form for animal locomotion.
\newblock \emph{ACM Transactions on Graphics (TOG)}, 28\penalty0 (3):\penalty0
  1--8, 2009.

\bibitem[Wang et~al.(2026{\natexlab{a}})Wang, Fang, Hanna, and
  Xiong]{wang2026efficient}
Hanwen Wang, Zhenlong Fang, Josiah Hanna, and Xiaobin Xiong.
\newblock Efficient and versatile quadrupedal skating: Optimal co-design via
  reinforcement learning and bayesian optimization.
\newblock \emph{arXiv preprint arXiv:2603.18408}, 2026{\natexlab{a}}.

\bibitem[Wang et~al.(2026{\natexlab{b}})Wang, Qian, Wang, Wu, Zong, Gupta, and
  Pathak]{wang2026bridge}
Jianren Wang, Letian Qian, Zikai Wang, Weiwei Wu, Junjie Zong, Abhinav Gupta,
  and Deepak Pathak.
\newblock Bridge: An open-source humanoid platform via morphology-control
  co-design for physical ai.
\newblock \emph{arXiv preprint arXiv:2609.03497}, 2026{\natexlab{b}}.

\bibitem[Wang et~al.(2020)Wang, Fonseca, and Tian]{wang2020lamcts}
Linnan Wang, Rodrigo Fonseca, and Yuandong Tian.
\newblock Learning search space partition for black-box optimization using
  {M}onte {C}arlo tree search.
\newblock In \emph{Advances in Neural Information Processing Systems},
  volume~33, pp.\  19511--19522, 2020.

\bibitem[Wang et~al.(2019)Wang, Zhou, Fidler, and Ba]{wang2019neural}
Tingwu Wang, Yuhao Zhou, Sanja Fidler, and Jimmy Ba.
\newblock Neural graph evolution: Towards efficient automatic robot design.
\newblock \emph{arXiv preprint arXiv:1906.05370}, 2019.

\bibitem[Wang et~al.(2023{\natexlab{a}})Wang, Ma, Spielberg, Xian, Zhang,
  Tenenbaum, Rus, and Gan]{wang2023softzoo}
Tsun-Hsuan Wang, Pingchuan Ma, Andrew~Everett Spielberg, Zhou Xian, Hao Zhang,
  Joshua~B. Tenenbaum, Daniela Rus, and Chuang Gan.
\newblock Softzoo: A soft robot co-design benchmark for locomotion in diverse
  environments.
\newblock In \emph{The Eleventh International Conference on Learning
  Representations}, 2023{\natexlab{a}}.
\newblock URL \url{https://openreview.net/forum?id=Xyme9p1rpZw}.

\bibitem[Wang et~al.(2023{\natexlab{b}})Wang, Zheng, Ma, Du, Kim, Spielberg,
  Tenenbaum, Gan, and Rus]{wang2023diffusebot}
Tsun-Hsuan~Johnson Wang, Juntian Zheng, Pingchuan Ma, Yilun Du, Byungchul Kim,
  Andrew Spielberg, Josh Tenenbaum, Chuang Gan, and Daniela Rus.
\newblock Diffusebot: Breeding soft robots with physics-augmented generative
  diffusion models.
\newblock \emph{Advances in Neural Information Processing Systems},
  36:\penalty0 44398--44423, 2023{\natexlab{b}}.

\bibitem[Weir(2021)]{weir2021projecthailmary}
Andy Weir.
\newblock \emph{Project Hail Mary}.
\newblock Ballantine Books, New York, 2021.
\newblock ISBN 9780593135204.

\bibitem[Xu et~al.(2021)Xu, Spielberg, Zhao, Rus, and Matusik]{xu2021multi}
Jie Xu, Andrew Spielberg, Allan Zhao, Daniela Rus, and Wojciech Matusik.
\newblock Multi-objective graph heuristic search for terrestrial robot design.
\newblock In \emph{2021 IEEE international conference on robotics and
  automation (ICRA)}, pp.\  9863--9869. IEEE, 2021.

\bibitem[Xu et~al.(2024)Xu, Ha, and Song]{xu2024dynamics}
Xiaomeng Xu, Huy Ha, and Shuran Song.
\newblock Dynamics-guided diffusion model for sensor-less robot manipulator
  design.
\newblock \emph{arXiv preprint arXiv:2402.15038}, 2024.

\bibitem[Yu et~al.(2026)Yu, Matthews, Wang, Gu, Blackiston, Rubenstein, and
  Kriegman]{yu2026agile}
Chen Yu, David Matthews, Jingxian Wang, Jing Gu, Douglas Blackiston, Michael
  Rubenstein, and Sam Kriegman.
\newblock Agile legged locomotion in reconfigurable modular robots.
\newblock \emph{Proceedings of the National Academy of Sciences}, 123\penalty0
  (10):\penalty0 e2519129123, 2026.

\bibitem[Yuan et~al.(2021)Yuan, Song, Luo, Sun, and
  Kitani]{yuan2021transform2act}
Ye~Yuan, Yuda Song, Zhengyi Luo, Wen Sun, and Kris Kitani.
\newblock Transform2act: Learning a transform-and-control policy for efficient
  agent design.
\newblock \emph{arXiv preprint arXiv:2110.03659}, 2021.

\bibitem[Zardini(2023)]{zardini2023co}
Gioele Zardini.
\newblock \emph{Co-design of complex systems: From autonomy to future mobility
  systems}.
\newblock PhD thesis, ETH Zurich, 2023.

\bibitem[Zhao et~al.(2020)Zhao, Xu, Konakovi{\'c}-Lukovi{\'c}, Hughes,
  Spielberg, Rus, and Matusik]{zhao2020robogrammar}
Allan Zhao, Jie Xu, Mina Konakovi{\'c}-Lukovi{\'c}, Josephine Hughes, Andrew
  Spielberg, Daniela Rus, and Wojciech Matusik.
\newblock Robogrammar: graph grammar for terrain-optimized robot design.
\newblock \emph{ACM Transactions on Graphics (TOG)}, 39\penalty0 (6):\penalty0
  1--16, 2020.

\end{thebibliography}
\bibliographystyle{thestyle}
\clearpage

\end{document}